\pdfoutput=1
\documentclass[11pt]{article}

\usepackage[final]{acl}

\usepackage{times}
\usepackage{latexsym}

\usepackage[T1]{fontenc}
\usepackage[utf8]{inputenc}

\usepackage{microtype}

\usepackage{inconsolata}

\usepackage{graphicx}
\usepackage{times}
\usepackage{latexsym}
\usepackage{amsmath}
\usepackage{amsfonts}
\usepackage{amssymb}
\usepackage{pifont}% http://ctan.org/pkg/pifont
\usepackage{hyperref}
\usepackage[nameinlink]{cleveref}
\usepackage{soul}
\usepackage{algorithm}
\usepackage{algpseudocode}
\usepackage{booktabs}
\usepackage{multirow}
\usepackage{arydshln}
\usepackage{subfigure}
\usepackage{lipsum}
\usepackage{xcolor}
\usepackage{mathtools}
\usepackage{amssymb}
\usepackage{amsthm}
\usepackage{soul}

\newcommand{\best}[1]{{\textbf{#1}}}
\newcommand{\framework}[1]{{\textsc{#1}}}
\title{\framework{Lion}s: An Empirically Optimized Approach to Align Language Models}
\author{Xiao Yu\thanks{~~denotes equal contribution.},
  Qingyang Wu$^{*}$,
  Yu Li,
  Zhou Yu
  \\
  Columbia University \\
  \texttt{\{xy2437, qw2345, yl5016, zy2461\}@columbia.edu}
}

\begin{document}
\maketitle
\begin{abstract}
% Background
Alignment is a crucial step to enhance the instruction-following and conversational abilities of language models.
% However,..
Despite many recent work proposing new algorithms, datasets, and training pipelines, there is a lack of comprehensive studies measuring the impact of various design choices throughout the whole training process.
% Despite many papers and libraries providing analysis and training platform for alignment, there is a lack of comprehensive studies covering the entire language model alignment pipeline.
% What we do
We first conduct a rigorous analysis over a three-stage training pipeline consisting of supervised fine-tuning, offline preference learning, and online preference learning.
% We conduct a rigorous analysis of the best strategies for supervised fine-tuning, offline, and online preference learning.
We have found that using techniques like sequence packing, loss masking in SFT, increasing the preference dataset size in DPO, and online DPO training can significantly improve the performance of language models.
We then train from Gemma-2b-base and LLama-3-8b-base, and find that our best models exceed the performance of the official instruct models tuned with closed-source data and algorithms.
Our code and models can be found at \url{https://github.com/Columbia-NLP-Lab/LionAlignment}.
% We will make our models, training datasets, and code publicly available. 

% Our best models, trained from Gemma-2b-base and LLama-3-8b-base, exceed the performance of officially tuned instruct counterparts while only using publicly available datasets and open-source algorithms.
% We provide an easily reproducible training recipe for future alignment research.\footnote{We make our models, curated datasets, and code publicly available.}
% We find that our model, tuned from the base models, can exceed the performance of official tuned instruct models while only using publicly available datasets.
% We find that our model can exceed the performance of official tuned instruct models while only using publicly available datasets.
% 1) replicating the complete alignment pipeline
% 2) exceed officially released instruct models
% We present a fully reproducible training pipeline and trained our models using publicly available datasets.
% Our models, trained from Gemma-2b and LLaMA-3 8b base models, have achieved comparable performance to the state-of-the art models.
% Our code and models will be made publicly available.
\end{abstract}

\section{Introduction}

% Alignment is a pivotal step in fine-tuning language models to enhance their instruction-following and conversational capabilities.
% Since DPO \cite{rafailov2023direct}, many recent work has focused on proposing new algorithms to improve DPO's performance or efficiency \cite{meng2024simpo,gorbatovski2024trdpo,guo2024direct-ipo,ethayarajh2024kto}, creating new datasets \cite{cui2023ultrafeedback,starling2023}, or building new training pipelines \cite{dong2024rlhf-workflow,snorkel2023iterative-dpo}.
% However, it is challenging to determine the source of improvements, and which aspects of these methods contribute the most.

Large language models (LLMs), pre-trained on datasets of trillion-scale tokens, have shown remarkable performance across a wide range of natural language processing tasks \cite{brown2020language,openai2024gpt4,touvron2023llama,llama3modelcard}. However, these pre-trained models often struggle to follow human instructions and generate responses that are unsafe or inappropriate \cite{wei2023jailbroken,deshpande2023toxicity}.
Recent research has increasingly focused on aligning LLMs: this includes many new algorithms based on reinforcement learning \cite{ouyang2022training,rafailov2023direct,meng2024simpo,guo2024direct-ipo}, new datasets to facilitate preference learning \cite{cui2023ultrafeedback,starling2023}, and new training pipelines to improve the overall alignment performance \cite{tunstall2023zephyr,snorkel2023iterative-dpo,dong2024rlhf-workflow}.
Although these contributions demonstrate sizable improvements, the training processes, datasets, and hyper-parameters often remain heterogeneous or closed-source \cite{gemmateam2024gemma,llama3modelcard}.
This makes it difficult to pinpoint the source of improvements and limits the development of more effective or efficient alignment algorithms.
% and it is difficult to pinpoint the sources of these improvements.
% % Despite plentiful works showing improvement in different directions, it becomes increasingly difficult to pinpoint the sources of these improvements and identify which aspects of the methods contribute the most.
% This limits researchers from understanding the critical components in the modern alignment pipeline, thereby hindering the development of more effective and efficient algorithms.

In this work, we replicate modern alignment pipelines \cite{tunstall2023zephyr,xu2024cringe-iterative-dpo} and analyze sources in the training process that could affect performance.
% In this work, we begin with a replication study of modern alignment pipelines , and analyze potential sources in the training process that could impact performance.
In the three-stage training of supervised fine-tuning (SFT), offline preference learning, and online preference learning, we find that: 1) sequence packing and loss masking significantly enhance SFT, 2) scaling offline preference datasets improves overall performance, and 3) online learning greatly benefits chat benchmarks.
% Throughout the three-stage training process, we find that 1) simple techniques such as sequence packing and loss masking can significantly enhance the performance of supervised fine-tuning; 
% 2) scaling offline preference learning datasets and training steps greatly improves final performance; and 3) training with online preference data strongly benefits chat benchmarks.

Then, we aggregate our findings and fine-tune from Gemma-2b-base \cite{gemmateam2024gemma} and LLaMA-3-8b-base \cite{llama3modelcard} using publicly available datasets and open-source algorithms.
We evaluate our models on popular benchmarks such as Arena-Hard \cite{zheng2023judging}, AlpacaEval-2 \cite{alpaca_eval}, MT-Bench \cite{zheng2023judging}, and OpenLLM \cite{open-llm-leaderboard}.
In all benchmarks, our models exceed the performance of the officially instruct models, which rely on closed-source datasets and algorithms.
We believe our easily reproducible study offers useful insights for alignment research, and our fine-tuned models are valuable for downstream applications.
% Compared to the officially released instruct models which relies on closed-source datasets and algorithms, our models 
% We find that our best Gemma-2b model exceeds the performance of various 7B models on benchmarks such as Arena-Hard-Auto \cite{zheng2023judging} and AlpacaEval-2 \cite{alpaca_eval}. 
% Our best LLaMA-3-8b model outperforms the officially released instruct model, while only using publicly available datasets and a simpler alignment pipeline.
% We believe our study offers useful insights for future alignment research, and our fine-tuned models are valuable for future downstream applications.

Our contributions are:
\vspace{-1mm}
\begin{itemize}
    \item We present a rigorous analysis of modern alignment training pipelines, and identify a set of design choices that significantly impact the performance of language models.
    \item We aggregate our empirical findings into a step-by-step recipe, and show that our models outperform the officially released instruct models, which relies on closed-source datasets and algorithms.
    \item We make our model, training dataset, and code publicly available for future research in language model alignment.
    % \item We aggregate our empirical findings into a step-by-step recipe, and release the \framework{Lion} series of models trained from Gemma-2b-base and LLaMA-3-8b-base, the datasets, and the code.
    % \item We show that using only publicly available datasets and a simple training pipeline, our models exceed the performance of the officially released instruct models as well as models of larger sizes.
\end{itemize}

\newcommand{\piref}{\pi_\text{ref}}
\newcommand{\pisft}{\pi^\text{SFT}} %

\section{Preliminaries}
\label{sec:Preliminaries}
Traditional reinforcement learning for human feedback (RLHF) methods typically starts with supervised fine-tuning, then trains a reward model $r$ mimicking human preferences, and finally optimizes a language model $\pi_\theta$ to maximize the reward \cite{ziegler2020finetuning,bai2022training,ouyang2022training}.
However, this process can be complex and difficult to tune.
Many recent works \cite{snorkel2023iterative-dpo,xu2024cringe-iterative-dpo} have proposed alternative algorithms.
In this work, we examine a three-stage RLHF pipeline:
1) supervised fine-tuning; 2) offline preference learning with DPO \cite{rafailov2023direct}; 
and 3) online preference learning with DPO. 
Below, we review each training stage.
% Training language models (LM) to follow human preferences \cite{ziegler2020finetuning,bai2022training,ouyang2022training} in the past utilizes a two-stage training process: 1) supervised finetuning; 2) reinforcement learning \cite{schulman2017proximal} under a trained reward model.
% However, prior work such as DPO \cite{rafailov2023direct} found that the second training stage can be simplified without much performance trade-offs.
% This enables many work \cite{snorkel2023iterative-dpo,xu2024cringe-iterative-dpo} to consider a three-stage training process: 1) supervised finetuning; 2) offline preference learning with DPO; and 3) online preference learning. We review each training stage below.

\paragraph{Supervised Fine-tuning Stage} Supervised Fine-tuning (SFT) maximizes the log likelihood of the ground truth response $y$ given a user's query $x$:  
\begin{equation*}
    p(y|x) = \frac{1}{|y|} \sum_{i=1}^{|y|} \log \pi_\theta (y_i | x, y_{<i}).
\end{equation*}
Efficiently optimizing the above objective is a challenging problem.
We explore various strategies for SFT, which are detailed in \Cref{subsec:Supervied Finetuning}.

\paragraph{Offline Preference Learning Stage} In the second phase, the SFT model $\pisft$ is trained offline with a collection of human preference data.
% Unlike the traditional approach of first learning a reward model $r$ and then optimizing an LM $\pi_\theta$ to maximize the reward \cite{ouyang2022training,schulman2017proximal}, recent methods such as Direct Preference Optimization (DPO) simplify the process \cite{rafailov2023direct}.
This is often done using DPO, which first re-parametrizes the reward $r$ in terms of the optimal policy\footnote{Under the formulation of reward maximization under a KL-divergence constraint}:
\[
r(x, y) = \beta \log \frac{\pi_\theta(y|x)}{\piref(y|x)} + \beta \log Z(x)
\] 
where $\piref$ is a reference policy, $\beta$ controls the strength of the KL-divergence, and $Z(x)$ is the partition function. Then, DPO optimizes the following objective treating the LMs as reward models:
\begin{equation*}
\label{eq:optimum_model}
    \mathcal{L}_\text{DPO} = -\mathbb{E}_{(x, y_w, y_l)\sim \mathcal{D}}\left[\log \sigma \left( r_w-r_l \right)\right].
\end{equation*}
where $(x, y_w, y_l)$ are preference pairs consisting of the prompt, the winning response, and the losing response, respectively; and $r_{w} = r(x_w, y_w)$, $r_{l} = r(x_l, y_l)$ are rewards given a choice of $\pi_\theta$ and $\piref$.
% A common choice is to use $\pi_\theta = \piref = \pisft$.
Despite its simplicity, practical concerns include the choice of reference model $\piref$, strength of KL-divergence $\beta$, scalability of training, and data filtering. We will analyze these in \Cref{subsec:Offline Preference Learning}.

\paragraph{Online Preference Learning Stage} After offline preference learning, the model can be further fine-tuned with online preference pairs to improve its performance \cite{xu2024cringe-iterative-dpo,guo2024onlinedpo,snorkel2023iterative-dpo}. 
Similar to the traditional RLHF pipeline, the process includes: 1) sampling multiple responses from $\pi_\theta$; 2) use a reward model or judge to rank the responses; and 3) optimize $\pi_\theta$ using DPO. 
In Section~\ref{subsec:Online Preference Learning}, we will examine the effectiveness of further online learning against the simple offline training process.

% Despites its simplicity, recent work has found that \cite{tunstall2023zephyr,dong2024rlhf-workflow} preference learning methods still require a multi-stage training process. Typically, $\pi_\theta$ is first trained with supervised fine-tuning (SFT) on a large dataset of high-quality responses to obtain $\pisft$, and then $\pisft$ is trained with DPO on a small dataset of preference pairs \cite{tunstall2023zephyr}. 
% The final model can be additionally fine-tuned with online preference pairs to further improve performance \cite{xu2024cringe-iterative-dpo,guo2024onlinedpo,snorkel2023iterative-dpo}. 
% A common choice to \# TODO for SFT; and to initialize $\pi_\theta = \piref = \pisft$ for DPO or online DPO training.

\section{Alignment Procedure Analysis}
\label{sec:Training Procedure Analysis}
This section explores and quantifies which choices are important to align language models.
We introduce our experimental setups in \Cref{subsec:Experiment Setup}, and analyze different configurations in each stage of the training process in \Cref{subsec:Supervied Finetuning}, \Cref{subsec:Offline Preference Learning}, and \Cref{subsec:Online Preference Learning}, respectively.
For a controlled study, we fix the model architecture to Gemma-2b \citep{gemmateam2024gemma}.

% For a controlled study, we keep the model architecture fixed to Gemma-2b \citep{gemmateam2024gemma}, and compare different configurations in each stage of the training process.

\subsection{Experiment Setup}
\label{subsec:Experiment Setup}

\paragraph{SFT Training Data}

To ensure the reproducibility, we carefully selected open-source, high-quality datasets for supervised fine-tuning. 
These include OpenHermes-2.5 \cite{OpenHermes2.5}, SlimOrca \cite{SlimOrca,longpre2023flan,mukherjee2023orca}, MetaMathQA \cite{yu2023metamath}, UltraChat \cite{ding2023enhancing}, OrcaMath \cite{mitra2024orcamath}, Capybara \cite{daniele2023amplify-instruct}, and Deita-10k \cite{liu2024what}.
Since OpenHermes-2.5 is a collection of many other smaller open-source datasets, we also implemented a deduplication process to remove duplicate samples across these datasets.
We use the latest version of those datasets to ensure that there is no contamination of data for our evaluation benchmarks. We summarize the dataset statistics in \Cref{tab:sft_data_stats}.

\begin{table}[t]
\centering
\scalebox{0.87}{
    \begin{tabular}{lrr}
    \toprule
    \textbf{Dataset} & \textbf{Samples (\%)} & \textbf{Tokens (\%)} \\ 
    \midrule
    OpenHermes-2.5 & 35.79\% & 28.77\% \\
    MetaMathQA & 20.96\% & 10.14\% \\
    SlimOrca & 19.74\% & 15.42\% \\
    UltraChat & 11.29\% & 27.89\% \\
    OrcaMath & 10.85\% & 7.60\% \\
    Capybara & 0.86\% & 1.68\% \\
    Deita-10k & 0.51\% & 8.51\% \\
    \midrule
    Total Count & 1.84M & 878M \\
    \bottomrule
    \end{tabular}
}
\caption{SFT dataset statistics}
\label{tab:sft_data_stats}
\end{table}

% \begin{table}[t]
% \centering
% \scalebox{0.8}{
%     \begin{tabular}{lrr}
%     \toprule
%     \textbf{Dataset} & \textbf{Samples (\%)} & \textbf{Tokens (\%)} \\ 
%     \midrule
%     OpenHermes-2.5 & 33.75\% & 26.93\% \\
%     MetaMathQA & 19.77\% & 9.49\% \\
%     SlimOrca & 18.62\% & 14.43\% \\
%     UltraChat & 10.65\% & 26.10\% \\
%     OrcaMath & 10.23\% & 7.11\% \\
%     Magicoder-Evol-Instruct & 5.69\% & 6.41\% \\
%     Capybara & 0.81\% & 1.57\% \\
%     Deita-10k & 0.48\% & 7.96\% \\
%     \midrule
%     Total Count & 1.95M & 938M \\
%     \bottomrule
%     \end{tabular}
% }
% \caption{SFT dataset statistics}
% \label{tab:sft_data_stats}
% \end{table}
% 
% 
% 
\begin{table}[!t]
\centering
\scalebox{0.8}{
    \begin{tabular}{lrr}
    \toprule
    \textbf{Dataset} & \textbf{Samples (\%)} & \textbf{Judge} \\ 
    \midrule
    HH-RLHF & 37.83\% & Human \\
    TLDR-Preference & 27.85\% & Human \\
    UltraFeedback & 23.04\% & GPT-4 \\
    Distilabel-Orca & 4.83\% & GPT-4-Turbo \\
    Py-DPO & 3.57\% & GPT-4-Turbo \\
    Distilabel-Capybara & 2.87\% & GPT-4-Turbo \\
    \midrule
    Total Count & 264K & - \\
    \bottomrule
    \end{tabular}
}
\caption{Offline DPO dataset statistics}
\label{tab:dpo_data_stats}
\end{table}
\paragraph{Offline Preference Learning Data} For a controlled study, all models are trained on a fixed pool of pairwise preference dataset. We follow \cite{dong2024rlhf-workflow,tunstall2023zephyr} and manually select a mixture of publicly available datasets such as UltraFeedback \cite{cui2023ultrafeedback}, HH-RLHF \cite{bai2022training}, and TLDR-preferences \cite{stienon2020learning}.
These datasets consist of chat responses generated by a variety of LMs, and the winning/losing response is decided by prompting a judge model (e.g., GPT-4) or by asking human raters. We present the dataset statistics in \Cref{tab:dpo_data_stats}.
For more details on these datasets, please refer to \Cref{subsec:DPO Analysis Dataset Selection}.
Unless otherwise indicated, all dataset \emph{subsets} mentioned in this section are randomly sampled from this 264K mixture.

\paragraph{Online Preference Learning Data}
We follow prior work \cite{meng2024simpo,xu2024cringe-iterative-dpo} and consider using data from UltraFeedback \cite{cui2023ultrafeedback} as prompts. We then sample multiple responses from $\pi_\theta$ and use Pair-RM \cite{jiang2023llmblender} as a judge to obtain preference pairs. This results in an online collected dataset of 60k in size. Unless otherwise indicated, all dataset \emph{subsets} related to online learning are randomly sampled from this 60k dataset.

\paragraph{Evaluation Benchmarks}
We assess our models using OpenLLM \cite{open-llm-leaderboard} and Arena-Hard-Auto \cite{arenahard2024}. The HuggingFace OpenLLM leaderboard evaluates an LM across a diverse set of reasoning, math, and knowledge tasks, and the average score is reported. 
Arena-Hard-Auto evaluates an LM's instruction-following ability using 500 challenging user queries curated from the live Chatbot Arena leaderboard \citep{zheng2023judging}.
To quantify the models' performance, it prompts a judge model (GPT-4-turbo) to compare the generated response against a reference response (by GPT-4), and uses the win rate as the final score.
Since evaluation with GPT-4-turbo is expensive and using GPT-4's answers provides reference answers that are too strong, we use GPT-4-Omni\footnote{We use version GPT-4o-2024-05-13} as the judge model and answers by GPT-3.5-turbo\footnote{We use version GPT-3.5-turbo-0125} as references for \Cref{subsec:Supervied Finetuning,subsec:Offline Preference Learning,subsec:Online Preference Learning}. We denote this modification as \emph{Arena-Hard-Auto*}.
\subsection{Supervised Fine-tuning}
\label{subsec:Supervied Finetuning}

Supervised fine-tuning (SFT) plays a critical role in aligning Large Language Models (LLMs), often serving as the first step of alignment.
However, different techniques, including sequence packing, padding, and loss masking, have been proposed for SFT \cite{vicuna2023,tunstall2023zephyr,shi2024instruction}.
We re-examine the effectiveness of these strategies within the context of alignment.
% However, there lacks comparative studies on the most effective SFT strategies. 
% This work explores and assesses different supervised fine-tuning strategies, which provides better instruction models for the further alignment phases.

\paragraph{Packing}
Packing optimizes the training efficiency by grouping sequences of varying lengths into a single long sequence without requiring any padding.
This technique, commonly used in LLM pre-training, is now also utilized in instruction-based supervised fine-tuning, as implemented by models like Zephyr \cite{tunstall2023zephyr}\footnote{\url{https://github.com/huggingface/alignment-handbook}}.
% We evaluate the performance of packing with and without loss masking in the SFT training process.
% This method ensures that each batch contains an equivalent number of tokens, which improves the overall training efficiency and load balance.
% It is commonly used in LLM pre-training before, and now is utilized in the instruction supervised fine-tuning for training models like Zephyr \cite{tunstall2023zephyr}\footnote{\url{https://github.com/huggingface/alignment-handbook}}.

\paragraph{Padding}
% The second strategy is padding, which is opposite to "packing". 
% It involves padding shorter sequences to a predefined maximum length and truncating longer sequences to the maximum length, which ensures that all sequences in a batch have the same length.
% It is often combined with loss masking to ignore the loss on the instruction and user tokens. 
% Padding is used for training Alpaca \cite{alpaca} and Vicuna \cite{vicuna2023}, and it is implemented in FastChat\footnote{https://github.com/lm-sys/FastChat}
% Padding allows the model to focus on the meaningful part of the data and let the model learn the training data more quickly.
In contrast to packing, padding extends shorter sequences with padding tokens and truncates longer ones to a fixed maximum length.
It is often paired with loss masking, and is implemented in training models like Alpaca \cite{alpaca} and Vicuna \cite{vicuna2023}\footnote{\url{https://github.com/lm-sys/FastChat}}.
% We similarly investigate the performance of padding with and without loss masking.

% In contrast, padding extends shorter sequences with padding tokens and truncates longer ones to a fixed maximum length.
% It is often paired with loss masking to only compute loss on the target output tokens.
% This method is implemented in training models like Alpaca \cite{alpaca} and Vicuna \cite{vicuna2023}\footnote{\url{https://github.com/lm-sys/FastChat}}. 
% By concentrating on significant data, padding expedites the learning process, allowing the model to assimilate training data more effectively.

\paragraph{Loss Masking}
The standard language model training computes loss across all tokens in a sequence. 
Loss masking, however, ignores loss computation on tokens that are not output tokens like user instructions.
It prevents the model from learning irrelevant information, alleviating catastrophic forgetting and overfitting.

% \cite{alignment_handbook2023}

% Non-packing achieved the best results in specific scenarios but showed a significant drop in performance when generalized to other input templates, especially in OpenLLM. Meanwhile, Packing with Tokens Masking consistently showed the best overall performance.

% SFT refers to the process of updating only a small subset of the model's parameters during the training phase. This approach can lead to more efficient training cycles and reduce the risk of overfitting, especially in large language models. Below is a detailed explanation of the SFT strategies employed in our experiments:

% These strategies are tested to determine their efficacy in adapting to different input templates and their overall impact on model robustness and performance in tasks evaluated by the OpenLLM suite.

% To test the effectiveness of different supervised fine-tuning strategies, we conduct experiments on a small dataset Deita \cite{liu2024makes-deita} with 10 thousand examples and 74 million tokens. 
% We also tested a larger dataset including Open-Hermes2.5, MetaMathQA and UltraChat with 1.6 million examples and 600 million tokens, to test the effect of scaling.
% We use a hyperparameter with batch size 32, sequence length with 8192, learning rate 2e-5, and 3 epochs.
% Gemma-2b is used as the base language model here.

\paragraph{}

Loss masking can be used in conjunction with both packing and padding strategies.
Packing without loss masking and padding with loss masking is widely adopted in SFT, but the combination of packing with loss masking is largely unexplored.
We evaluate the performance of these strategies on both small and large datasets. For each dataset size $|D|$, we train all models from Gemma-2b-base over 3 epochs using a batch size of 32, a sequence length of 2,048 tokens, a learning rate of 2e-5.
We then repeat this with $|D|$=10K with DEITA-10k \cite{liu2024makes-deita} and $|D|$=1.6M with Open-Hermes2.5 \cite{OpenHermes2.5}, MetaMathQA \cite{yu2023metamath}, and UltraChat \cite{ding2023enhancing}.
% To evaluate the performance of different supervised fine-tuning strategies, we conducted experiments using both small and large datasets. Initially, we used the Deita dataset \cite{liu2024makes-deita}, including 10k examples with a total of 74 million tokens. 
% To assess the scalability of these strategies, we also tested a larger dataset that includes Open-Hermes2.5 \cite{OpenHermes2.5}, MetaMathQA \cite{yu2023metamath}, and UltraChat \cite{ding2023enhancing}, which collectively contain 1.6 million examples and around 600 million tokens. 
% For these experiments, we use a batch size of 32, a sequence length of 8,192 tokens, a learning rate of \(2 \times 10^{-5}\), and a training duration of three epochs. 
% The Gemma-2b base model served as the starting base language model for training.

\Cref{tab:sft_results} summarizes our results.
We find that combining packing with loss masking consistently yields the best performance across both dataset scales.
We believe this is because other strategies may overfit chat templates: the starting tokens in each batch remain unchanged, leading to poor adaptation to unseen templates used in benchmarks such as OpenLLM.
Next, we find increasing dataset size widens the performance gap between packing with and without loss masking. 
This may be due to the increasing number of user instructions as the dataset size grows, which is unnecessary for the model to learn.
Overall, this indicates that $\pisft$ should be trained with packing and loss masking, over a large collection of high-quality datasets as in \Cref{tab:sft_data_stats}.

\begin{table}[t]
\centering
\scalebox{0.8}{
\begin{tabular}{lcc}
\toprule
\textbf{Model} & \textbf{OpenLLM} & \textbf{Arena Hard Auto*} \\
\midrule
Gemma-2b & 46.51 & - \\
% Gemma-2b-it & 42.75 & 9.0 \\
% gemma-1.1-2b-it & 42.1 & 71.4 & 42.3 & - & 65.4 & 17.7 & xx \\
\midrule
\midrule
$|D|=10K$ \\
\midrule
Padding & 42.49 &  3.1 \\
\quad + Loss Mask & 43.62 & 2.4 \\
Packing & 47.95 &  5.1  \\
\quad + Loss Mask & \textbf{48.34} & \textbf{5.2} \\
\midrule
\midrule
$|D|=1.6M$ \\
\midrule
Packing & 47.14 & 3.9 \\
\quad + Loss Mask & \textbf{53.80} & \textbf{8.8} \\
\bottomrule
\end{tabular}
}
\caption{Performance comparison for different SFT strategies on OpenLLM and Arena-Hard-Auto*.}
\vspace{-5pt}
\label{tab:sft_results}
\end{table}
\begin{figure*}[t!]
    \centering
    \subfigure[OpenLLM]{
        \includegraphics[scale=0.36]{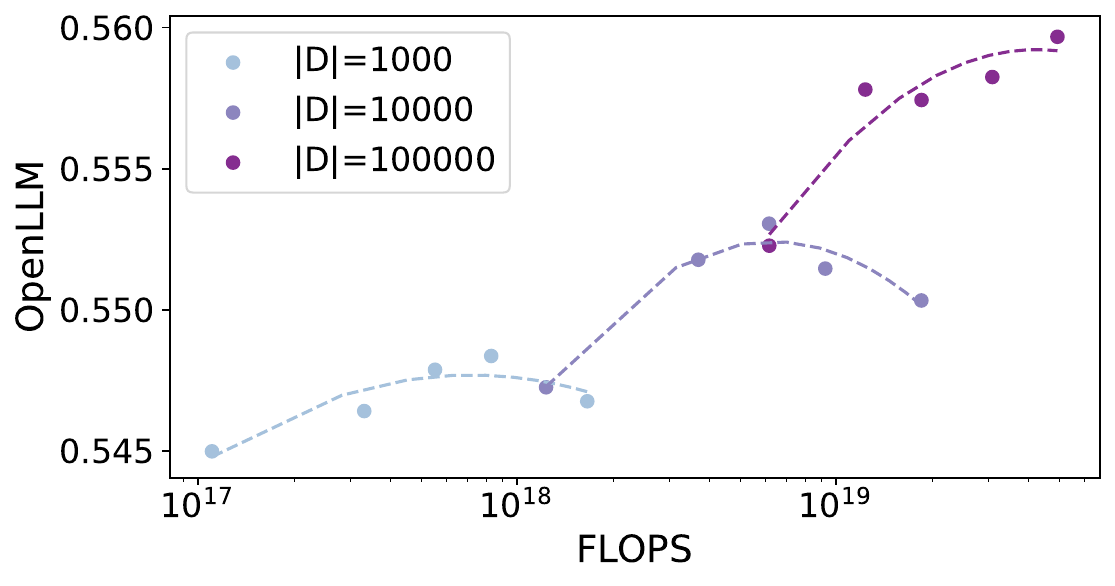}
    }
    \qquad
    \subfigure[Arena Hard Auto*]{
        \includegraphics[scale=0.36]{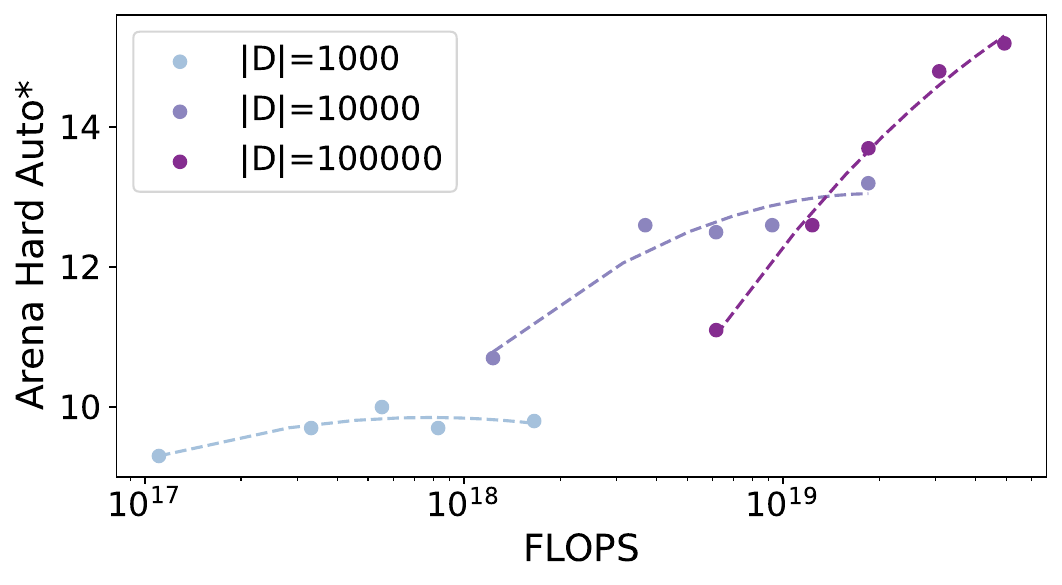}
    }
\caption{
Measuring the effect of dataset size ($|D|$) and training steps (FLOPs) on final performance. While performance can quickly saturate given a fixed $|D|$, increasing the dataset size increases the point of saturation. Dotted lines are our interpolation using a degree 2 polynomial.
% Varying the number of training steps with DPO. For each dataset size $|D|$, we vary the number of training steps and record their performance changes. Dotted lines are our interpolation using an degree 2 polynomial.
}
\label{fig:scaling-dpo}
\end{figure*}
\begin{figure}[t!]
    \centering
    \includegraphics[scale=0.28]{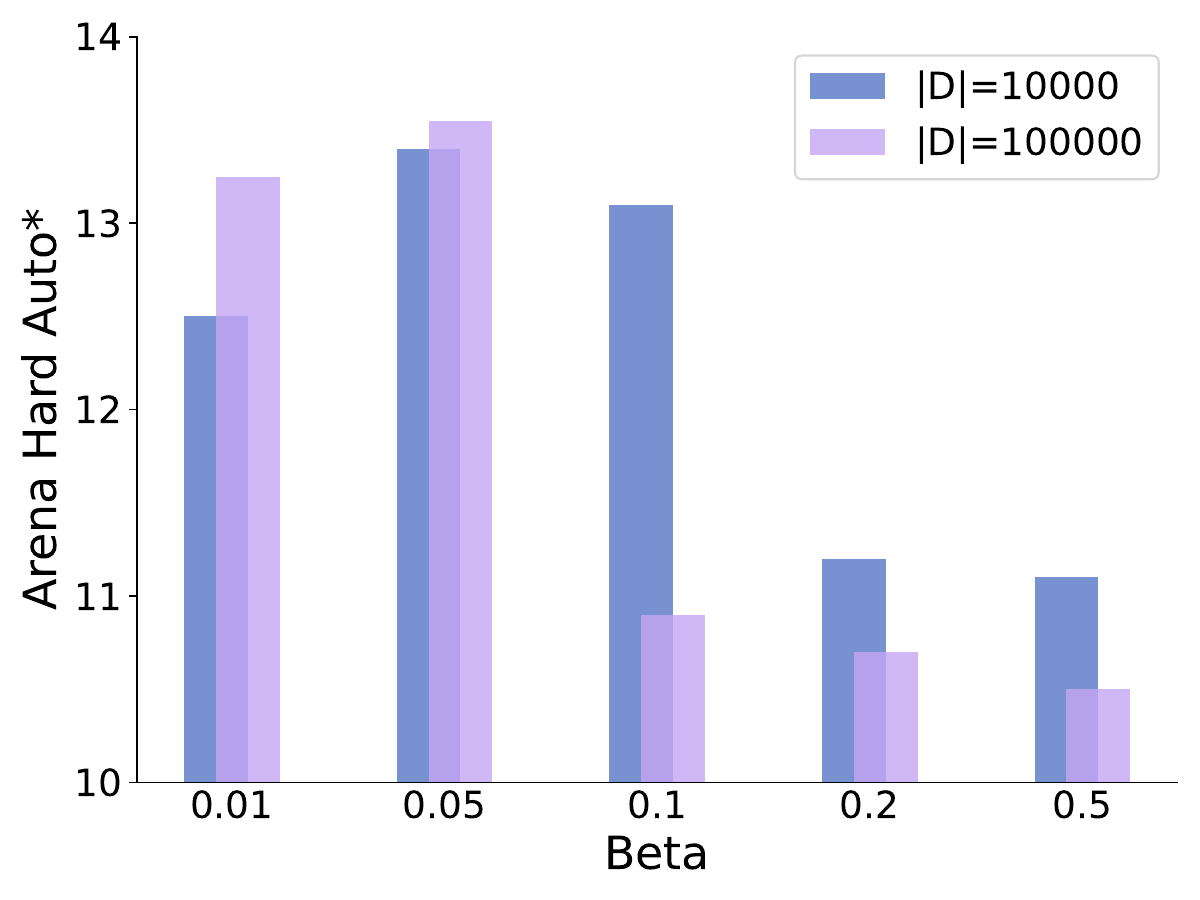}
\caption{Varying KL-divergence strength ($\beta$) under different training data sizes. We find the best $\beta$ stays relatively consistent across different dataset sizes.}
\label{fig:dpo-hyperparam-beta}
\end{figure}
\subsection{Offline Preference Learning}
\label{subsec:Offline Preference Learning}
Following prior work, we use DPO \cite{rafailov2023direct} and continue training from the last iteration of SFT from \Cref{subsec:Supervied Finetuning}.
We selected DPO for our study because it is one of the most widely used algorithms for preference optimization \cite{tunstall2023zephyr,jiang2024mixtralexperts,yang2023baichuan2openlargescale,yuan2024selfrewardinglanguagemodels}, despite the recent appearance of many alternatives \cite{meng2024simpo,gorbatovski2024trdpo,ethayarajh2024kto}.
We then compare different training settings such as choosing sequence length/reference model; tuning beta; scaling offline alignment; and filtering preference datasets.

\begin{table}[t]
\centering
  \scalebox{0.82}{
\begin{tabular}{lcc}
\toprule
\textbf{Model} & \textbf{OpenLLM} & \textbf{Arena Hard Auto*} \\
\midrule
% v0.5-sft
gemma-2b-sft & 54.67 & 8.8 \\
 + default DPO & 55.13 & 11.6\\
% gemma-1.1-2b-it & 42.1 & 71.4 & 42.3 & - & 65.4 & 17.7 & xx \\
\cmidrule(lr){1-3}
+ sqlen=2048 & 55.31 & 12.7 \\
\midrule
+ $\piref$=SFT chosen & 55.39 & \best{13.1} \\
+ $\piref$=DPO & \best{55.42} & 12.5 \\
+ $\piref$=LLaMA-3-8b & 55.31 & 12.7 \\
\bottomrule
\end{tabular}
}
\caption{Effect of training with longer sequences and using different reference models. \emph{sqlen} refers to maximum sequence length. \emph{$\piref=$x} refers to DPO training with different reference models.}
\label{tbl:seq-len-n-ref-model}
\end{table}

\paragraph{Choosing Sequence Length/Reference Model}
Popular implementations of DPO \cite{alignment_handbook2023,tunstall2023zephyr} use a sequence length of 1024, and a reference model $\piref=\pisft$. However, many recent work differs in this setting: using either a longer sequence length \cite{meng2024simpo}, or a different reference model \cite{rafailov2023direct,gorbatovski2024trdpo}. We compare these configurations by training Gemma-2b on a \emph{10k randomly sampled subset} from the mixture dataset and evaluating on OpenLLM and Arena-Hard-Auto*.
Specially, we measure the impact of 1) using a longer sequence length of 2048, and 2) using different reference models proposed by prior work.
The latter includes using $\pisft$ after further SFT on all chosen responses in the 10k subset  (denoted as \emph{$\piref$=SFT chosen}); $\pisft$ after additional DPO training on the 10k subset (denoted as \emph{$\piref$=DPO}); and using a stronger model such as LLaMA-3-8b (denoted as \emph{$\piref$=LLaMA-3-8b}).
% Many recent work considers DPO training with a sequence length of 1024 \citep{tunstall2023zephyr}, and using a reference model being the $\pisft$ itself \citep{tunstall2023zephyr,meng2024simpo}, $\pisft$ after additional SFT on the chosen responses \citep{rafailov2023direct}, or any other model with a stronger capability \citep{gorbatovski2024trdpo}. We compare these configurations by training Gemma-2b on a 10k subset from the mixture dataset and evaluating on OpenLLM and Arena-Hard-Auto*. We report the results in \Cref{tbl:seq-len-n-ref-model}.

\Cref{tbl:seq-len-n-ref-model} shows that training with a longer sequence length of 2048 significantly improves performance on both benchmarks.
We believe this is because multi-turn chat data are intrinsically long, and that longer responses may contain more complex reasoning compared to shorter answers \cite{zhao2024longismore}.
We also find using different reference models such as \emph{$\piref$=SFT chosen} or \emph{$\piref$=DPO} slightly improves performance.
However, as these methods require additional training, for simplicity we use $\piref=\pisft$ for the rest of the experiments.
% Default configuration: seq len = 2048, bsz = 128, beta=0.1, epoch=6.

\paragraph{Tuning Beta} Beta $\beta$ is a hyperparameter in DPO that controls the strength of KL-divergence.
Besides sequence length and reference model, many prior work \cite{tunstall2023zephyr,gorbatovski2024trdpo} also differs in the choice of $\beta$.
It is unclear whether $\beta$ can critically affect performance, and how other factors such as training data size can interact with $\beta$.

To investigate this, we first fix a dataset size $|D|$, and vary $\beta \in \{0.01, 0.05, 0.1, 0.2, 0.5\}$. We then repeat this process for different dataset sizes.
% For each training dataset size, we fix the number of training epochs, and vary $\beta$ from 0.01 to 0.5. 
In \Cref{fig:dpo-hyperparam-beta}, we find that 1) using a high KL-divergence $\beta$ significantly harms performance, and 2) the best $\beta$ stays relatively consistent across different training data sizes.
% This shows that besides hyperparameters such as learning rate, $\beta$ can also significantly affect performance in DPO.
This indicates that $\beta$ can be tuned using only a small subset of the data\footnote{However, we note that using a subset too small (e.g., $|D|=1000$) do not yield meaningful variations across runs.
}, which is much more compute-efficient than sweeping using the full dataset.
\begin{figure*}[t!]
    \centering
    \subfigure[OpenLLM]{
        \includegraphics[scale=0.32]{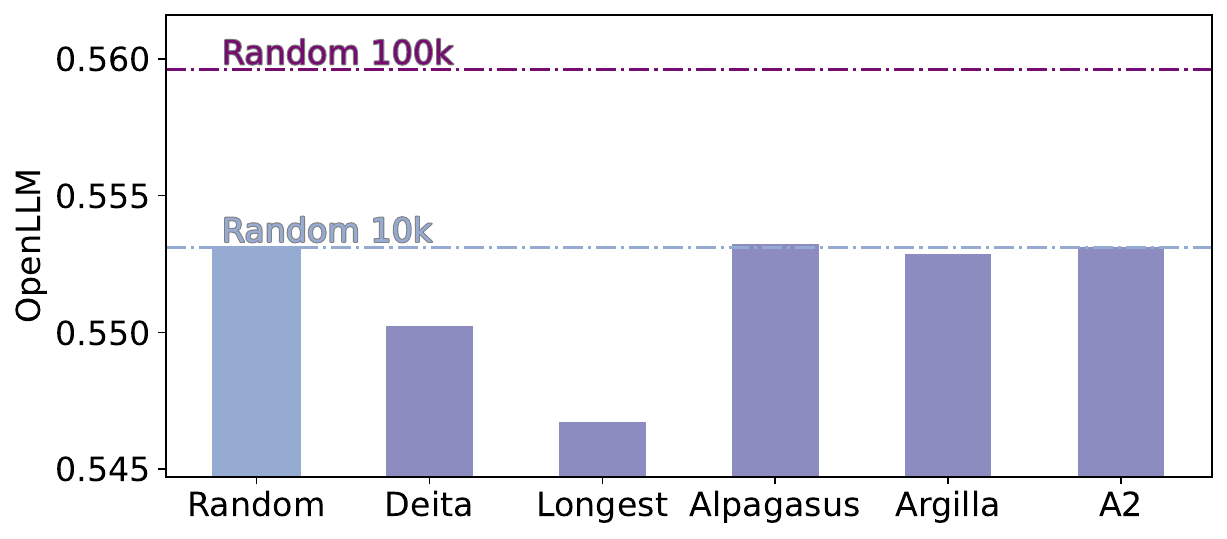}
    }
    \subfigure[Arena Hard Auto*]{
        \includegraphics[scale=0.32]{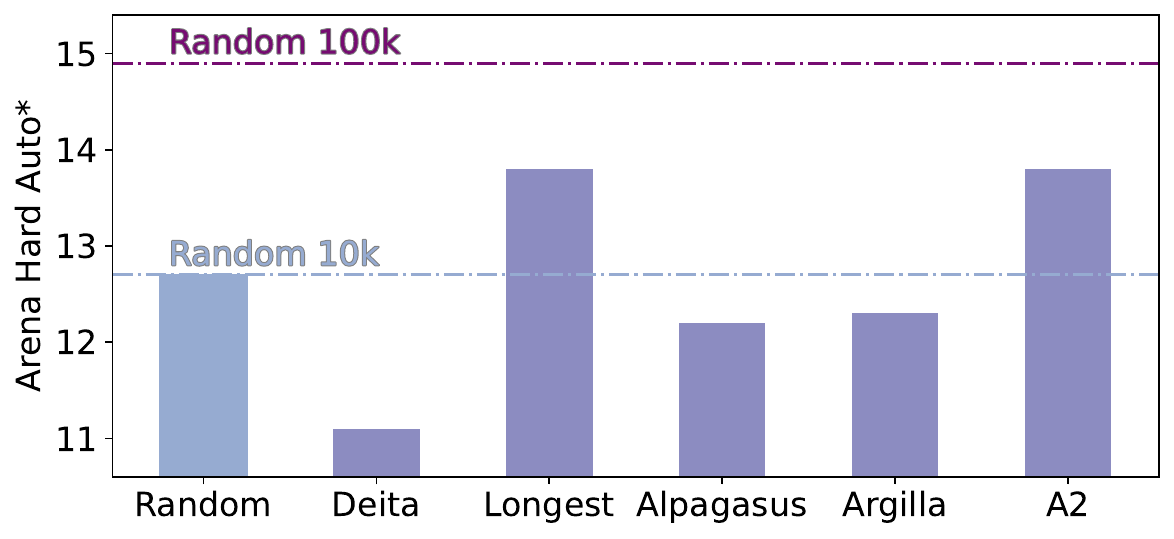}
    }
\caption{Effect of training on 10k data selected using different filtering algorithms. We find that simply training on a larger dataset (100k) outperforms all methods.}
\label{fig:dpo-dset-algo}
\end{figure*}
\paragraph{Scaling Offline Alignment} Prior work in SFT shows that scaling high-quality data during pretraining can significantly improve performance \cite{hoffmann2022compute-optimal,kaplan2020scaling-llm}. We investigate whether a similar scaling law exists in offline preference alignment. For a given dataset size $|D|$, we fix all training hyperparameters (e.g., $\beta=0.1$ and a learning rate of 5e-7) and only vary the number of training steps. We then repeat this process for $|D|=\{1,10,100\}\times 10^{3}$, all randomly sampled from the dataset in \Cref{tab:dpo_data_stats}. For other training details, please refer to \Cref{subsec:Training Hyperparams for Scaling DPO}.
We present the results in \Cref{fig:scaling-dpo}.

In \Cref{fig:scaling-dpo}, we find that 1) under a fixed dataset size, performance quickly saturates/over-optimizes as training step increases \cite{rafailov2024scaling-dpo,gao2022scaling-reward-model}; and 2) increasing dataset size raises the point of saturation.
We believe this indicates that similar to SFT, scaling law exists in offline preference learning so that scaling both dataset size and training steps can improve performance. 
We note that this finding contrasts many DPO training configurations in prior work, where either a 2-10 times smaller dataset is used \cite{tunstall2023zephyr,ivison2023camels-dpo}, or 2-4 times fewer training steps are performed \cite{meng2024simpo,gorbatovski2024trdpo}.

\paragraph{Filtering Preference Datasets} Besides increasing training data, several prior work \cite{liu2024makes-deita,zhou2023lima,zhao2024longismore} have also explored scaling \emph{down} training data. These work finds that training with a small selection of ``highest-quality'' data can match or outperform training with the full dataset.
To measure the effectiveness of these approaches, we considered training with a 10k data budget obtained from different data selection algorithms: \framework{Deita} \cite{liu2024makes-deita}, \framework{Longest} \cite{zhao2024longismore}, \framework{Alpagasus} \cite{chen2024alpagasus}, and \framework{Argilla} \cite{argilla-dpo-mix}. These methods select data based on their response length, response quality, prompt diversity, or a mixture of them\footnote{
Some algorithms such as \framework{DEITA} are originally designed for SFT datasets and uses a single prompt-response pair $(x,y)$. In these cases, we use $(x,y_w)$ from DPO datasets as $(x,y)$.
}.
We also consider \framework{A2}, our simple heuristic that filters data based on a combination of score difference \cite{argilla-dpo-mix,wang2024secrets} and high rating alike \framework{AlpaGasus}.
On a high level, our method 1) remove preference pairs that have a score difference of less than 2.0, which is similar to \citet{argilla-dpo-mix} that treats pairs with small score differences as noisy training data; and 2) bin preference pairs by their score differences, and uniformly sample preference pairs that has the highest chosen score from each bin, which is similar to \citet{chen2024alpagasus} that trains on the highest-quality sequences judged by GPT-3.5-turbo/GPT-4-turbo.
For more implementation details on these algorithms, please refer to \Cref{subsec:Dataset Filtering Algorithms}.

\Cref{fig:dpo-dset-algo} summarizes the results.
We find that 1) simply training on a 10 times larger dataset (100k) outperforms all data filtering methods; and 2) \framework{A2} is the only method that is competitive with random sampling in \emph{both} benchmarks.
We believe the former result strengthens the importance of data quantity and diversity in offline preference learning. The latter indicates that filtering methods based on attributes about the data itself may be insufficient for DPO, and that ``better'' data may be model dependent \cite{xia2024less,yu2024teaching}.
\begin{table}[t]
    \centering
      \scalebox{0.82}{
    \begin{tabular}{lcc}
    \toprule
    \textbf{Model} & \textbf{OpenLLM} & \textbf{Arena Hard Auto*} \\
    \midrule
    % v0.5-sft
    gemma-2b-sft & 54.67 & 8.8 \\
    + offline DPO (10k) & 55.31 & 12.7 \\
    + offline DPO (100k) & 55.96 & 14.9 \\
    \cmidrule(lr){1-3}
    + ODPO (1k) & 55.32 & 12.7 \\
    + ODPO (5k) & 55.31 & 13.4 \\
    + ODPO (10k) & 55.32 & 14.6 \\
    % + ODPO 60k & - & - \\
    \bottomrule
    \end{tabular}
    }
\caption{Effect of Online DPO (denoted as \emph{ODPO}). We initialize all ODPO runs from the DPO (10k) checkpoint, and investigate the effect of different training data sizes.}
\label{tbl:online-dpo}
\vspace{-5pt}
\end{table}
\subsection{Online Preference Learning}
\label{subsec:Online Preference Learning}
Finetuning $\pi_\theta$ with preference data obtained online has proven highly effective in further enhancing model performance \cite{dong2024rlhf-workflow,guo2024onlinedpo}.
Given the high computational complexity of online training \cite{schulman2017proximal,snorkel2023iterative-dpo}, we investigate whether it remains ``essential'' compared to the much more efficient offline alternative (\Cref{subsec:Offline Preference Learning}).
% Given the high computational complexity of online training \cite{schulman2017proximal}, we use a fixed reward model and investigate the impact of different training dataset sizes on the final performance.

We measure the effect of various online training data sizes on the final performance, and compare it against offline DPO.
Specifically, we follow \citet{meng2024simpo} and first sample $n=5$ responses with a temperature of $0.8$ for each prompt from the UltraFeedback dataset. We then use Pair-RM\footnote{We note that results may vary with different reward models \cite{lambert2024rewardbench}, which we leave for future work.} \citep{jiang2023llmblender} as a judge and use the best and worst response as $y_w$ and $y_l$, respectively. Similar to \citet{meng2024simpo}, we perform one iteration of online training using DPO.
We train all models from the DPO checkpoint trained with 10k randomly sampled data (denoted as \emph{offline DPO (10k)}).

In \Cref{tbl:online-dpo}, we first find that online training mainly benefits chat benchmarks (Arena Hard Auto*) but not core capability/knowledgege benchmarks (OpenLLM).
We believe this is because online preference pairs are derived from $\pi_\theta$ itself, making it unlikely for $\pi_\theta$ to acquire \emph{new} knowledge or skills.
Next, we find that increasing the number of online training samples to 10k reaches comparable performance to offline DPO with 100k data.
This indicates that online training remains competitive, and can be much more sample efficient than offline training for chat benchmarks.

% it seems that training with half as many epochs as the offline DPO training.

\section{The \framework{Lion} Series}
In the previous section, we empirically analyzed the best strategies to perform supervised fine-tuning, offline preference learning, and online preference learning.
We aggregate these findings to train a series of models, the \framework{Lion} series, and evaluate them on numerous LLM benchmarks.

\begin{table*}[!t]
\setlength{\tabcolsep}{4pt}
  \centering
  \scalebox{0.86}{
    \begin{tabular}{l lc cccc}
      \toprule
      \textbf{Model} & {\textbf{Method}} & \textbf{Size} 
      & \textbf{Arena-Hard} & \textbf{AlpacaEval-2} 
      & \textbf{MT-bench} & \textbf{OpenLLM} \\
      \midrule
      Gemma-2b        & -       & 2B & - & - & - & 46.69\\
      Gemma-2b-it     & SFT+RLHF & 2B & 3.4  & 5.44 & 5.63 & 42.75\\
      % wandb's run
      Gemma-2b-zephyr & SFT+DPO & 2B & 0.9 & 2.65 & 4.13 & 46.92\\
      LLaMA-2-7b-chat      & SFT     & 7B & 4.6  & 5.35 & 6.22 &  53.16\\
      Vicuna-7b-v1.5  & SFT     & 7B & 2.5 & 7.62 & 6.57 & 52.06\\
      % LLaMA-2-13b-chat     & SFT     &13B &   & 8.44 &  &  58.27\\
      Gemma-2b-lion-sft (ours)  & SFT & 2B &
      2.4 & 7.79 & 6.37 & 54.78 \\
      % gemma-2b-lion-v0.7-full-240k-beta0.05-epoch2-bsz64-zero1
      % Gemma-2b-lion-dpo (ours)  & S∂FT+DPO & 2B &
      % \textbf{5.4} & \textbf{7.86} & \textbf{6.74} & \textbf{55.38}\\
      % gemma-2b-lion-v0.6-full-185k-beta0.05-epoch4-from2-bsz64-zero1
      % Gemma-2b-lion-dpo (ours)  & SFT+DPO & 2B &
      % \textbf{5.2} & \textbf{7.91} & \textbf{6.61} & \textbf{56.15}\\
      % gemma-2b-lion-v0.6-full-165k-beta0.05-epoch4-from2-bsz64-zero1
      Gemma-2b-lion-dpo (ours)  & SFT+DPO & 2B &
      4.6 & 8.75 & 6.58 & 55.35 \\
      % gemma-2b-lion-v0.6-165k-2epoch-online-60k-bigmix-beta0.05-epoch1-bsz64-zero1
      Gemma-2b-lion-odpo (ours)  & SFT+DPO+ODPO & 2B &
      \best{5.0} & \best{9.57} & \textbf{6.75} & \best{55.98}\\
      \cmidrule(lr){1-7}
      LLaMA-3-8b      & -          & 8B & - & - & - & 63.05\\
      % manually evaled
      LLaMA-3-8b-it   & SFT+RS+DPO+PPO & 8B & 20.6 & 22.9 & 8.00 & 68.28 \\
      LLaMA-3-8b-lion-sft  (ours)  & SFT          & 8B & 11.3 & 17.9 & 7.58 & 68.71 \\
      LLaMA-3-8b-lion-dpo  (ours)  & SFT+DPO      & 8B & 19.1 & 21.8 & 8.12 & 71.28 \\
      % llama-8b-v0.2-dpo2-online-60k-beta0.01-epoch1-bsz128-zero3
      LLaMA-3-8b-lion-odpo (ours)  & SFT+DPO+ODPO & 8B & \best{22.0} & \best{26.8} & \best{8.19} & \best{71.41} \\
      \cmidrule(lr){1-7}
      LLaMA-3-70B-it     & SFT+RS+DPO+PPO & 70B & 41.1 & 34.4 & 8.95 & 73.96\\
      GPT-3.5-turbo-0125 & -          & -   & 24.8 & 22.7 & 8.39 & - \\
      GPT-4 Turbo        & -          & -   & 82.6 & 55.0 & 9.32 & - \\
      % GPT-4 Omni         & -          & -   & 0 & 0 & 0 & 0\\
      \bottomrule
    \end{tabular}
  }
  \caption{Evaluating the \framework{Lion} series across multiple chat and core knowledge benchmarks. We report the win rate and length-controlled win rate for Arena-Hard-Auto and AlpacaEval-2, respectively.}
  % \vspace{-5pt}
  \label{tbl:main_exp}
  \setlength{\tabcolsep}{6pt}
\end{table*}
\subsection{Experiment Setup}

% \paragraph{Training Datasets} Following prior work \cite{dong2024rlhf-workflow,meng2024simpo} and \Cref{sec:Training Procedure Analysis}, we manually pick a mixture of high-quality SFT and DPO datasets. 
% Given the scaling trends in \Cref{fig:scaling-dpo}, we additionally included large-scale datasets such as OpenHermes-2.5 \cite{OpenHermes2.5}, SlimOrca \cite{SlimOrca,longpre2023flan,mukherjee2023orca}, MetaMathQA \cite{yu2023metamath}, UltraChat \cite{ding2023enhancing}, OrcaMath \cite{mitra2024orcamath}, Magicoder-Evol-Instruct \cite{luo2023wizardcoder}, Capybara \cite{daniele2023amplify-instruct}, and Deita-10k \cite{liu2024what} for SFT and Nectar \cite{starling2023} for offline preference learning. This results in 1.95m data for SFT and 180k data for DPO. For online DPO, we follow \cite{meng2024simpo} and use prompts from UltraFeedback \cite{cui2023ultrafeedback}. We summarize the datasets used in \Cref{tbl:training_datasets}.

\paragraph{Training Recipe} We aggregate our findings from \Cref{sec:Training Procedure Analysis} into a single training recipe.
During the SFT stage, we use the packing with loss masking strategy. During the DPO stage, we 1) use a sequence length of 2048 and $\piref=\pisft$, 2) sweep for the optimal $\beta$ using a small (10k) subset, and 3) train our models over a large dataset with a compute budget equivalent to the best model in \Cref{fig:scaling-dpo}.
For online DPO, we follow \cite{meng2024simpo} and use Pair-RM \cite{jiang2023llmblender} as a judge. We perform one iteration of online DPO and train on the full 60k online preference data.
% analysis gemma best data=100k, 8 epoch, batch size=128
% equivalent compute gemma2b with data=180k: 4 epoch bsz=128 or (equiv steps) 2 epoch bsz=64
% equivalent compute llama8b with data=180k: 1 epoch bsz=128

\paragraph{Training Datasets} We use the same datasets from \Cref{subsec:Supervied Finetuning} for SFT. Given the scaling trends for offline preference learning (\Cref{subsec:Offline Preference Learning}), we additionally add Nectar \cite{starling2023}, HelpSteer \cite{wang2023helpsteer}, and PKU-SafeRLHF \cite{safe-rlhf}.
For online learning, we follow \cite{meng2024simpo} and use prompts from UltraFeedback. We summarize the datasets used in \Cref{subsec:LION Training Datasets}.

\paragraph{Evaluation Benchmarks} To holistically evaluate our models performance, we follow prior work and consider in total four benchmarks: Arena-Hard-Auto \cite{arenahard2024}, AlpacaEval-2 \cite{dubois2024length,alpaca_eval}, MT-Bench \cite{zheng2023judging}, and OpenLLM \cite{open-llm-leaderboard}. In addition to the evaluation methods used in \Cref{sec:Training Procedure Analysis}, AlpacaEval-2 uses a length-controlled (LC) metric to evaluate the model's instruction-following ability; and MT-Bench uses GPT-4 as a judge to score the model's response on a diverse set of QA tasks.
We use the standard evaluation setting for all benchmarks, such as using GPT-4-turbo as the judge model and GPT-4 as the reference for Arena-Hard-Auto (c.f. \Cref{subsec:Experiment Setup}).

\subsection{Models and Baselines}
\label{subsec:models and baselines}
\paragraph{Models} We train all of our models from Gemma-2b-base \cite{gemmateam2024gemma} and LLaMa-3-8B-base \cite{llama3modelcard}.
These models are pre-trained with trillions of tokens from the web, and highly performant for a wide range of text generation tasks \cite{open-llm-leaderboard}.
We denote our models trained after each phase as \emph{-lion-sft}, \emph{-lion-dpo}, and \emph{-lion-odpo}, representing the SFT, offline DPO, and online DPO stages, respectively.

\paragraph{Baselines}
We mainly compare our method against the officially released instruct models, Gemma-2b-it and Llama-3-8b-it.
% We compare against the offically released instruct models Gemma-2b-it and Llama-3-8b-it.
From the base model, Gemma-2b-it is first trained using SFT, and further finetuned using a novel, close-sourced RLHF algorithm \cite{gemmateam2024gemma}.
LLaMA-3-8b-it is trained using a combination of SFT, rejection sampling, PPO, and DPO \cite{llama3modelcard}.
Both models are trained using close-sourced data.
\begin{figure*}[t!]
\centering
\subfigure[undertrained]{
    \includegraphics[scale=0.26]{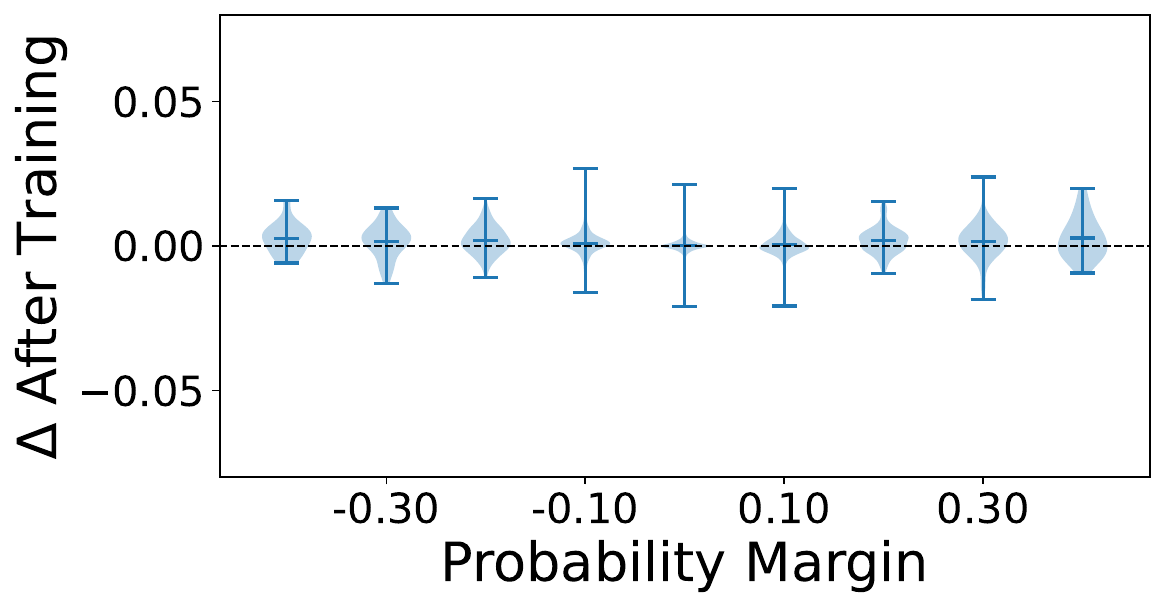}
    \label{fig:dpo-seqprob-undertrained}
}
\subfigure[best performing]{
    \includegraphics[scale=0.26]{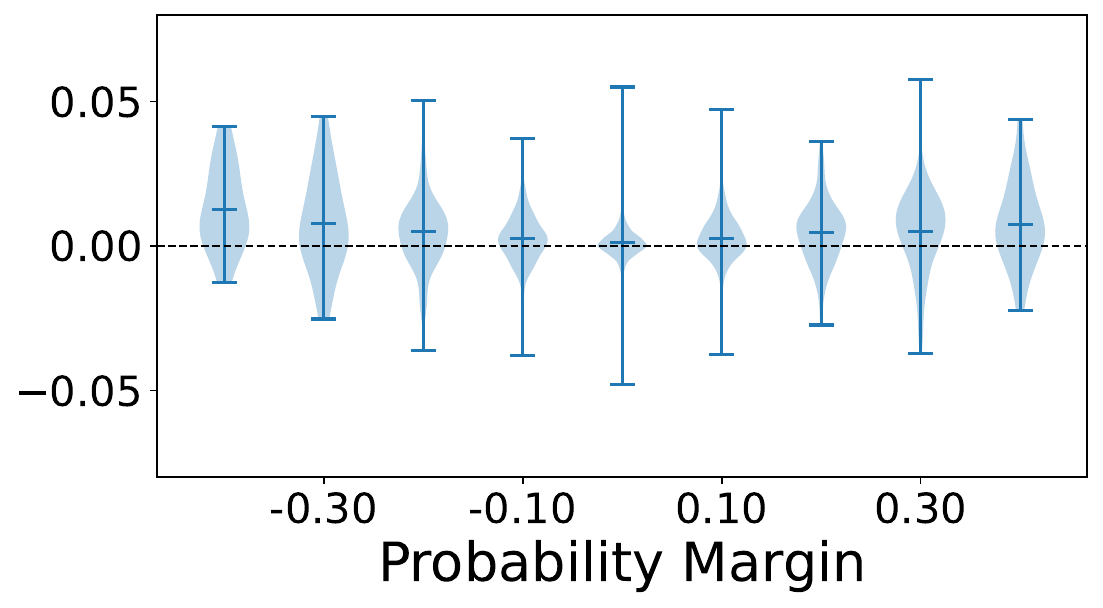}
    \label{fig:dpo-seqprob-best}
}
\subfigure[overtrained]{
    \includegraphics[scale=0.26]{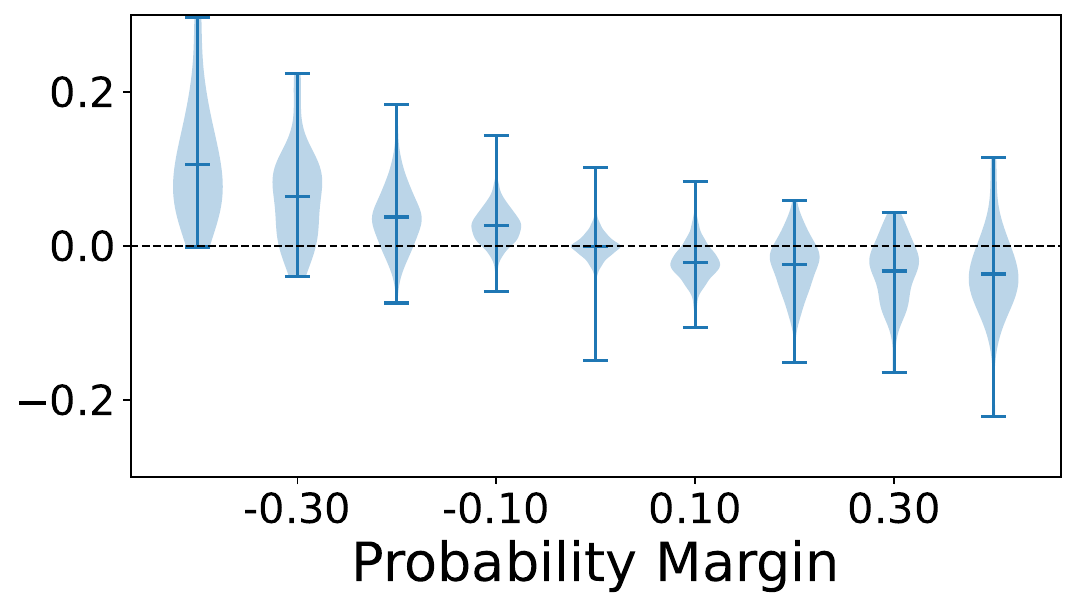}
    \label{fig:dpo-seqprob-overtrained}
}
\caption{
We track the changes in the probability margin $\pi_\theta(y_w) - \pi_\theta(y_l)$ under various training configurations, and find that the best-performing models exhibit a parabolic pattern.
Arena-Hard-Auto* results from left to right are 10.7, 14.8, and 13.2. Test loss from left to right is 0.63, 0.65, and 1.33.
% Visualizing change in sequence probability for $y_w$ and $y_l$ after DPO training. We track the changes in the probability margin $\pi_\theta(y_w) - \pi_\theta(y_l)$ under various training configurations, and find that the best-performing models exhibit a parabolic pattern, where preference pairs with the largest margin show the most improvement.
% Arena-Hard-Auto* result from left to right is 10.7, 14.8, and 13.2. Test loss from left to right is 0.63, 0.65, and 1.33.
}
\label{fig:dpo-seqprob-change-full}
\vspace{-12pt}
\end{figure*}
\subsection{Main Results}
\label{subsec:Main Results}

We present the main results in \Cref{tbl:main_exp}. We find that after the SFT and offline DPO training phase, our Gemma-2b model already outperforms the Gemma-2b-it on all benchmarks. 
It also matches or surpasses various popular 7b models, such as LLaMA-2-7b-chat \cite{touvron2023llama} and Vicuna-7b \cite{vicuna2023}. Similarly, our LLaMA-3-8b-lion-dpo shows competitive performance against the LLaMA-3-8b-it, despite only being trained with SFT and DPO. Finally, after online DPO training, our models further improve, surpassing the officially released instruct models in all benchmarks. We believe this result indicates the effectiveness of our training recipe throughout the three stages of alignment training.

\subsection{Qualitative Analysis}
\label{subsec:Qualitative Analysis}
To provide insights into the opaque training process during preference learning, we additionally measure how the sequence probability for $y_w$ and $y_l$ change after DPO training.
We use a fixed unseen test set of $(x,y_w,y_l)$ obtained from public datasets (see \Cref{subsec:DPO Analysis Dataset Selection}), and compare $\pi_\theta(y_w) - \pi_\theta(y_l)$ for each preference pair before and after training. We find that many of the best-performing models have a parabolic shape as shown in \Cref{fig:dpo-seqprob-best}.
This shows that well-trained models learns to improve confidence not only in pairs they could already distinguish correctly before training (i.e., $\pi_\theta(y_w) > \pi_\theta(y_l)$) but also equally in pairs they previously could not (i.e., $\pi_\theta(y_w) < \pi_\theta(y_l)$). Please refer to \Cref{subsec:Details on Qualitative Analysis} for more details.
% To provide insights into the opaque training process during preference learning, we additionally record how the sequence probability for $y_w$ and $y_l$ change after DPO training.
% We use a fixed unseen test set of $(x,y_w,y_l)$ obtained from public datasets (see \Cref{subsec:DPO Analysis Dataset Selection}), and compute $\pi_\theta(y_w) - \pi_\theta(y_l)$ for each preference pair before and after training.
% We then qualitatively compare between various models from \Cref{tbl:main_exp} and from \Cref{sec:Training Procedure Analysis}. We present the visualizations in \Cref{fig:dpo-seqprob-change-full}.

% In \Cref{fig:dpo-seqprob-change-full}, we find that many of the best-performing models have a parabolic shape as shown in \Cref{fig:dpo-seqprob-best}.
% This indicates that well-trained models learn to not only increase their confidence in pairs they could already distinguish correctly before training (i.e., $\pi_\theta(y_w) > \pi_\theta(y_l)$), but also improve on pairs where they previously could not distinguish (i.e., $\pi_\theta(y_w) < \pi_\theta(y_l)$).
% Undertrained models (\Cref{fig:dpo-seqprob-undertrained}) achieves a similar shape but with a much smaller magnitude.
% While overtrained models (\Cref{fig:dpo-seqprob-overtrained}) show change in probability at an even greater scale (e.g., for $\pi_\theta(y_w) < \pi_\theta(y_l)$), it is often achieved by sacrifising performance on the other side of the plot (e.g., $\pi_\theta(y_w) > \pi_\theta(y_l)$).

\section{Related Work}

Many recent studies extensively explored alignment methods to improve LLMs' ability to follow human instructions.
% Large language model alignment been effe
Early approaches train LLMs with supervised fine-tuning (SFT) using high-quality human-written demonstrations \cite{Sanh2021MultitaskPT,wei2022finetuned,chung2022scaling,Mishra2021CrossTaskGV}.
This method enjoys various properties such as fast convergence and scaling laws in training data/model sizes \cite{kaplan2020scaling-llm,hoffmann2022compute-optimal}.
% However, SFT is vulnerable to exposure bias, and thus does not align models well with nuanced human preferences.
However, SFT is vulnerable to exposure bias, and can generate outputs that do not align well with human intent.
To this end, reinforcement learning from human feedback \cite{Schulman2017ProximalPO,Stiennon2020LearningTS,Ouyang2022TrainingLM} was proposed. These work typically employs an online RL algorithm (such as PPO, \citet{schulman2017proximal}) to optimize LLMs towards a reward model mimicking human preferences.
Although effective, PPO can be complex to implement and often suffers from high reward variance, making it challenging to maintain stable performance \cite{Xu2024IsDS}.

% To address the limitations of PPO, offline preference learning algorithms like Direct Preference Optimization (DPO, \citet{rafailov2023direct}) were developed.
To address the limitations of PPO, many recent work focused on offline preference learning algorithms, such as Direct Preference Optimization \citep{rafailov2023direct}, KTO \cite{ethayarajh2024kto}, CTO \cite{xu2024contrastive}, and more \cite{guo2024direct-ipo,hong2024orpo}.
These algorithms are much easier to use, and spurred many recent studies to explore: 1) collecting high-quality offline preference data \cite{cui2023ultrafeedback,safe-rlhf,starling2023}, and 2) designing better training pipelines such as iterative DPO and online DPO \cite{guo2024onlinedpo,xu2024cringe-iterative-dpo,snorkel2023iterative-dpo,Xu2024IsDS}.
Although these contributions demonstrate improvements, the training processes, datasets, and hyper-parameters often remain heterogeneous, and it is difficult to understand the source of improvements.
Our goal is to provide a comprehensive analysis of the alignment pipeline starting from SFT to online preference learning, and to serve as a reference point for better understanding in the alignment process.

\section{Conclusion}

% Motivation
We present a detailed analysis of modern alignment pipelines, and present a step-by-step recipe to finetune models using only publicly available datasets and open-sourced algorithms.
% Findings
We find that model performance substantially improves by 1) using packing and loss making in SFT, 2) scaling 
offline preference datasets and training steps in DPO, and 3) training with online preference data to improve chat performance.
We then aggregate our findings and train the \framework{Lion} series, and show that they outperform the officially released instruct models as well as models of larger sizes on benchmarks such as Arena-Hard-Auto, AlpacaEval-2, MT-bench, and OpenLLM.
These results illustrate the importance of many previously overlooked design choices, and serve as a reference point for future work in alignment research.

% indicate that the integration of specific training strategies, particularly the use of token masking in supervised fine-tuning and the optimization of training steps and beta values in direct preference optimization, substantially improves the performance of language models.

% We introduce a robust framework for both supervised fine-tuning and preference learning—online and offline—that advances the current standards in language model training.

% % TODO, modify this
% Our findings indicate that the integration of specific training strategies, particularly the use of token masking in supervised fine-tuning and the optimization of training steps and beta values in direct preference optimization, substantially improves the performance of language models. Models such as Gemma-2b and LLama-3-8b, trained using these optimized strategies on publicly available datasets, demonstrated superior abilities in adhering to complex instructions and engaging in nuanced conversations compared to traditionally tuned models.

% % TODO: Modify this
% Moreover, we have made our models, the curated datasets, and the detailed training recipes publicly accessible. This transparency not only fosters further academic and practical validation but also encourages the community to replicate and build upon our work, ensuring continuous improvements in the field of natural language processing.

\section{Limitations}
\subsection{Sensitivity of Model Backbone}
In our analysis, we investigated the effect of various training strategies during each stage \emph{independently}, and aggregate a training recipe using the best results from each stage.
However, it is possible that there are combined effects between two or more stages (e.g., modify SFT and offline DPO simultaneously), which could lead to different or better results.
Since this would result in an exponentially larger search space for training strategies, we chose to conduct our experiments in a sequential manner. We leave this exploration for future work.
% We focus on fixing model from each stage, and investigate other stages independently. It may be possible that there are combined effects.

\subsection{Sensitivity of Training Data}
Unlike SFT, in our prior experiments we find that offline preference datasets can vary significantly in quality and quantity. We therefore manually selected a mixture of high-quality datasets for our experiments in \Cref{subsec:Offline Preference Learning} based on some empirical heuristics (\Cref{subsec:DPO Analysis Dataset Selection}).
Since this choice is empirical, we believe results may vary when, in the future, datasets of higher quality and larger sizes become available.
We believe creating new, higher-quality datasets is perpendicular to our work, and we leave this for future work.

\subsection{More Model Architectures}
\label{subsec:More Model Architectures}
Our analysis primarily focuses on the Gemma-2b-base model.
This is because 1) Gemma-2b is a light-weight yet performant model used widely in the community, and 2) it requires significantly less compute to conduct analysis as in \Cref{sec:Training Procedure Analysis} compared to using larger models such as LLaMA-3-8B-base.
However, we believe it would be beneficial to extend our analysis to other model architectures such as LLaMA-3-8b \cite{llama3modelcard} and Mistral-7b \cite{jiang2023mistral}. We plan to
extend our experiments with models of different sizes and architectures in future work.
% Could have trained on more models such as Mistral, Phi, etc.

\section{Ethical Considerations}

In this work, we focus on the reproduction study and step-by-step recipe to align large language models. Our model was trained on publicly available alignment datasets. 
Despite our efforts to carefully examine and curate our training data sources, there is a possibility that malicious or harmful content may still be present. 
To mitigate these risks, we acknowledge the necessity of incorporating more datasets specifically focused on safety, harmfulness, and bias. 
Furthermore, for future work, we commit to conducting more comprehensive evaluations on safety, harmfulness, and bias to enhance the robustness and ethical standards of language model alignment.

% \bibliography{custom}
\bibliography{iclr2024_conference,custom}

\begin{thebibliography}{80}
\providecommand{\natexlab}[1]{#1}

\bibitem[{AI@Meta(2024)}]{llama3modelcard}
AI@Meta. 2024.
\newblock \href {https://github.com/meta-llama/llama3/blob/main/MODEL_CARD.md} {Llama 3 model card}.

\bibitem[{Argilla(2024)}]{argilla-dpo-mix}
Team Argilla. 2024.
\newblock \href {https://huggingface.co/datasets/argilla/dpo-mix-7k} {Argilla: Dpo-mix-7k}.

\bibitem[{Bai et~al.(2022)Bai, Jones, Ndousse, Askell, Chen, DasSarma, Drain, Fort, Ganguli, Henighan, Joseph, Kadavath, Kernion, Conerly, El-Showk, Elhage, Hatfield-Dodds, Hernandez, Hume, Johnston, Kravec, Lovitt, Nanda, Olsson, Amodei, Brown, Clark, McCandlish, Olah, Mann, and Kaplan}]{bai2022training}
Yuntao Bai, Andy Jones, Kamal Ndousse, Amanda Askell, Anna Chen, Nova DasSarma, Dawn Drain, Stanislav Fort, Deep Ganguli, Tom Henighan, Nicholas Joseph, Saurav Kadavath, Jackson Kernion, Tom Conerly, Sheer El-Showk, Nelson Elhage, Zac Hatfield-Dodds, Danny Hernandez, Tristan Hume, Scott Johnston, Shauna Kravec, Liane Lovitt, Neel Nanda, Catherine Olsson, Dario Amodei, Tom Brown, Jack Clark, Sam McCandlish, Chris Olah, Ben Mann, and Jared Kaplan. 2022.
\newblock \href {https://arxiv.org/abs/2204.05862} {Training a helpful and harmless assistant with reinforcement learning from human feedback}.
\newblock \emph{Preprint}, arXiv:2204.05862.

\bibitem[{Banghua et~al.(2023)Banghua, Evan, Tianhao, Hanlin, and Jiantao}]{starling2023}
Zhu Banghua, Frick Evan, Wu~Tianhao, Zhu Hanlin, and Jiao Jiantao. 2023.
\newblock \href {https://starling.cs.berkeley.edu/} {Starling-7b: Improving llm helpfulness and harmlessness with rlaif}.

\bibitem[{Beeching et~al.(2023)Beeching, Fourrier, Habib, Han, Lambert, Rajani, Sanseviero, Tunstall, and Wolf}]{open-llm-leaderboard}
Edward Beeching, Clémentine Fourrier, Nathan Habib, Sheon Han, Nathan Lambert, Nazneen Rajani, Omar Sanseviero, Lewis Tunstall, and Thomas Wolf. 2023.
\newblock Open llm leaderboard.
\newblock \url{https://huggingface.co/spaces/open-llm-leaderboard/open_llm_leaderboard}.

\bibitem[{Brown et~al.(2020)Brown, Mann, Ryder, Subbiah, Kaplan, Dhariwal, Neelakantan, Shyam, Sastry, Askell, Agarwal, Herbert-Voss, Krueger, Henighan, Child, Ramesh, Ziegler, Wu, Winter, Hesse, Chen, Sigler, Litwin, Gray, Chess, Clark, Berner, McCandlish, Radford, Sutskever, and Amodei}]{brown2020language}
Tom~B. Brown, Benjamin Mann, Nick Ryder, Melanie Subbiah, Jared Kaplan, Prafulla Dhariwal, Arvind Neelakantan, Pranav Shyam, Girish Sastry, Amanda Askell, Sandhini Agarwal, Ariel Herbert-Voss, Gretchen Krueger, Tom Henighan, Rewon Child, Aditya Ramesh, Daniel~M. Ziegler, Jeffrey Wu, Clemens Winter, Christopher Hesse, Mark Chen, Eric Sigler, Mateusz Litwin, Scott Gray, Benjamin Chess, Jack Clark, Christopher Berner, Sam McCandlish, Alec Radford, Ilya Sutskever, and Dario Amodei. 2020.
\newblock \href {https://arxiv.org/abs/2005.14165} {Language models are few-shot learners}.
\newblock \emph{Preprint}, arXiv:2005.14165.

\bibitem[{Chen et~al.(2024)Chen, Li, Yan, Wang, Gunaratna, Yadav, Tang, Srinivasan, Zhou, Huang, and Jin}]{chen2024alpagasus}
Lichang Chen, Shiyang Li, Jun Yan, Hai Wang, Kalpa Gunaratna, Vikas Yadav, Zheng Tang, Vijay Srinivasan, Tianyi Zhou, Heng Huang, and Hongxia Jin. 2024.
\newblock \href {https://arxiv.org/abs/2307.08701} {Alpagasus: Training a better alpaca with fewer data}.
\newblock \emph{Preprint}, arXiv:2307.08701.

\bibitem[{Chiang et~al.(2023)Chiang, Li, Lin, Sheng, Wu, Zhang, Zheng, Zhuang, Zhuang, Gonzalez, Stoica, and Xing}]{vicuna2023}
Wei-Lin Chiang, Zhuohan Li, Zi~Lin, Ying Sheng, Zhanghao Wu, Hao Zhang, Lianmin Zheng, Siyuan Zhuang, Yonghao Zhuang, Joseph~E. Gonzalez, Ion Stoica, and Eric~P. Xing. 2023.
\newblock \href {https://lmsys.org/blog/2023-03-30-vicuna/} {Vicuna: An open-source chatbot impressing gpt-4 with 90\%* chatgpt quality}.

\bibitem[{Chung et~al.(2022)Chung, Hou, Longpre, Zoph, Tay, Fedus, Li, Wang, Dehghani, Brahma, Webson, Gu, Dai, Suzgun, Chen, Chowdhery, Castro-Ros, Pellat, Robinson, Valter, Narang, Mishra, Yu, Zhao, Huang, Dai, Yu, Petrov, Chi, Dean, Devlin, Roberts, Zhou, Le, and Wei}]{chung2022scaling}
Hyung~Won Chung, Le~Hou, Shayne Longpre, Barret Zoph, Yi~Tay, William Fedus, Yunxuan Li, Xuezhi Wang, Mostafa Dehghani, Siddhartha Brahma, Albert Webson, Shixiang~Shane Gu, Zhuyun Dai, Mirac Suzgun, Xinyun Chen, Aakanksha Chowdhery, Alex Castro-Ros, Marie Pellat, Kevin Robinson, Dasha Valter, Sharan Narang, Gaurav Mishra, Adams Yu, Vincent Zhao, Yanping Huang, Andrew Dai, Hongkun Yu, Slav Petrov, Ed~H. Chi, Jeff Dean, Jacob Devlin, Adam Roberts, Denny Zhou, Quoc~V. Le, and Jason Wei. 2022.
\newblock \href {https://arxiv.org/abs/2210.11416} {Scaling instruction-finetuned language models}.
\newblock \emph{Preprint}, arXiv:2210.11416.

\bibitem[{Cobbe et~al.(2021)Cobbe, Kosaraju, Bavarian, Chen, Jun, Kaiser, Plappert, Tworek, Hilton, Nakano, Hesse, and Schulman}]{cobbe2021gsm8k}
Karl Cobbe, Vineet Kosaraju, Mohammad Bavarian, Mark Chen, Heewoo Jun, Lukasz Kaiser, Matthias Plappert, Jerry Tworek, Jacob Hilton, Reiichiro Nakano, Christopher Hesse, and John Schulman. 2021.
\newblock Training verifiers to solve math word problems.
\newblock \emph{arXiv preprint arXiv:2110.14168}.

\bibitem[{Cui et~al.(2023)Cui, Yuan, Ding, Yao, Zhu, Ni, Xie, Liu, and Sun}]{cui2023ultrafeedback}
Ganqu Cui, Lifan Yuan, Ning Ding, Guanming Yao, Wei Zhu, Yuan Ni, Guotong Xie, Zhiyuan Liu, and Maosong Sun. 2023.
\newblock \href {https://arxiv.org/abs/2310.01377} {Ultrafeedback: Boosting language models with high-quality feedback}.
\newblock \emph{Preprint}, arXiv:2310.01377.

\bibitem[{Dai et~al.(2024)Dai, Pan, Sun, Ji, Xu, Liu, Wang, and Yang}]{safe-rlhf}
Josef Dai, Xuehai Pan, Ruiyang Sun, Jiaming Ji, Xinbo Xu, Mickel Liu, Yizhou Wang, and Yaodong Yang. 2024.
\newblock \href {https://openreview.net/forum?id=TyFrPOKYXw} {Safe rlhf: Safe reinforcement learning from human feedback}.
\newblock In \emph{The Twelfth International Conference on Learning Representations}.

\bibitem[{Daniele and Suphavadeeprasit(2023)}]{daniele2023amplify-instruct}
Luigi Daniele and Suphavadeeprasit. 2023.
\newblock \href {https://huggingface.co/datasets/LDJnr/Capybara} {Amplify-instruct: Synthetically generated diverse multi-turn conversations for efficient llm training.}
\newblock \emph{arXiv preprint arXiv:(coming soon)}.

\bibitem[{Dao(2024)}]{dao2023flashattention2}
Tri Dao. 2024.
\newblock Flash{A}ttention-2: Faster attention with better parallelism and work partitioning.
\newblock In \emph{International Conference on Learning Representations (ICLR)}.

\bibitem[{Deshpande et~al.(2023)Deshpande, Murahari, Rajpurohit, Kalyan, and Narasimhan}]{deshpande2023toxicity}
Ameet Deshpande, Vishvak Murahari, Tanmay Rajpurohit, Ashwin Kalyan, and Karthik Narasimhan. 2023.
\newblock \href {https://arxiv.org/abs/2304.05335} {Toxicity in chatgpt: Analyzing persona-assigned language models}.
\newblock \emph{Preprint}, arXiv:2304.05335.

\bibitem[{Ding et~al.(2023)Ding, Chen, Xu, Qin, Zheng, Hu, Liu, Sun, and Zhou}]{ding2023enhancing}
Ning Ding, Yulin Chen, Bokai Xu, Yujia Qin, Zhi Zheng, Shengding Hu, Zhiyuan Liu, Maosong Sun, and Bowen Zhou. 2023.
\newblock Enhancing chat language models by scaling high-quality instructional conversations.
\newblock \emph{arXiv preprint arXiv:2305.14233}.

\bibitem[{Dong et~al.(2024)Dong, Xiong, Pang, Wang, Zhao, Zhou, Jiang, Sahoo, Xiong, and Zhang}]{dong2024rlhf-workflow}
Hanze Dong, Wei Xiong, Bo~Pang, Haoxiang Wang, Han Zhao, Yingbo Zhou, Nan Jiang, Doyen Sahoo, Caiming Xiong, and Tong Zhang. 2024.
\newblock \href {https://arxiv.org/abs/2405.07863} {Rlhf workflow: From reward modeling to online rlhf}.
\newblock \emph{Preprint}, arXiv:2405.07863.

\bibitem[{Dubois et~al.(2024)Dubois, Galambosi, Liang, and Hashimoto}]{dubois2024length}
Yann Dubois, Bal{\'a}zs Galambosi, Percy Liang, and Tatsunori~B Hashimoto. 2024.
\newblock Length-controlled alpacaeval: A simple way to debias automatic evaluators.
\newblock \emph{arXiv preprint arXiv:2404.04475}.

\bibitem[{Ethayarajh et~al.(2024)Ethayarajh, Xu, Muennighoff, Jurafsky, and Kiela}]{ethayarajh2024kto}
Kawin Ethayarajh, Winnie Xu, Niklas Muennighoff, Dan Jurafsky, and Douwe Kiela. 2024.
\newblock \href {https://arxiv.org/abs/2402.01306} {Kto: Model alignment as prospect theoretic optimization}.
\newblock \emph{Preprint}, arXiv:2402.01306.

\bibitem[{Gao et~al.(2022)Gao, Schulman, and Hilton}]{gao2022scaling-reward-model}
Leo Gao, John Schulman, and Jacob Hilton. 2022.
\newblock \href {https://arxiv.org/abs/2210.10760} {Scaling laws for reward model overoptimization}.
\newblock \emph{Preprint}, arXiv:2210.10760.

\bibitem[{Gemma et~al.(2024)Gemma, Mesnard, Hardin, Dadashi, Bhupatiraju, Pathak, Sifre, Rivière, Kale, Love, Tafti, Hussenot, Sessa, Chowdhery, Roberts, Barua, Botev, Castro-Ros, Slone, Héliou, Tacchetti, Bulanova, Paterson, Tsai, Shahriari, Lan, Choquette-Choo, Crepy, Cer, Ippolito, Reid, Buchatskaya, Ni, Noland, Yan, Tucker, Muraru, Rozhdestvenskiy, Michalewski, Tenney, Grishchenko, Austin, Keeling, Labanowski, Lespiau, Stanway, Brennan, Chen, Ferret, Chiu, Mao-Jones, Lee, Yu, Millican, Sjoesund, Lee, Dixon, Reid, Mikuła, Wirth, Sharman, Chinaev, Thain, Bachem, Chang, Wahltinez, Bailey, Michel, Yotov, Chaabouni, Comanescu, Jana, Anil, McIlroy, Liu, Mullins, Smith, Borgeaud, Girgin, Douglas, Pandya, Shakeri, De, Klimenko, Hennigan, Feinberg, Stokowiec, hui Chen, Ahmed, Gong, Warkentin, Peran, Giang, Farabet, Vinyals, Dean, Kavukcuoglu, Hassabis, Ghahramani, Eck, Barral, Pereira, Collins, Joulin, Fiedel, Senter, Andreev, and Kenealy}]{gemmateam2024gemma}
Team Gemma, Thomas Mesnard, Cassidy Hardin, Robert Dadashi, Surya Bhupatiraju, Shreya Pathak, Laurent Sifre, Morgane Rivière, Mihir~Sanjay Kale, Juliette Love, Pouya Tafti, Léonard Hussenot, Pier~Giuseppe Sessa, Aakanksha Chowdhery, Adam Roberts, Aditya Barua, Alex Botev, Alex Castro-Ros, Ambrose Slone, Amélie Héliou, Andrea Tacchetti, Anna Bulanova, Antonia Paterson, Beth Tsai, Bobak Shahriari, Charline~Le Lan, Christopher~A. Choquette-Choo, Clément Crepy, Daniel Cer, Daphne Ippolito, David Reid, Elena Buchatskaya, Eric Ni, Eric Noland, Geng Yan, George Tucker, George-Christian Muraru, Grigory Rozhdestvenskiy, Henryk Michalewski, Ian Tenney, Ivan Grishchenko, Jacob Austin, James Keeling, Jane Labanowski, Jean-Baptiste Lespiau, Jeff Stanway, Jenny Brennan, Jeremy Chen, Johan Ferret, Justin Chiu, Justin Mao-Jones, Katherine Lee, Kathy Yu, Katie Millican, Lars~Lowe Sjoesund, Lisa Lee, Lucas Dixon, Machel Reid, Maciej Mikuła, Mateo Wirth, Michael Sharman, Nikolai Chinaev, Nithum Thain, Olivier Bachem,
  Oscar Chang, Oscar Wahltinez, Paige Bailey, Paul Michel, Petko Yotov, Rahma Chaabouni, Ramona Comanescu, Reena Jana, Rohan Anil, Ross McIlroy, Ruibo Liu, Ryan Mullins, Samuel~L Smith, Sebastian Borgeaud, Sertan Girgin, Sholto Douglas, Shree Pandya, Siamak Shakeri, Soham De, Ted Klimenko, Tom Hennigan, Vlad Feinberg, Wojciech Stokowiec, Yu~hui Chen, Zafarali Ahmed, Zhitao Gong, Tris Warkentin, Ludovic Peran, Minh Giang, Clément Farabet, Oriol Vinyals, Jeff Dean, Koray Kavukcuoglu, Demis Hassabis, Zoubin Ghahramani, Douglas Eck, Joelle Barral, Fernando Pereira, Eli Collins, Armand Joulin, Noah Fiedel, Evan Senter, Alek Andreev, and Kathleen Kenealy. 2024.
\newblock \href {https://arxiv.org/abs/2403.08295} {Gemma: Open models based on gemini research and technology}.
\newblock \emph{Preprint}, arXiv:2403.08295.

\bibitem[{Gorbatovski et~al.(2024)Gorbatovski, Shaposhnikov, Malakhov, Surnachev, Aksenov, Maksimov, Balagansky, and Gavrilov}]{gorbatovski2024trdpo}
Alexey Gorbatovski, Boris Shaposhnikov, Alexey Malakhov, Nikita Surnachev, Yaroslav Aksenov, Ian Maksimov, Nikita Balagansky, and Daniil Gavrilov. 2024.
\newblock \href {https://arxiv.org/abs/2404.09656} {Learn your reference model for real good alignment}.
\newblock \emph{Preprint}, arXiv:2404.09656.

\bibitem[{Guo et~al.(2024{\natexlab{a}})Guo, Zhang, Liu, Liu, Khalman, Llinares, Rame, Mesnard, Zhao, Piot, Ferret, and Blondel}]{guo2024direct-ipo}
Shangmin Guo, Biao Zhang, Tianlin Liu, Tianqi Liu, Misha Khalman, Felipe Llinares, Alexandre Rame, Thomas Mesnard, Yao Zhao, Bilal Piot, Johan Ferret, and Mathieu Blondel. 2024{\natexlab{a}}.
\newblock \href {https://arxiv.org/abs/2402.04792} {Direct language model alignment from online ai feedback}.
\newblock \emph{Preprint}, arXiv:2402.04792.

\bibitem[{Guo et~al.(2024{\natexlab{b}})Guo, Zhang, Liu, Liu, Khalman, Llinares, Rame, Mesnard, Zhao, Piot, Ferret, and Blondel}]{guo2024onlinedpo}
Shangmin Guo, Biao Zhang, Tianlin Liu, Tianqi Liu, Misha Khalman, Felipe Llinares, Alexandre Rame, Thomas Mesnard, Yao Zhao, Bilal Piot, Johan Ferret, and Mathieu Blondel. 2024{\natexlab{b}}.
\newblock \href {https://arxiv.org/abs/2402.04792} {Direct language model alignment from online ai feedback}.
\newblock \emph{Preprint}, arXiv:2402.04792.

\bibitem[{Hendrycks et~al.(2021)Hendrycks, Burns, Kadavath, Arora, Basart, Tang, Song, and Steinhardt}]{hendrycksmath2021}
Dan Hendrycks, Collin Burns, Saurav Kadavath, Akul Arora, Steven Basart, Eric Tang, Dawn Song, and Jacob Steinhardt. 2021.
\newblock Measuring mathematical problem solving with the math dataset.
\newblock \emph{NeurIPS}.

\bibitem[{Hoffmann et~al.(2022)Hoffmann, Borgeaud, Mensch, Buchatskaya, Cai, Rutherford, de~Las~Casas, Hendricks, Welbl, Clark, Hennigan, Noland, Millican, van~den Driessche, Damoc, Guy, Osindero, Simonyan, Elsen, Rae, Vinyals, and Sifre}]{hoffmann2022compute-optimal}
Jordan Hoffmann, Sebastian Borgeaud, Arthur Mensch, Elena Buchatskaya, Trevor Cai, Eliza Rutherford, Diego de~Las~Casas, Lisa~Anne Hendricks, Johannes Welbl, Aidan Clark, Tom Hennigan, Eric Noland, Katie Millican, George van~den Driessche, Bogdan Damoc, Aurelia Guy, Simon Osindero, Karen Simonyan, Erich Elsen, Jack~W. Rae, Oriol Vinyals, and Laurent Sifre. 2022.
\newblock \href {https://arxiv.org/abs/2203.15556} {Training compute-optimal large language models}.
\newblock \emph{Preprint}, arXiv:2203.15556.

\bibitem[{Hong et~al.(2024)Hong, Lee, and Thorne}]{hong2024orpo}
Jiwoo Hong, Noah Lee, and James Thorne. 2024.
\newblock \href {https://arxiv.org/abs/2403.07691} {Orpo: Monolithic preference optimization without reference model}.
\newblock \emph{Preprint}, arXiv:2403.07691.

\bibitem[{Ivison et~al.(2023)Ivison, Wang, Pyatkin, Lambert, Peters, Dasigi, Jang, Wadden, Smith, Beltagy, and Hajishirzi}]{ivison2023camels-dpo}
Hamish Ivison, Yizhong Wang, Valentina Pyatkin, Nathan Lambert, Matthew Peters, Pradeep Dasigi, Joel Jang, David Wadden, Noah~A. Smith, Iz~Beltagy, and Hannaneh Hajishirzi. 2023.
\newblock \href {https://arxiv.org/abs/2311.10702} {Camels in a changing climate: Enhancing lm adaptation with tulu 2}.
\newblock \emph{Preprint}, arXiv:2311.10702.

\bibitem[{Jiang et~al.(2023{\natexlab{a}})Jiang, Sablayrolles, Mensch, Bamford, Chaplot, de~las Casas, Bressand, Lengyel, Lample, Saulnier, Lavaud, Lachaux, Stock, Scao, Lavril, Wang, Lacroix, and Sayed}]{jiang2023mistral}
Albert~Q. Jiang, Alexandre Sablayrolles, Arthur Mensch, Chris Bamford, Devendra~Singh Chaplot, Diego de~las Casas, Florian Bressand, Gianna Lengyel, Guillaume Lample, Lucile Saulnier, Lélio~Renard Lavaud, Marie-Anne Lachaux, Pierre Stock, Teven~Le Scao, Thibaut Lavril, Thomas Wang, Timothée Lacroix, and William~El Sayed. 2023{\natexlab{a}}.
\newblock \href {https://arxiv.org/abs/2310.06825} {Mistral 7b}.
\newblock \emph{Preprint}, arXiv:2310.06825.

\bibitem[{Jiang et~al.(2024)Jiang, Sablayrolles, Roux, Mensch, Savary, Bamford, Chaplot, de~las Casas, Hanna, Bressand, Lengyel, Bour, Lample, Lavaud, Saulnier, Lachaux, Stock, Subramanian, Yang, Antoniak, Scao, Gervet, Lavril, Wang, Lacroix, and Sayed}]{jiang2024mixtralexperts}
Albert~Q. Jiang, Alexandre Sablayrolles, Antoine Roux, Arthur Mensch, Blanche Savary, Chris Bamford, Devendra~Singh Chaplot, Diego de~las Casas, Emma~Bou Hanna, Florian Bressand, Gianna Lengyel, Guillaume Bour, Guillaume Lample, Lélio~Renard Lavaud, Lucile Saulnier, Marie-Anne Lachaux, Pierre Stock, Sandeep Subramanian, Sophia Yang, Szymon Antoniak, Teven~Le Scao, Théophile Gervet, Thibaut Lavril, Thomas Wang, Timothée Lacroix, and William~El Sayed. 2024.
\newblock \href {https://arxiv.org/abs/2401.04088} {Mixtral of experts}.
\newblock \emph{Preprint}, arXiv:2401.04088.

\bibitem[{Jiang et~al.(2023{\natexlab{b}})Jiang, Ren, and Lin}]{jiang2023llmblender}
Dongfu Jiang, Xiang Ren, and Bill~Yuchen Lin. 2023{\natexlab{b}}.
\newblock \href {https://arxiv.org/abs/2306.02561} {Llm-blender: Ensembling large language models with pairwise ranking and generative fusion}.
\newblock \emph{Preprint}, arXiv:2306.02561.

\bibitem[{Kaplan et~al.(2020)Kaplan, McCandlish, Henighan, Brown, Chess, Child, Gray, Radford, Wu, and Amodei}]{kaplan2020scaling-llm}
Jared Kaplan, Sam McCandlish, Tom Henighan, Tom~B. Brown, Benjamin Chess, Rewon Child, Scott Gray, Alec Radford, Jeffrey Wu, and Dario Amodei. 2020.
\newblock \href {https://arxiv.org/abs/2001.08361} {Scaling laws for neural language models}.
\newblock \emph{Preprint}, arXiv:2001.08361.

\bibitem[{Lambert et~al.(2024)Lambert, Pyatkin, Morrison, Miranda, Lin, Chandu, Dziri, Kumar, Zick, Choi, Smith, and Hajishirzi}]{lambert2024rewardbench}
Nathan Lambert, Valentina Pyatkin, Jacob Morrison, LJ~Miranda, Bill~Yuchen Lin, Khyathi Chandu, Nouha Dziri, Sachin Kumar, Tom Zick, Yejin Choi, Noah~A. Smith, and Hannaneh Hajishirzi. 2024.
\newblock \href {https://arxiv.org/abs/2403.13787} {Rewardbench: Evaluating reward models for language modeling}.
\newblock \emph{Preprint}, arXiv:2403.13787.

\bibitem[{Li et~al.(2023{\natexlab{a}})Li, Hammoud, Itani, Khizbullin, and Ghanem}]{li2023camel}
Guohao Li, Hasan Abed Al~Kader Hammoud, Hani Itani, Dmitrii Khizbullin, and Bernard Ghanem. 2023{\natexlab{a}}.
\newblock \href {https://arxiv.org/abs/2303.17760} {Camel: Communicative agents for "mind" exploration of large scale language model society}.
\newblock \emph{Preprint}, arXiv:2303.17760.

\bibitem[{Li et~al.(2024)Li, Chiang, Frick, Dunlap, Zhu, Gonzalez, and Stoica}]{arenahard2024}
Tianle Li, Wei-Lin Chiang, Evan Frick, Lisa Dunlap, Banghua Zhu, Joseph~E. Gonzalez, and Ion Stoica. 2024.
\newblock \href {https://lmsys.org/blog/2024-04-19-arena-hard/} {From live data to high-quality benchmarks: The arena-hard pipeline}.

\bibitem[{Li et~al.(2023{\natexlab{b}})Li, Zhang, Dubois, Taori, Gulrajani, Guestrin, Liang, and Hashimoto}]{alpaca_eval}
Xuechen Li, Tianyi Zhang, Yann Dubois, Rohan Taori, Ishaan Gulrajani, Carlos Guestrin, Percy Liang, and Tatsunori~B. Hashimoto. 2023{\natexlab{b}}.
\newblock Alpacaeval: An automatic evaluator of instruction-following models.
\newblock \url{https://github.com/tatsu-lab/alpaca_eval}.

\bibitem[{Lian et~al.(2023)Lian, Wang, Goodson, Pentland, Cook, Vong, and "Teknium"}]{SlimOrca}
Wing Lian, Guan Wang, Bleys Goodson, Eugene Pentland, Austin Cook, Chanvichet Vong, and "Teknium". 2023.
\newblock \href {https://https://huggingface.co/Open-Orca/SlimOrca} {Slimorca: An open dataset of gpt-4 augmented flan reasoning traces, with verification}.

\bibitem[{Liu et~al.(2024{\natexlab{a}})Liu, Zeng, He, Jiang, and He}]{liu2024what}
Wei Liu, Weihao Zeng, Keqing He, Yong Jiang, and Junxian He. 2024{\natexlab{a}}.
\newblock \href {https://openreview.net/forum?id=BTKAeLqLMw} {What makes good data for alignment? a comprehensive study of automatic data selection in instruction tuning}.
\newblock In \emph{The Twelfth International Conference on Learning Representations}.

\bibitem[{Liu et~al.(2024{\natexlab{b}})Liu, Zeng, He, Jiang, and He}]{liu2024makes-deita}
Wei Liu, Weihao Zeng, Keqing He, Yong Jiang, and Junxian He. 2024{\natexlab{b}}.
\newblock \href {https://arxiv.org/abs/2312.15685} {What makes good data for alignment? a comprehensive study of automatic data selection in instruction tuning}.
\newblock \emph{Preprint}, arXiv:2312.15685.

\bibitem[{Longpre et~al.(2023)Longpre, Hou, Vu, Webson, Chung, Tay, Zhou, Le, Zoph, Wei, and Roberts}]{longpre2023flan}
Shayne Longpre, Le~Hou, Tu~Vu, Albert Webson, Hyung~Won Chung, Yi~Tay, Denny Zhou, Quoc~V. Le, Barret Zoph, Jason Wei, and Adam Roberts. 2023.
\newblock \href {https://arxiv.org/abs/2301.13688} {The flan collection: Designing data and methods for effective instruction tuning}.
\newblock \emph{Preprint}, arXiv:2301.13688.

\bibitem[{Luo et~al.(2023)Luo, Xu, Zhao, Sun, Geng, Hu, Tao, Ma, Lin, and Jiang}]{luo2023wizardcoder}
Ziyang Luo, Can Xu, Pu~Zhao, Qingfeng Sun, Xiubo Geng, Wenxiang Hu, Chongyang Tao, Jing Ma, Qingwei Lin, and Daxin Jiang. 2023.
\newblock Wizardcoder: Empowering code large language models with evol-instruct.

\bibitem[{Meng et~al.(2024)Meng, Xia, and Chen}]{meng2024simpo}
Yu~Meng, Mengzhou Xia, and Danqi Chen. 2024.
\newblock \href {https://arxiv.org/abs/2405.14734} {Simpo: Simple preference optimization with a reference-free reward}.
\newblock \emph{Preprint}, arXiv:2405.14734.

\bibitem[{Mishra et~al.(2021)Mishra, Khashabi, Baral, and Hajishirzi}]{Mishra2021CrossTaskGV}
Swaroop Mishra, Daniel Khashabi, Chitta Baral, and Hannaneh Hajishirzi. 2021.
\newblock \href {https://api.semanticscholar.org/CorpusID:237421373} {Cross-task generalization via natural language crowdsourcing instructions}.
\newblock In \emph{Annual Meeting of the Association for Computational Linguistics}.

\bibitem[{Mitra et~al.(2024)Mitra, Khanpour, Rosset, and Awadallah}]{mitra2024orcamath}
Arindam Mitra, Hamed Khanpour, Corby Rosset, and Ahmed Awadallah. 2024.
\newblock \href {https://arxiv.org/abs/2402.14830} {Orca-math: Unlocking the potential of slms in grade school math}.
\newblock \emph{Preprint}, arXiv:2402.14830.

\bibitem[{Mukherjee et~al.(2023)Mukherjee, Mitra, Jawahar, Agarwal, Palangi, and Awadallah}]{mukherjee2023orca}
Subhabrata Mukherjee, Arindam Mitra, Ganesh Jawahar, Sahaj Agarwal, Hamid Palangi, and Ahmed Awadallah. 2023.
\newblock \href {https://arxiv.org/abs/2306.02707} {Orca: Progressive learning from complex explanation traces of gpt-4}.
\newblock \emph{Preprint}, arXiv:2306.02707.

\bibitem[{OpenAI et~al.(2024)OpenAI, Achiam, Adler, Agarwal, Ahmad, Akkaya, Aleman, Almeida, Altenschmidt, Altman, Anadkat, Avila, Babuschkin, Balaji, Balcom, Baltescu, Bao, and et~al.}]{openai2024gpt4}
OpenAI, Josh Achiam, Steven Adler, Sandhini Agarwal, Lama Ahmad, Ilge Akkaya, Florencia~Leoni Aleman, Diogo Almeida, Janko Altenschmidt, Sam Altman, Shyamal Anadkat, Red Avila, Igor Babuschkin, Suchir Balaji, Valerie Balcom, Paul Baltescu, Haiming Bao, and et~al. 2024.
\newblock \href {https://arxiv.org/abs/2303.08774} {Gpt-4 technical report}.
\newblock \emph{Preprint}, arXiv:2303.08774.

\bibitem[{Ouyang et~al.(2022{\natexlab{a}})Ouyang, Wu, Jiang, Almeida, Wainwright, Mishkin, Zhang, Agarwal, Slama, Ray, Schulman, Hilton, Kelton, Miller, Simens, Askell, Welinder, Christiano, Leike, and Lowe}]{ouyang2022training}
Long Ouyang, Jeff Wu, Xu~Jiang, Diogo Almeida, Carroll~L. Wainwright, Pamela Mishkin, Chong Zhang, Sandhini Agarwal, Katarina Slama, Alex Ray, John Schulman, Jacob Hilton, Fraser Kelton, Luke Miller, Maddie Simens, Amanda Askell, Peter Welinder, Paul Christiano, Jan Leike, and Ryan Lowe. 2022{\natexlab{a}}.
\newblock \href {https://arxiv.org/abs/2203.02155} {Training language models to follow instructions with human feedback}.
\newblock \emph{Preprint}, arXiv:2203.02155.

\bibitem[{Ouyang et~al.(2022{\natexlab{b}})Ouyang, Wu, Jiang, Almeida, Wainwright, Mishkin, Zhang, Agarwal, Slama, Ray, Schulman, Hilton, Kelton, Miller, Simens, Askell, Welinder, Christiano, Leike, and Lowe}]{Ouyang2022TrainingLM}
Long Ouyang, Jeff Wu, Xu~Jiang, Diogo Almeida, Carroll~L. Wainwright, Pamela Mishkin, Chong Zhang, Sandhini Agarwal, Katarina Slama, Alex Ray, John Schulman, Jacob Hilton, Fraser Kelton, Luke~E. Miller, Maddie Simens, Amanda Askell, Peter Welinder, Paul~Francis Christiano, Jan Leike, and Ryan~J. Lowe. 2022{\natexlab{b}}.
\newblock \href {https://api.semanticscholar.org/CorpusID:246426909} {Training language models to follow instructions with human feedback}.
\newblock \emph{ArXiv}, abs/2203.02155.

\bibitem[{Rafailov et~al.(2024)Rafailov, Chittepu, Park, Sikchi, Hejna, Knox, Finn, and Niekum}]{rafailov2024scaling-dpo}
Rafael Rafailov, Yaswanth Chittepu, Ryan Park, Harshit Sikchi, Joey Hejna, Bradley Knox, Chelsea Finn, and Scott Niekum. 2024.
\newblock \href {https://arxiv.org/abs/2406.02900} {Scaling laws for reward model overoptimization in direct alignment algorithms}.
\newblock \emph{Preprint}, arXiv:2406.02900.

\bibitem[{Rafailov et~al.(2023)Rafailov, Sharma, Mitchell, Manning, Ermon, and Finn}]{rafailov2023direct}
Rafael Rafailov, Archit Sharma, Eric Mitchell, Christopher~D Manning, Stefano Ermon, and Chelsea Finn. 2023.
\newblock \href {https://openreview.net/forum?id=HPuSIXJaa9} {Direct preference optimization: Your language model is secretly a reward model}.
\newblock In \emph{Thirty-seventh Conference on Neural Information Processing Systems}.

\bibitem[{Rasley et~al.(2020)Rasley, Rajbhandari, Ruwase, and He}]{Rasley2020DeepSpeedSO}
Jeff Rasley, Samyam Rajbhandari, Olatunji Ruwase, and Yuxiong He. 2020.
\newblock \href {https://api.semanticscholar.org/CorpusID:221191193} {Deepspeed: System optimizations enable training deep learning models with over 100 billion parameters}.
\newblock \emph{Proceedings of the 26th ACM SIGKDD International Conference on Knowledge Discovery \& Data Mining}.

\bibitem[{Sanh et~al.(2021)Sanh, Webson, Raffel, Bach, Sutawika, Alyafeai, Chaffin, Stiegler, Scao, Raja, Dey, Bari, Xu, Thakker, Sharma, Szczechla, Kim, Chhablani, Nayak, Datta, Chang, Jiang, Wang, Manica, Shen, Yong, Pandey, Bawden, Wang, Neeraj, Rozen, Sharma, Santilli, F{\'e}vry, Fries, Teehan, Biderman, Gao, Bers, Wolf, and Rush}]{Sanh2021MultitaskPT}
Victor Sanh, Albert Webson, Colin Raffel, Stephen~H. Bach, Lintang Sutawika, Zaid Alyafeai, Antoine Chaffin, Arnaud Stiegler, Teven~Le Scao, Arun Raja, Manan Dey, M~Saiful Bari, Canwen Xu, Urmish Thakker, Shanya Sharma, Eliza Szczechla, Taewoon Kim, Gunjan Chhablani, Nihal~V. Nayak, Debajyoti Datta, Jonathan~D. Chang, Mike Tian-Jian Jiang, Han Wang, Matteo Manica, Sheng Shen, Zheng-Xin Yong, Harshit Pandey, Rachel Bawden, Thomas Wang, Trishala Neeraj, Jos Rozen, Abheesht Sharma, Andrea Santilli, Thibault F{\'e}vry, Jason~Alan Fries, Ryan Teehan, Stella Biderman, Leo Gao, Tali Bers, Thomas Wolf, and Alexander~M. Rush. 2021.
\newblock \href {https://api.semanticscholar.org/CorpusID:239009562} {Multitask prompted training enables zero-shot task generalization}.
\newblock \emph{ArXiv}, abs/2110.08207.

\bibitem[{Schulman et~al.(2017{\natexlab{a}})Schulman, Wolski, Dhariwal, Radford, and Klimov}]{schulman2017proximal}
John Schulman, Filip Wolski, Prafulla Dhariwal, Alec Radford, and Oleg Klimov. 2017{\natexlab{a}}.
\newblock \href {https://arxiv.org/abs/1707.06347} {Proximal policy optimization algorithms}.
\newblock \emph{Preprint}, arXiv:1707.06347.

\bibitem[{Schulman et~al.(2017{\natexlab{b}})Schulman, Wolski, Dhariwal, Radford, and Klimov}]{Schulman2017ProximalPO}
John Schulman, Filip Wolski, Prafulla Dhariwal, Alec Radford, and Oleg Klimov. 2017{\natexlab{b}}.
\newblock \href {https://api.semanticscholar.org/CorpusID:28695052} {Proximal policy optimization algorithms}.
\newblock \emph{ArXiv}, abs/1707.06347.

\bibitem[{Shi et~al.(2024)Shi, Yang, Wu, Aitchison, Yilmaz, and Lipani}]{shi2024instruction}
Zhengyan Shi, Adam~X. Yang, Bin Wu, Laurence Aitchison, Emine Yilmaz, and Aldo Lipani. 2024.
\newblock \href {https://arxiv.org/abs/2405.14394} {Instruction tuning with loss over instructions}.
\newblock \emph{Preprint}, arXiv:2405.14394.

\bibitem[{Stiennon et~al.(2020{\natexlab{a}})Stiennon, Ouyang, Wu, Ziegler, Lowe, Voss, Radford, Amodei, and Christiano}]{stienon2020learning}
Nisan Stiennon, Long Ouyang, Jeff Wu, Daniel~M. Ziegler, Ryan Lowe, Chelsea Voss, Alec Radford, Dario Amodei, and Paul Christiano. 2020{\natexlab{a}}.
\newblock Learning to summarize from human feedback.
\newblock In \emph{NeurIPS}.

\bibitem[{Stiennon et~al.(2020{\natexlab{b}})Stiennon, Ouyang, Wu, Ziegler, Lowe, Voss, Radford, Amodei, and Christiano}]{Stiennon2020LearningTS}
Nisan Stiennon, Long Ouyang, Jeff Wu, Daniel~M. Ziegler, Ryan~J. Lowe, Chelsea Voss, Alec Radford, Dario Amodei, and Paul Christiano. 2020{\natexlab{b}}.
\newblock \href {https://api.semanticscholar.org/CorpusID:221665105} {Learning to summarize from human feedback}.
\newblock \emph{ArXiv}, abs/2009.01325.

\bibitem[{Taori et~al.(2023)Taori, Gulrajani, Zhang, Dubois, Li, Guestrin, Liang, and Hashimoto}]{alpaca}
Rohan Taori, Ishaan Gulrajani, Tianyi Zhang, Yann Dubois, Xuechen Li, Carlos Guestrin, Percy Liang, and Tatsunori~B. Hashimoto. 2023.
\newblock Stanford alpaca: An instruction-following llama model.
\newblock \url{https://github.com/tatsu-lab/stanford_alpaca}.

\bibitem[{Teknium(2023)}]{OpenHermes2.5}
Teknium. 2023.
\newblock \href {https://huggingface.co/datasets/teknium/OpenHermes-2.5} {Openhermes 2.5: An open dataset of synthetic data for generalist llm assistants}.

\bibitem[{Touvron et~al.(2023)Touvron, Martin, Stone, Albert, Almahairi, Babaei, Bashlykov, Batra, Bhargava, Bhosale, Bikel, Blecher, Ferrer, Chen, Cucurull, Esiobu, Fernandes, Fu, Fu, Fuller, Gao, Goswami, Goyal, Hartshorn, Hosseini, Hou, Inan, Kardas, Kerkez, Khabsa, Kloumann, Korenev, Koura, Lachaux, Lavril, Lee, Liskovich, Lu, Mao, Martinet, Mihaylov, Mishra, Molybog, Nie, Poulton, Reizenstein, Rungta, Saladi, Schelten, Silva, Smith, Subramanian, Tan, Tang, Taylor, Williams, Kuan, Xu, Yan, Zarov, Zhang, Fan, Kambadur, Narang, Rodriguez, Stojnic, Edunov, and Scialom}]{touvron2023llama}
Hugo Touvron, Louis Martin, Kevin Stone, Peter Albert, Amjad Almahairi, Yasmine Babaei, Nikolay Bashlykov, Soumya Batra, Prajjwal Bhargava, Shruti Bhosale, Dan Bikel, Lukas Blecher, Cristian~Canton Ferrer, Moya Chen, Guillem Cucurull, David Esiobu, Jude Fernandes, Jeremy Fu, Wenyin Fu, Brian Fuller, Cynthia Gao, Vedanuj Goswami, Naman Goyal, Anthony Hartshorn, Saghar Hosseini, Rui Hou, Hakan Inan, Marcin Kardas, Viktor Kerkez, Madian Khabsa, Isabel Kloumann, Artem Korenev, Punit~Singh Koura, Marie-Anne Lachaux, Thibaut Lavril, Jenya Lee, Diana Liskovich, Yinghai Lu, Yuning Mao, Xavier Martinet, Todor Mihaylov, Pushkar Mishra, Igor Molybog, Yixin Nie, Andrew Poulton, Jeremy Reizenstein, Rashi Rungta, Kalyan Saladi, Alan Schelten, Ruan Silva, Eric~Michael Smith, Ranjan Subramanian, Xiaoqing~Ellen Tan, Binh Tang, Ross Taylor, Adina Williams, Jian~Xiang Kuan, Puxin Xu, Zheng Yan, Iliyan Zarov, Yuchen Zhang, Angela Fan, Melanie Kambadur, Sharan Narang, Aurelien Rodriguez, Robert Stojnic, Sergey Edunov, and Thomas
  Scialom. 2023.
\newblock \href {https://arxiv.org/abs/2307.09288} {Llama 2: Open foundation and fine-tuned chat models}.
\newblock \emph{Preprint}, arXiv:2307.09288.

\bibitem[{Tran et~al.(2023)Tran, Glaze, and Hancock}]{snorkel2023iterative-dpo}
Hoang Tran, Chris Glaze, and Braden Hancock. 2023.
\newblock Iterative dpo alignment.

\bibitem[{Tunstall et~al.(2023{\natexlab{a}})Tunstall, Beeching, Lambert, Rajani, Huang, Rasul, Rush, and Wolf}]{alignment_handbook2023}
Lewis Tunstall, Edward Beeching, Nathan Lambert, Nazneen Rajani, Shengyi Huang, Kashif Rasul, Alexander~M. Rush, and Thomas Wolf. 2023{\natexlab{a}}.
\newblock The alignment handbook.
\newblock \url{https://github.com/huggingface/alignment-handbook}.

\bibitem[{Tunstall et~al.(2023{\natexlab{b}})Tunstall, Beeching, Lambert, Rajani, Rasul, Belkada, Huang, von Werra, Fourrier, Habib, Sarrazin, Sanseviero, Rush, and Wolf}]{tunstall2023zephyr}
Lewis Tunstall, Edward Beeching, Nathan Lambert, Nazneen Rajani, Kashif Rasul, Younes Belkada, Shengyi Huang, Leandro von Werra, Clémentine Fourrier, Nathan Habib, Nathan Sarrazin, Omar Sanseviero, Alexander~M. Rush, and Thomas Wolf. 2023{\natexlab{b}}.
\newblock \href {https://arxiv.org/abs/2310.16944} {Zephyr: Direct distillation of lm alignment}.
\newblock \emph{Preprint}, arXiv:2310.16944.

\bibitem[{Wang et~al.(2024)Wang, Zheng, Chen, Liu, Dou, Huang, Shen, Jin, Zhou, Shi, Gao, Xu, Zhou, Fan, Xi, Zhao, Wang, Ji, Yan, Shen, Chen, Gui, Zhang, Qiu, Huang, Wu, and Jiang}]{wang2024secrets}
Binghai Wang, Rui Zheng, Lu~Chen, Yan Liu, Shihan Dou, Caishuang Huang, Wei Shen, Senjie Jin, Enyu Zhou, Chenyu Shi, Songyang Gao, Nuo Xu, Yuhao Zhou, Xiaoran Fan, Zhiheng Xi, Jun Zhao, Xiao Wang, Tao Ji, Hang Yan, Lixing Shen, Zhan Chen, Tao Gui, Qi~Zhang, Xipeng Qiu, Xuanjing Huang, Zuxuan Wu, and Yu-Gang Jiang. 2024.
\newblock \href {https://arxiv.org/abs/2401.06080} {Secrets of rlhf in large language models part ii: Reward modeling}.
\newblock \emph{Preprint}, arXiv:2401.06080.

\bibitem[{Wang et~al.(2023)Wang, Dong, Zeng, Adams, Sreedhar, Egert, Delalleau, Scowcroft, Kant, Swope, and Kuchaiev}]{wang2023helpsteer}
Zhilin Wang, Yi~Dong, Jiaqi Zeng, Virginia Adams, Makesh~Narsimhan Sreedhar, Daniel Egert, Olivier Delalleau, Jane~Polak Scowcroft, Neel Kant, Aidan Swope, and Oleksii Kuchaiev. 2023.
\newblock \href {https://arxiv.org/abs/2311.09528} {Helpsteer: Multi-attribute helpfulness dataset for steerlm}.
\newblock \emph{Preprint}, arXiv:2311.09528.

\bibitem[{Wei et~al.(2023)Wei, Haghtalab, and Steinhardt}]{wei2023jailbroken}
Alexander Wei, Nika Haghtalab, and Jacob Steinhardt. 2023.
\newblock \href {https://arxiv.org/abs/2307.02483} {Jailbroken: How does llm safety training fail?}
\newblock \emph{Preprint}, arXiv:2307.02483.

\bibitem[{Wei et~al.(2022)Wei, Bosma, Zhao, Guu, Yu, Lester, Du, Dai, and Le}]{wei2022finetuned}
Jason Wei, Maarten Bosma, Vincent~Y. Zhao, Kelvin Guu, Adams~Wei Yu, Brian Lester, Nan Du, Andrew~M. Dai, and Quoc~V. Le. 2022.
\newblock \href {https://arxiv.org/abs/2109.01652} {Finetuned language models are zero-shot learners}.
\newblock \emph{Preprint}, arXiv:2109.01652.

\bibitem[{Xia et~al.(2024)Xia, Malladi, Gururangan, Arora, and Chen}]{xia2024less}
Mengzhou Xia, Sadhika Malladi, Suchin Gururangan, Sanjeev Arora, and Danqi Chen. 2024.
\newblock \href {https://arxiv.org/abs/2402.04333} {Less: Selecting influential data for targeted instruction tuning}.
\newblock \emph{Preprint}, arXiv:2402.04333.

\bibitem[{Xu et~al.(2023)Xu, Sun, Zheng, Geng, Zhao, Feng, Tao, and Jiang}]{Xu2023WizardLMEL}
Can Xu, Qingfeng Sun, Kai Zheng, Xiubo Geng, Pu~Zhao, Jiazhan Feng, Chongyang Tao, and Daxin Jiang. 2023.
\newblock \href {https://api.semanticscholar.org/CorpusID:258298159} {Wizardlm: Empowering large language models to follow complex instructions}.
\newblock \emph{ArXiv}, abs/2304.12244.

\bibitem[{Xu et~al.(2024{\natexlab{a}})Xu, Sharaf, Chen, Tan, Shen, Durme, Murray, and Kim}]{xu2024contrastive}
Haoran Xu, Amr Sharaf, Yunmo Chen, Weiting Tan, Lingfeng Shen, Benjamin~Van Durme, Kenton Murray, and Young~Jin Kim. 2024{\natexlab{a}}.
\newblock \href {https://arxiv.org/abs/2401.08417} {Contrastive preference optimization: Pushing the boundaries of llm performance in machine translation}.
\newblock \emph{Preprint}, arXiv:2401.08417.

\bibitem[{Xu et~al.(2024{\natexlab{b}})Xu, Lee, Sukhbaatar, and Weston}]{xu2024cringe-iterative-dpo}
Jing Xu, Andrew Lee, Sainbayar Sukhbaatar, and Jason Weston. 2024{\natexlab{b}}.
\newblock \href {https://arxiv.org/abs/2312.16682} {Some things are more cringe than others: Iterative preference optimization with the pairwise cringe loss}.
\newblock \emph{Preprint}, arXiv:2312.16682.

\bibitem[{Xu et~al.(2024{\natexlab{c}})Xu, Fu, Gao, Ye, Liu, Mei, Wang, Yu, and Wu}]{Xu2024IsDS}
Shusheng Xu, Wei Fu, Jiaxuan Gao, Wenjie Ye, Weiling Liu, Zhiyu Mei, Guangju Wang, Chao Yu, and Yi~Wu. 2024{\natexlab{c}}.
\newblock \href {https://api.semanticscholar.org/CorpusID:269157140} {Is dpo superior to ppo for llm alignment? a comprehensive study}.
\newblock \emph{ArXiv}, abs/2404.10719.

\bibitem[{Yang et~al.(2023)Yang, Xiao, Wang, Zhang, Bian, Yin, Lv, Pan, Wang, Yan, Yang, Deng, Wang, Liu, Ai, Dong, Zhao, Xu, Sun, Zhang, Liu, Ji, Xie, Dai, Fang, Su, Song, Liu, Ru, Ma, Wang, Liu, Lin, Nie, Guo, Sun, Zhang, Li, Li, Cheng, Chen, Zeng, Wang, Chen, Men, Yu, Pan, Shen, Wang, Li, Jiang, Gao, Zhang, Zhou, and Wu}]{yang2023baichuan2openlargescale}
Aiyuan Yang, Bin Xiao, Bingning Wang, Borong Zhang, Ce~Bian, Chao Yin, Chenxu Lv, Da~Pan, Dian Wang, Dong Yan, Fan Yang, Fei Deng, Feng Wang, Feng Liu, Guangwei Ai, Guosheng Dong, Haizhou Zhao, Hang Xu, Haoze Sun, Hongda Zhang, Hui Liu, Jiaming Ji, Jian Xie, JunTao Dai, Kun Fang, Lei Su, Liang Song, Lifeng Liu, Liyun Ru, Luyao Ma, Mang Wang, Mickel Liu, MingAn Lin, Nuolan Nie, Peidong Guo, Ruiyang Sun, Tao Zhang, Tianpeng Li, Tianyu Li, Wei Cheng, Weipeng Chen, Xiangrong Zeng, Xiaochuan Wang, Xiaoxi Chen, Xin Men, Xin Yu, Xuehai Pan, Yanjun Shen, Yiding Wang, Yiyu Li, Youxin Jiang, Yuchen Gao, Yupeng Zhang, Zenan Zhou, and Zhiying Wu. 2023.
\newblock \href {https://arxiv.org/abs/2309.10305} {Baichuan 2: Open large-scale language models}.
\newblock \emph{Preprint}, arXiv:2309.10305.

\bibitem[{Yu et~al.(2023)Yu, Jiang, Shi, Yu, Liu, Zhang, Kwok, Li, Weller, and Liu}]{yu2023metamath}
Longhui Yu, Weisen Jiang, Han Shi, Jincheng Yu, Zhengying Liu, Yu~Zhang, James~T Kwok, Zhenguo Li, Adrian Weller, and Weiyang Liu. 2023.
\newblock Metamath: Bootstrap your own mathematical questions for large language models.
\newblock \emph{arXiv preprint arXiv:2309.12284}.

\bibitem[{Yu et~al.(2024)Yu, Peng, Galley, Gao, and Yu}]{yu2024teaching}
Xiao Yu, Baolin Peng, Michel Galley, Jianfeng Gao, and Zhou Yu. 2024.
\newblock \href {https://arxiv.org/abs/2310.13522} {Teaching language models to self-improve through interactive demonstrations}.
\newblock \emph{Preprint}, arXiv:2310.13522.

\bibitem[{Yuan et~al.(2024)Yuan, Pang, Cho, Li, Sukhbaatar, Xu, and Weston}]{yuan2024selfrewardinglanguagemodels}
Weizhe Yuan, Richard~Yuanzhe Pang, Kyunghyun Cho, Xian Li, Sainbayar Sukhbaatar, Jing Xu, and Jason Weston. 2024.
\newblock \href {https://arxiv.org/abs/2401.10020} {Self-rewarding language models}.
\newblock \emph{Preprint}, arXiv:2401.10020.

\bibitem[{Zhao et~al.(2024)Zhao, Andriushchenko, Croce, and Flammarion}]{zhao2024longismore}
Hao Zhao, Maksym Andriushchenko, Francesco Croce, and Nicolas Flammarion. 2024.
\newblock \href {https://arxiv.org/abs/2402.04833} {Long is more for alignment: A simple but tough-to-beat baseline for instruction fine-tuning}.
\newblock \emph{Preprint}, arXiv:2402.04833.

\bibitem[{Zheng et~al.(2023)Zheng, Chiang, Sheng, Zhuang, Wu, Zhuang, Lin, Li, Li, Xing, Zhang, Gonzalez, and Stoica}]{zheng2023judging}
Lianmin Zheng, Wei-Lin Chiang, Ying Sheng, Siyuan Zhuang, Zhanghao Wu, Yonghao Zhuang, Zi~Lin, Zhuohan Li, Dacheng Li, Eric~P. Xing, Hao Zhang, Joseph~E. Gonzalez, and Ion Stoica. 2023.
\newblock \href {https://arxiv.org/abs/2306.05685} {Judging llm-as-a-judge with mt-bench and chatbot arena}.
\newblock \emph{Preprint}, arXiv:2306.05685.

\bibitem[{Zhou et~al.(2023)Zhou, Liu, Xu, Iyer, Sun, Mao, Ma, Efrat, Yu, Yu, Zhang, Ghosh, Lewis, Zettlemoyer, and Levy}]{zhou2023lima}
Chunting Zhou, Pengfei Liu, Puxin Xu, Srini Iyer, Jiao Sun, Yuning Mao, Xuezhe Ma, Avia Efrat, Ping Yu, Lili Yu, Susan Zhang, Gargi Ghosh, Mike Lewis, Luke Zettlemoyer, and Omer Levy. 2023.
\newblock \href {https://arxiv.org/abs/2305.11206} {Lima: Less is more for alignment}.
\newblock \emph{Preprint}, arXiv:2305.11206.

\bibitem[{Ziegler et~al.(2020)Ziegler, Stiennon, Wu, Brown, Radford, Amodei, Christiano, and Irving}]{ziegler2020finetuning}
Daniel~M. Ziegler, Nisan Stiennon, Jeffrey Wu, Tom~B. Brown, Alec Radford, Dario Amodei, Paul Christiano, and Geoffrey Irving. 2020.
\newblock \href {https://arxiv.org/abs/1909.08593} {Fine-tuning language models from human preferences}.
\newblock \emph{Preprint}, arXiv:1909.08593.

\end{thebibliography}

\clearpage
\appendix
\setcounter{table}{0}
\renewcommand{\thetable}{A\arabic{table}}
\setcounter{figure}{0}
\renewcommand{\thefigure}{A\arabic{figure}}

\section{More Details on Alignment Procedure Analysis}
\label{sec:More Details on Analyzing Offline Preference Learning}

% \subsection{Qingyang's OpenLLM}

\subsection{Offline Preference Dataset Curation}
\label{subsec:DPO Analysis Dataset Selection}
Gathering a high-quality dataset of sufficient scale is imperative to study various properties of current offline preference learning algorithms.
% To study various properties of current offline preference learning algorithms, gathering a high-quality dataset of sufficiently large scale is imperative. 
There are many open-source preference labeled datasets available online, including Ultrafeedback \cite{cui2023ultrafeedback}, TLDR-Preferences \cite{stienon2020learning}, and Nectar \cite{starling2023}.
However, they show significant differences in 1) the quality and diversity of the prompts, 2) models used to generate the responses, 3) judge models, and 4) the number of preference pairs.
It is therefore unclear which dataset is of sufficient quality to train a model on, and how the model's performance changes on downstream benchmarks.
% At the time of writing, there are many open-source perference labeled datasets available online, including Ultrafeedback being ... as well as OpenHermes Preference, PRM pairs, etc being created by individuals. These datasets altogether amounts to over 1 million trainable preference pairs, however it is unclear if many of them are of sufficient quality.

\begin{figure}[t]
    \centering
    \includegraphics[scale=0.42]{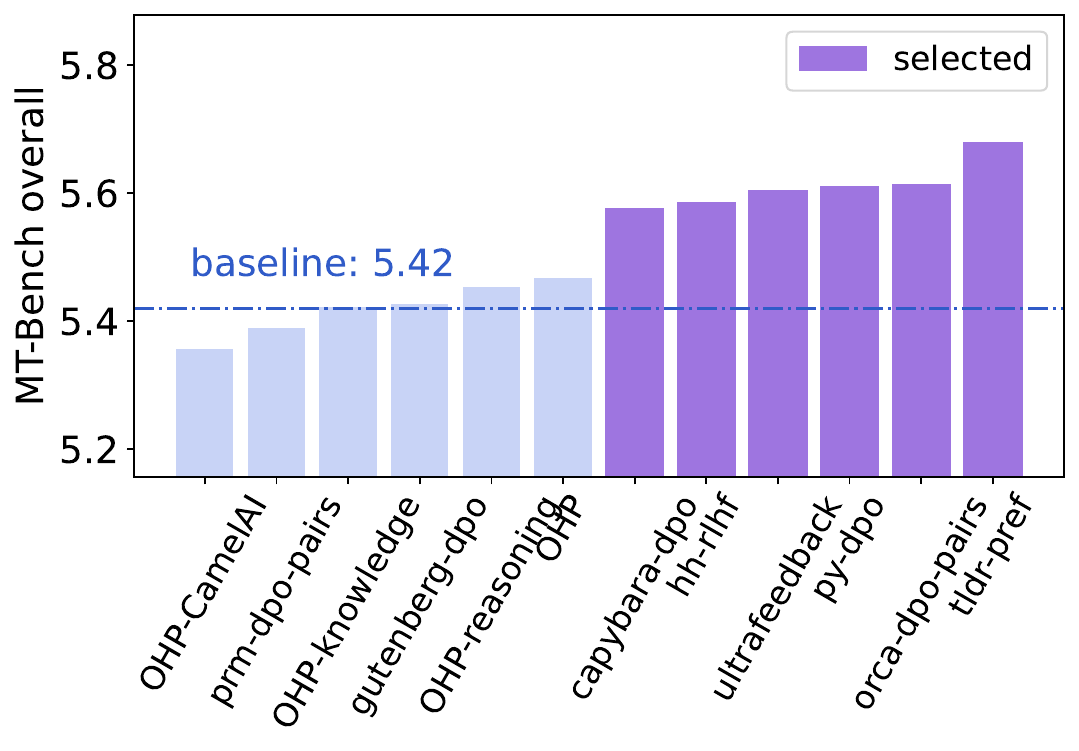}
    \caption{MT-bench score after training Gemma-2b on upto 10k samples from each dataset. Datasets we used in \Cref{sec:Training Procedure Analysis} is colored in purple. ``OHP'' stands for Openhermes-Preference.}
    \label{fig:data-eval-mt-bench}
\end{figure}

\begin{figure*}[t!]
\centering
\subfigure[Coding]{
    \includegraphics[scale=0.34]{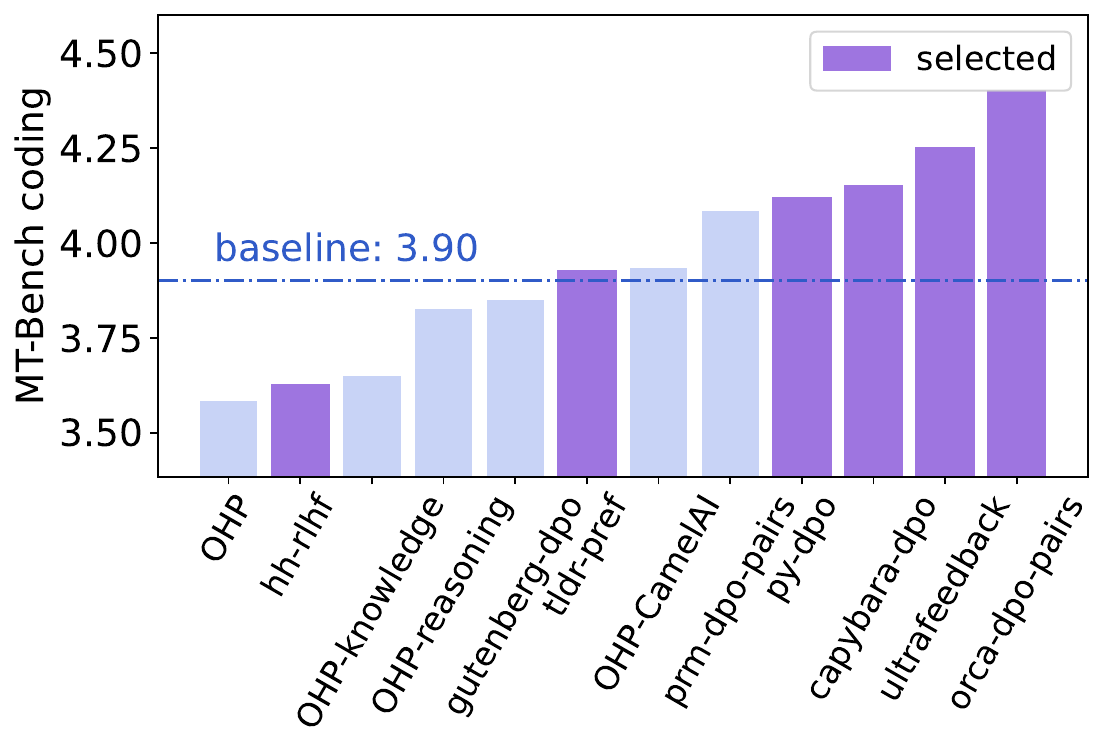}
}
\qquad
\subfigure[Extraction]{
    \includegraphics[scale=0.34]{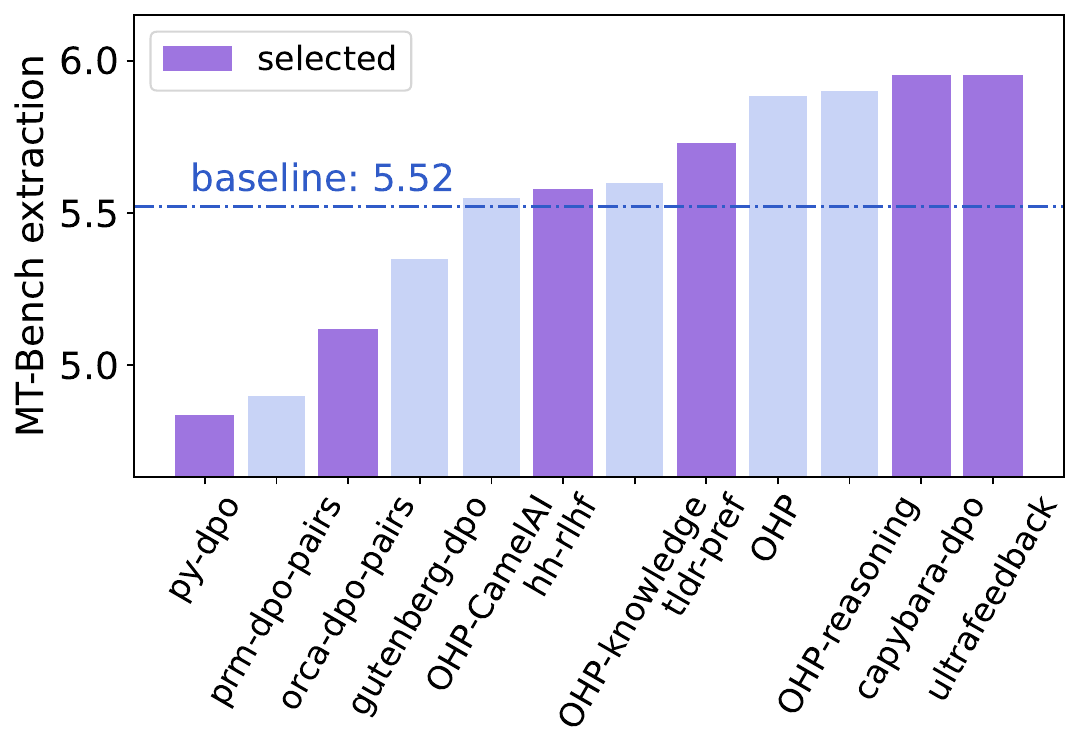}
}
\subfigure[Humanities]{
    \includegraphics[scale=0.34]{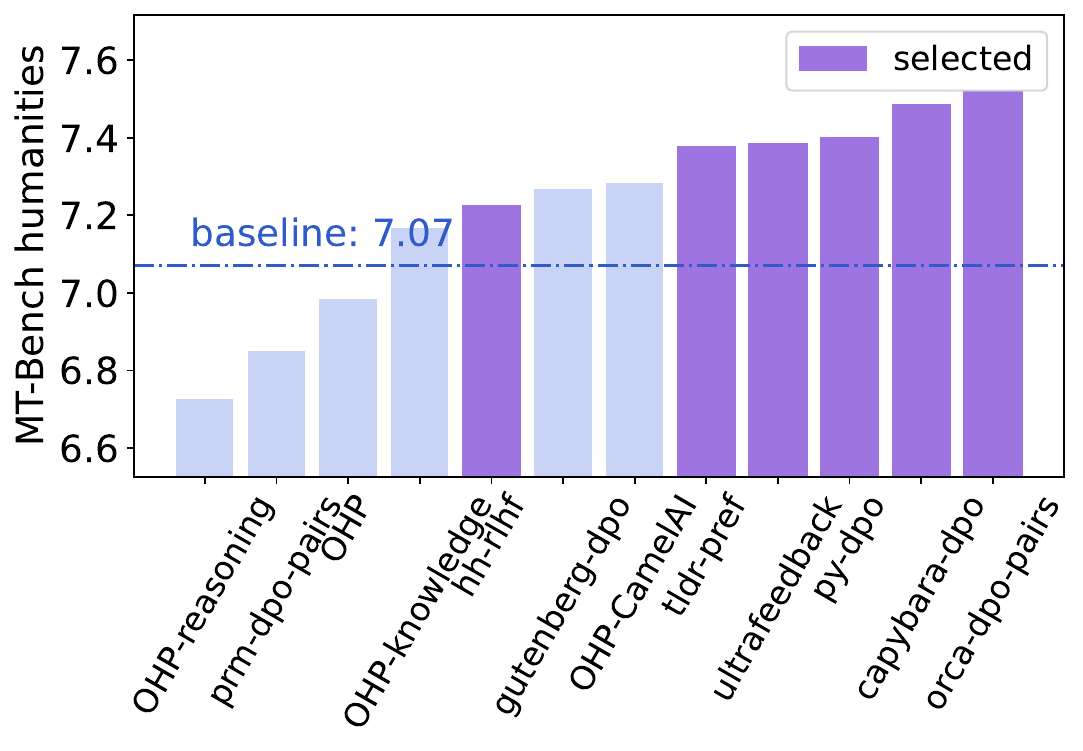}
}
\qquad
\subfigure[Math]{
    \includegraphics[scale=0.34]{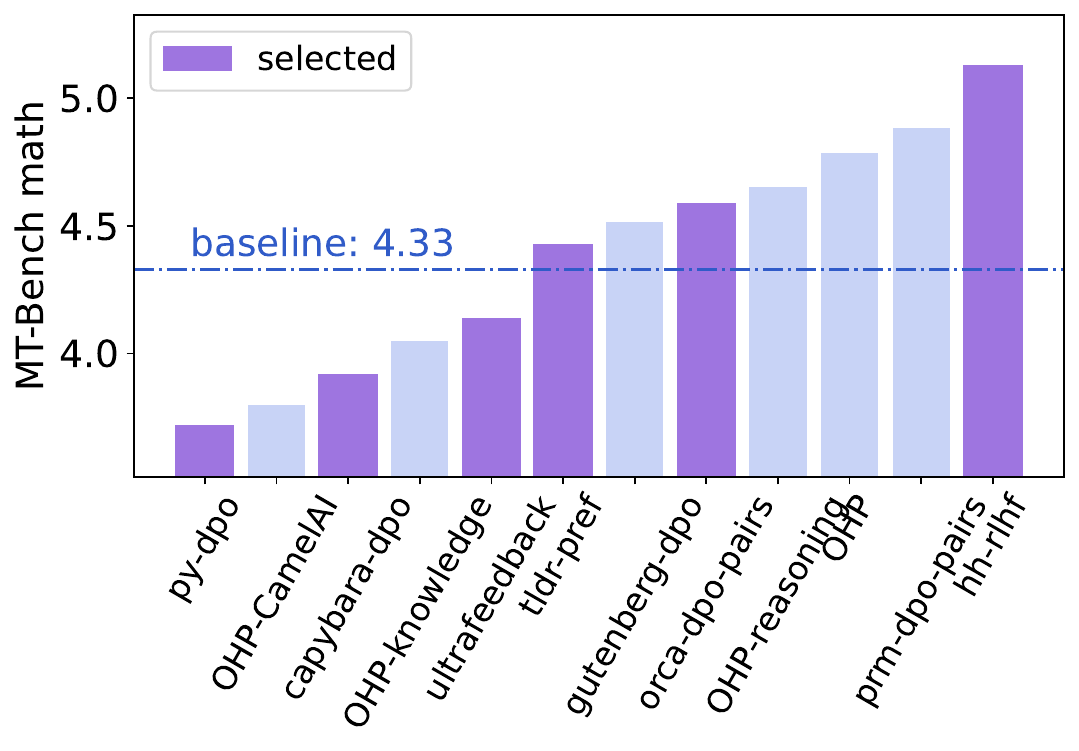}
}
\subfigure[Reasoning]{
    \includegraphics[scale=0.34]{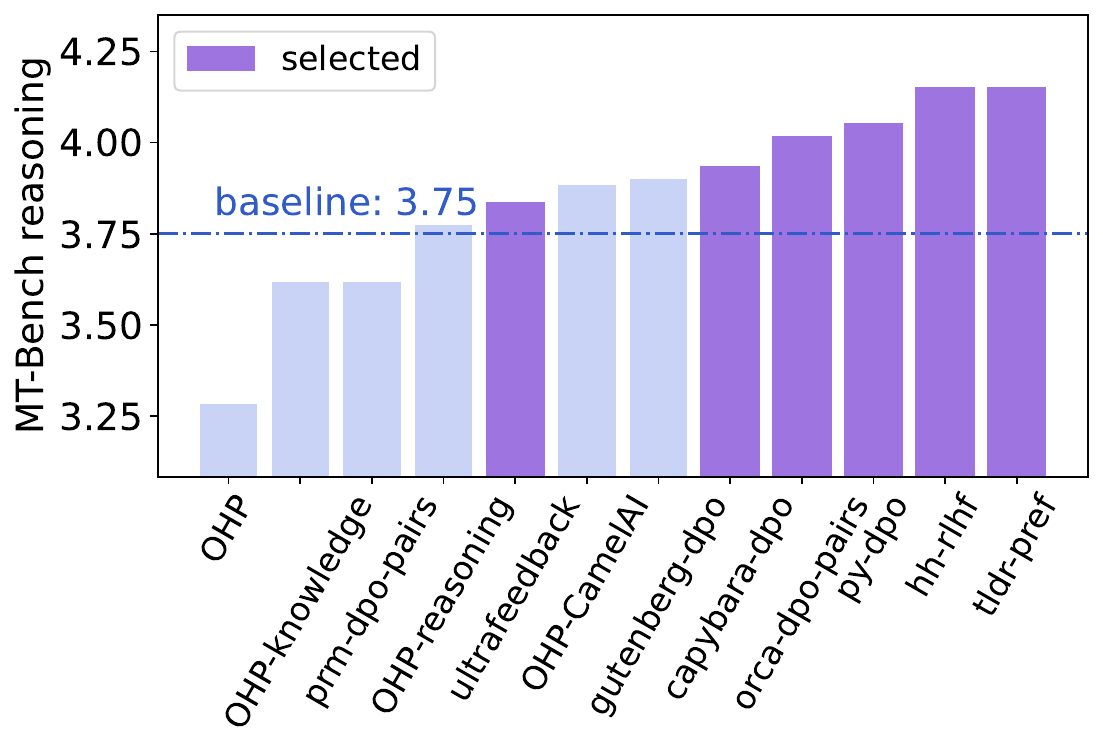}
}
\qquad
\subfigure[Roleplay]{
    \includegraphics[scale=0.34]{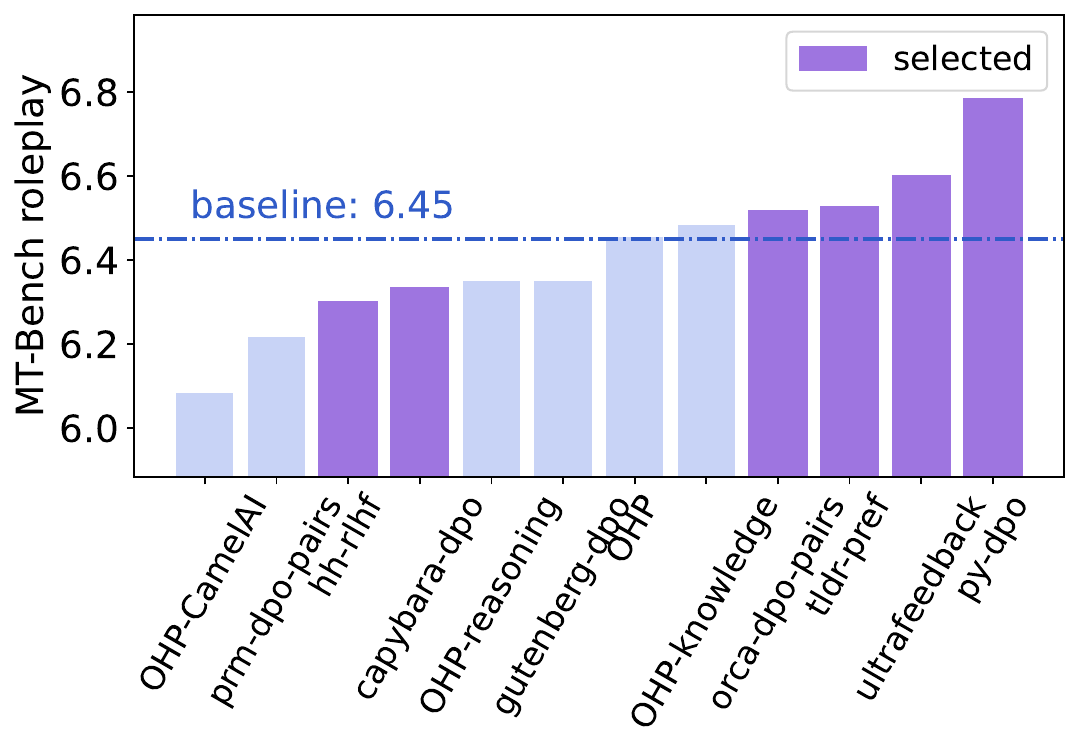}
}
\subfigure[Stem]{
    \includegraphics[scale=0.34]{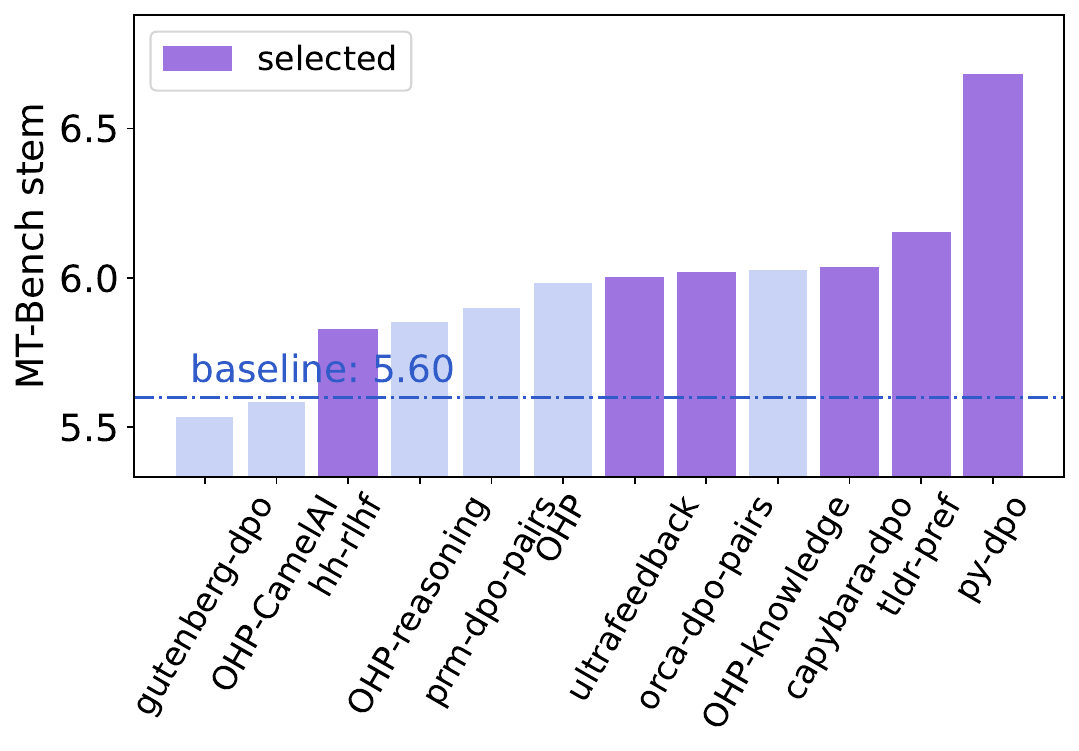}
}
\qquad
\subfigure[Writing]{
    \includegraphics[scale=0.34]{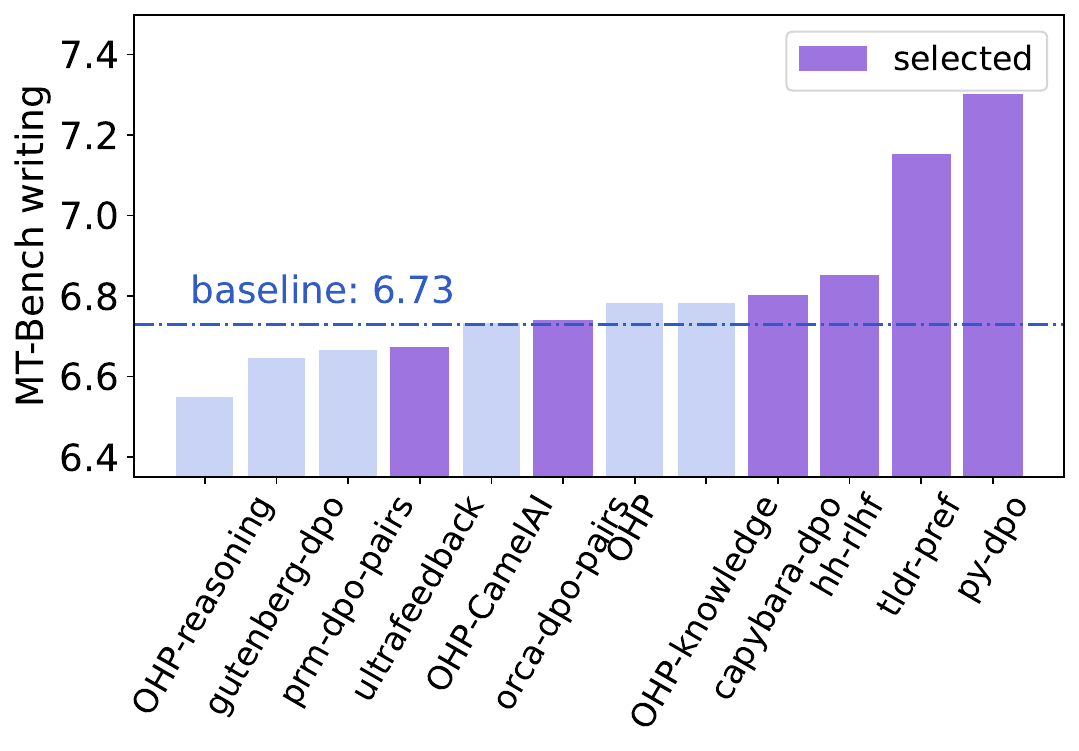}
}
\caption{MT-bench performance after training Gemma-2b on upto 10k samples from each dataset. Datasets we used in \Cref{subsec:Offline Preference Learning} are colored in purple.}
\label{fig:data-eval-mt-bench-dist}
\end{figure*}

To this end, we first selected a collection of 12 datasets, and empirically measure each dataset's quality by 1) sample upto 10k samples from each dataset, 2) train SFT-finetuned Gemma-2b and record its performance on MT-bench. We present the results in \Cref{fig:data-eval-mt-bench} and \Cref{fig:data-eval-mt-bench-dist}. We then used the top-six datasets according to their overall score in our offline preference learning experiments in \Cref{subsec:Offline Preference Learning}. The baseline is our $\pisft$ model.
% To this end, we first selected a collection of 12 datasets, and measure each dataset's quality by training a language model on it and recording its performance change on MT-bench, and then construct a mixture that is based on the performance improvements. Specifically, we first measure each dataset's quality by training Gemma-2b on a 10k subset (for dataset less than 10k in size, we use the full dataset) and measure its performance on MT-bench. We then included the top-six datasets according their overall score, and any dataset that achieves top-one score in any individual category. Conveniently, we find that the top-six datasets not only significantly outperform baseline, they also cover top-one dataset for any category.

\subsection{Training Hyperparams for Scaling DPO}
\label{subsec:Training Hyperparams for Scaling DPO}
For all runs, we finetune from the best $\pisft$ obtained from \Cref{subsec:Supervied Finetuning}. We use a sequence length of 2048, $\beta=0.1$, batch size of 128, learning rate of 5e-7, and vary training steps for each run. For $|D|$=100k, we continue training from previous runs instead of starting from scratch to save compute.

In addition to our result in \Cref{fig:scaling-dpo} measuring performance in OpenLLM and Arena-Hard-Auto*, we also present other metrics such as evaluation loss, reward margin, and reward accuracy in \Cref{tbl:eval-inconsistent}. We note that the evaluation loss and reward margin are \emph{inconsistent} with the performance in Arena-Hard-Auto* (or OpenLLM). This indicates that simple metrics such as evaluation loss and reward margin may not be good indicators of model final performance.
\subsection{Reward Annotation for Data Filtering}
\label{subsec:reward_annotation}

While datasets such as UltraFeedback \cite{cui2023ultrafeedback} provide ratings in [1,10] for chosen/rejected responses, other datasets such as TLDR \cite{stienon2020learning} and HH-RLHF \cite{bai2022training} does not. This makes methods such as filtering based on score difference (e.g., \framework{Argilla}) not applicable.

\begin{figure*}[t!]
\centering
\subfigure[Original]{
    \includegraphics[scale=0.34]{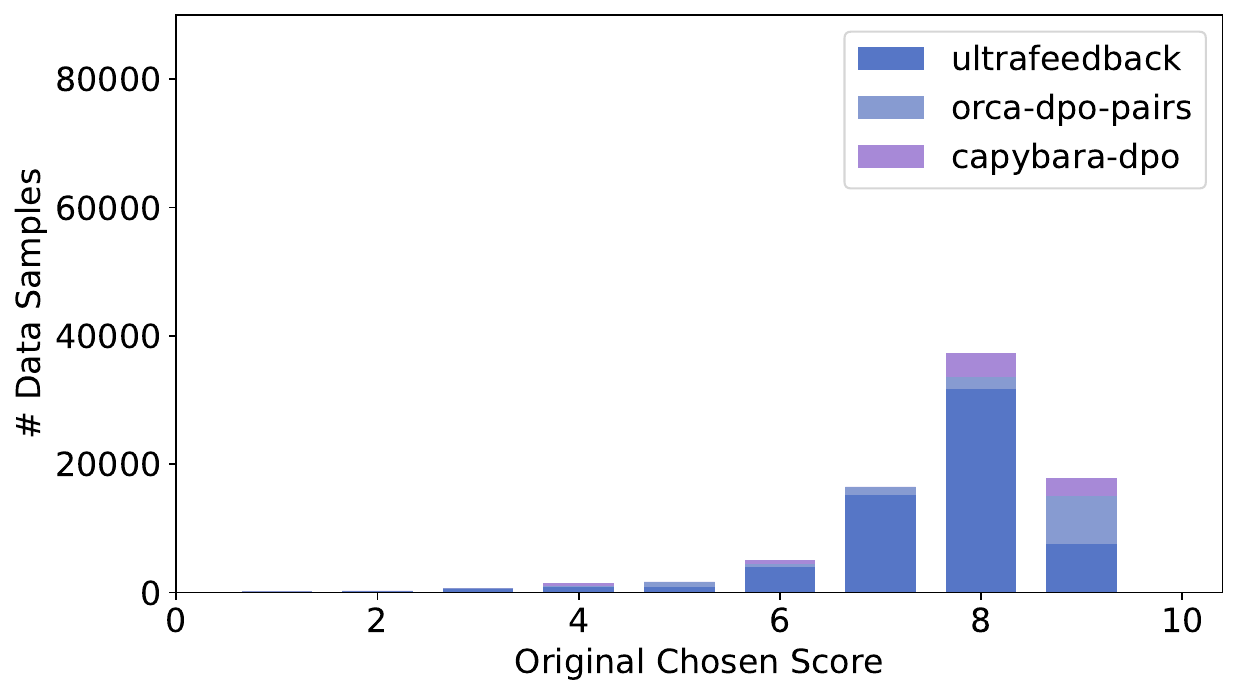}
    \includegraphics[scale=0.34]{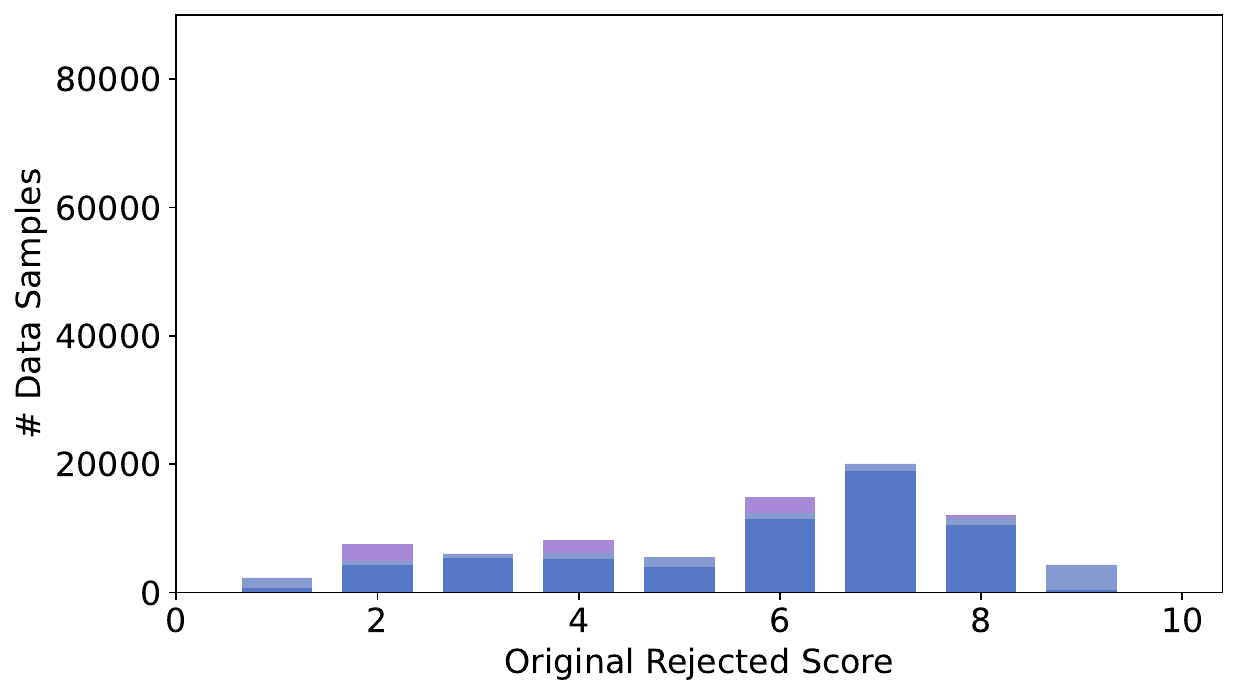}
}
\subfigure[Predicted]{
    \includegraphics[scale=0.34]{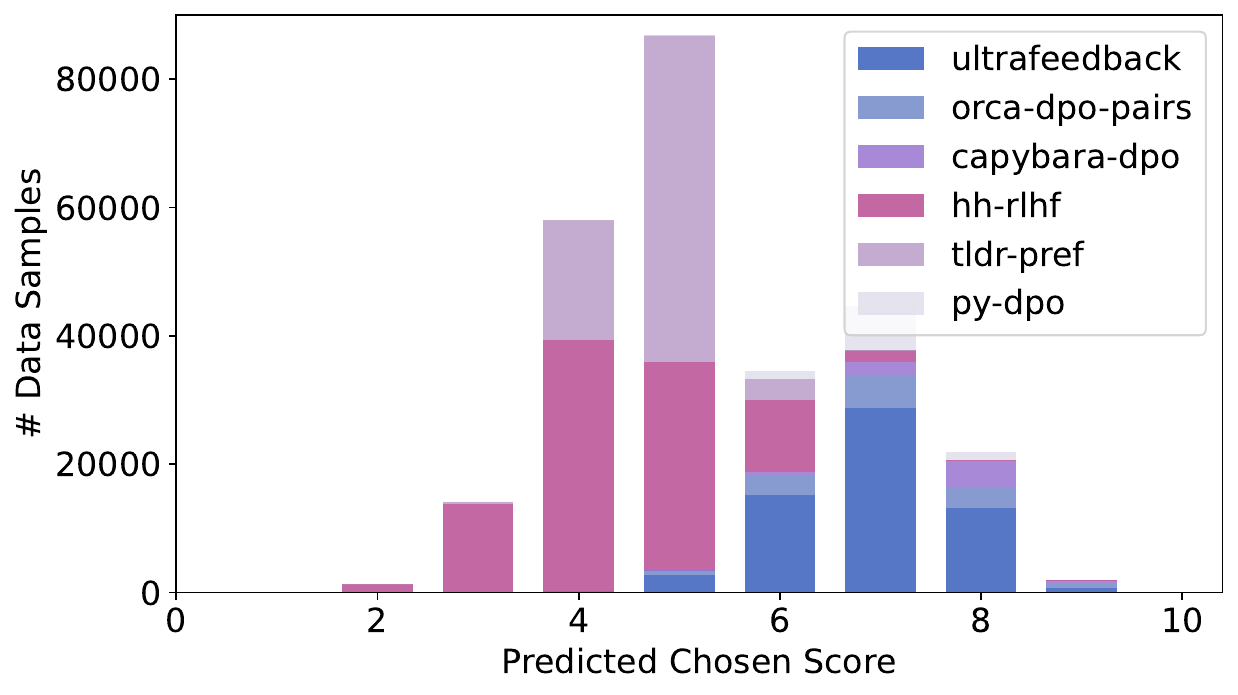}
    \includegraphics[scale=0.34]{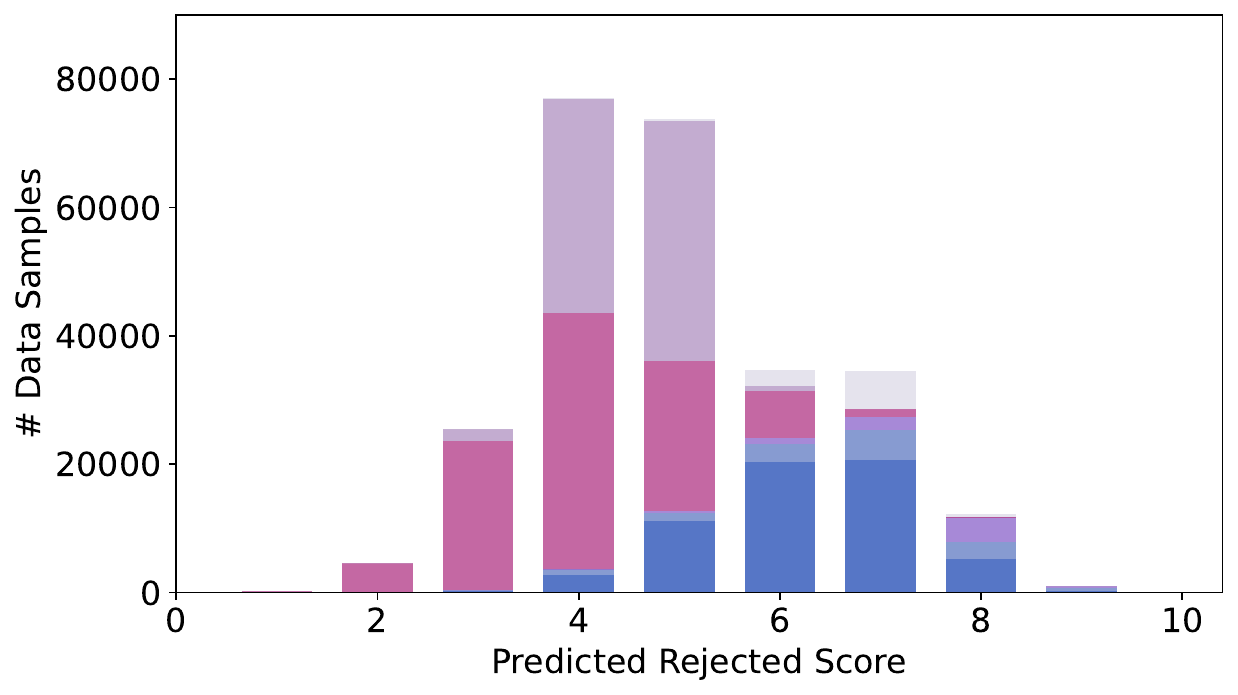}
}
\caption{Score distribution for the dataset we used in \Cref{subsec:Offline Preference Learning}. (a) Responses from Ultrafeedback, Orca-DPO-pairs, and Capybara-DPO already contain scores annotated by GPT-4/GPT-4-turbo. (b) We rescaled the score prediction produced by Nexusflow/Starling-RM-34B.}
\label{fig:rm-labeled-dist}
\end{figure*}

To this end, we consider a simple approach to use Nexusflow/Starling-RM-34B \cite{starling2023}, the best reward model according reward-bench \cite{lambert2024rewardbench}, to provide a score prediction to all of our training data. Specifically, we first used the reward model to compute a real value score for the prompt + chosen response and prompt + rejected response separately.
Next, since the predicted score is a real value, we 1) rescale it to $[0,1]$ to obtain $\tilde{s}$, and then 2) consider a least square solution to find $a,b$ under the function:
$$
\hat{s} = \text{clip}(a\cdot \tilde{s} + b, \min=0, \max=10)
$$
where the least square error between the true score $s$ for data that contains a GPT-4 annotated score and the rescaled $\hat{s}$ is minimized. This results in $a=11.1745,b=1.1791$. The average least square error and absolute error is $3.5389$ and $1.4278$, respectively. Finally, we augment the entire 264K training data with this score $\hat{s}$, and present the score distribution in \Cref{fig:rm-labeled-dist}.
\begin{figure*}[t!]
    \includegraphics[scale=0.75]{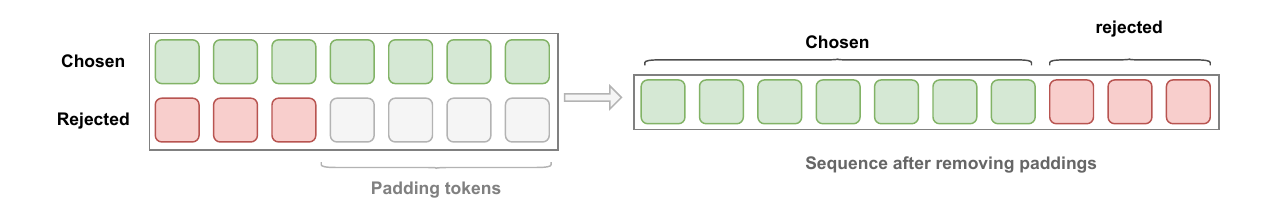}
    \caption{Illustration of efficient DPO implementation. Traditional DPO training requires adding padding tokens to the batch. Our implementation can remove the need of paddding tokens, and thus improving the training efficiency.}
    \label{fig:fast_dpo_implementation}
\end{figure*}

\begin{table*}[!t]
    \centering
    \scalebox{0.82}{
      \begin{tabular}{l lc cccc}
        \toprule
        FLOPs & $|D|$ & \textbf{Arena-Hard-Auto*} & \textbf{Eval Loss} & \textbf{Eval Reward Margin} & \textbf{Eval Reward Accurarcy}\\
        \midrule
  % gemma-2b-lion-v0.5-mix-100k-beta0.1-epoch5-from3
  % gemma-2b-lion-v0.5-mix-10k-beta0.1-epoch30-from26
  % gemma-2b-lion-v0.5-mix-100k-beta0.1-epoch2
  % gemma-2b-lion-v0.5-mix-10k-beta0.1-epoch2
  % gemma-2b-lion-v0.5-mix-1k-beta0.1-epoch2
        3.1e19 & 100k & \textbf{14.8} & 0.6507 & 0.5375 & 0.6554\\
        1.8e19 & 10k  & 13.2 & 1.3300 & \textbf{0.7851} & 0.5738\\
        6.2e18 & 100k & 11.1 & 0.6389 & 0.3970 & 0.6334\\
        1.2e18 & 10k  & 10.7 & \textbf{0.6333} & 0.2082 & \textbf{0.6922}\\
        1.1e17 & 1k   & 9.3  & 0.6940 & -0.0039 & 0.4858\\
        \bottomrule
      \end{tabular}
    }
    \caption{Automatic evaluation metrics such as loss, reward margin, and reward accuracies on test set is inconsistent with final performance on benchmarks such as Arena-Hard-Auto*. All runs used $\beta=0.1$.}
    \label{tbl:eval-inconsistent}
    \vspace{-5pt}
  \end{table*}
\begin{figure*}[h!]
\centering
\subfigure[OpenLLM Performance]{
    \includegraphics[scale=0.37]{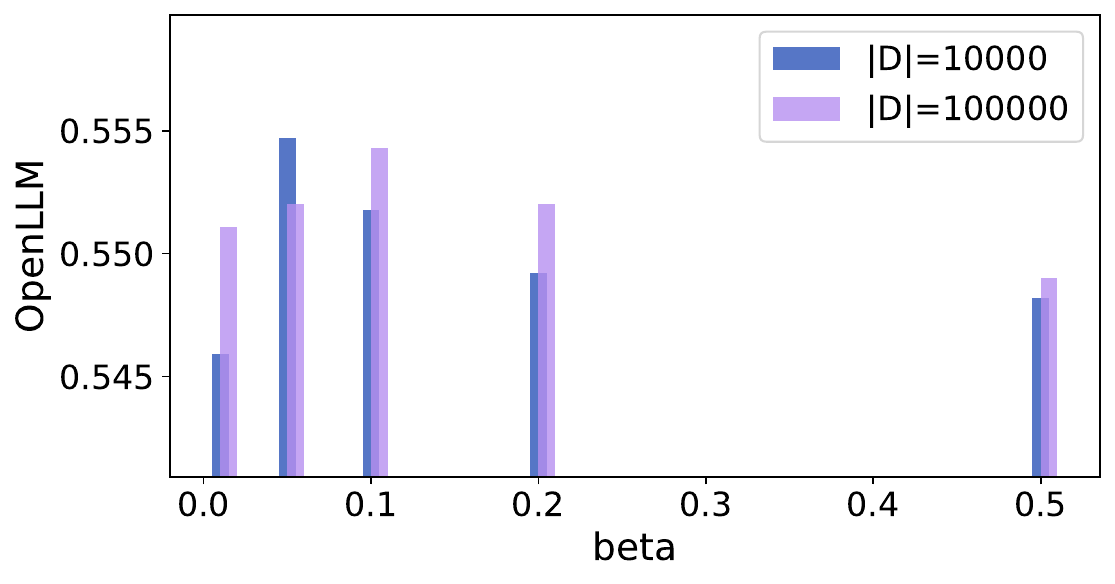}
}
\subfigure[Arena Hard Auto* Performance]{
    \includegraphics[scale=0.37]{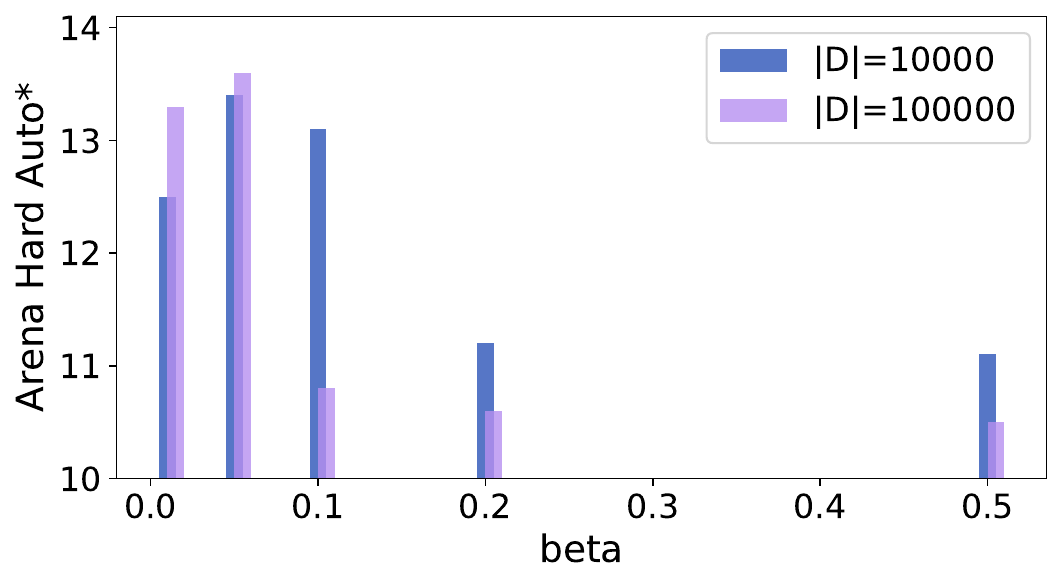}
}
\caption{Effect of $\beta$ on model performance across datasets of different sizes.}
\label{fig:dpo-hyperparam-beta-full}
\end{figure*}
\begin{figure*}[t!]
    \centering
    \subfigure{
        \raisebox{0.18\height}{
            \includegraphics[scale=0.32]{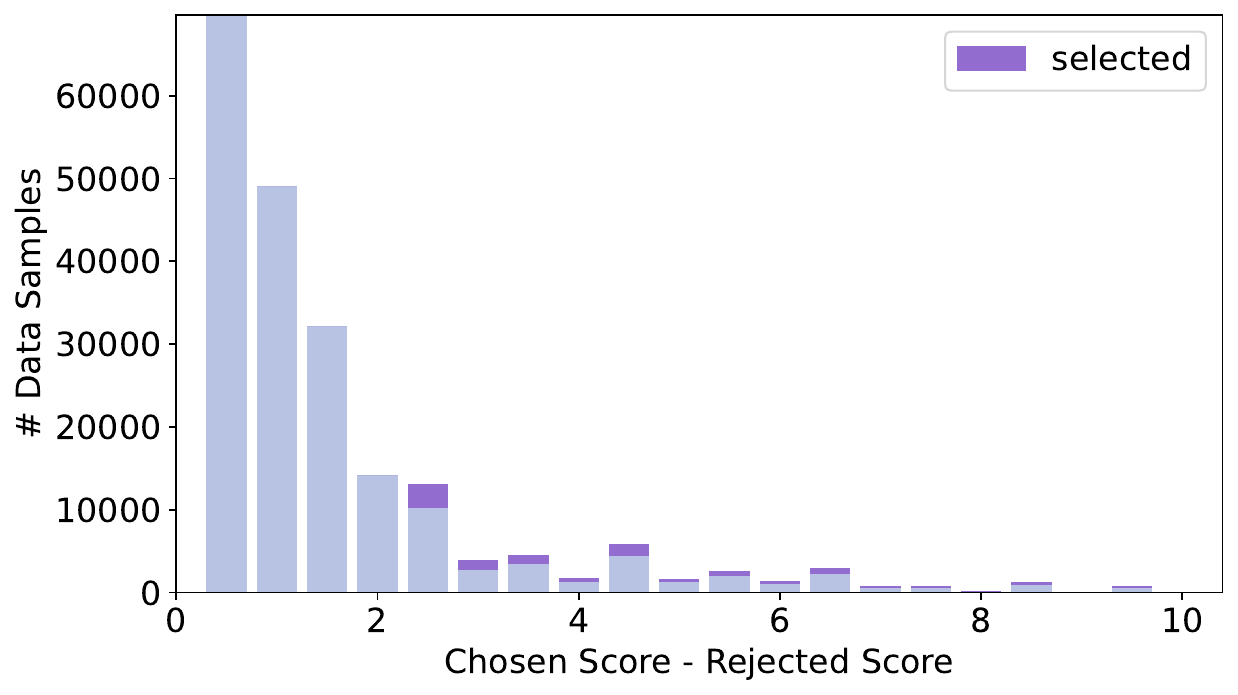}
        }
        \includegraphics[scale=0.32]{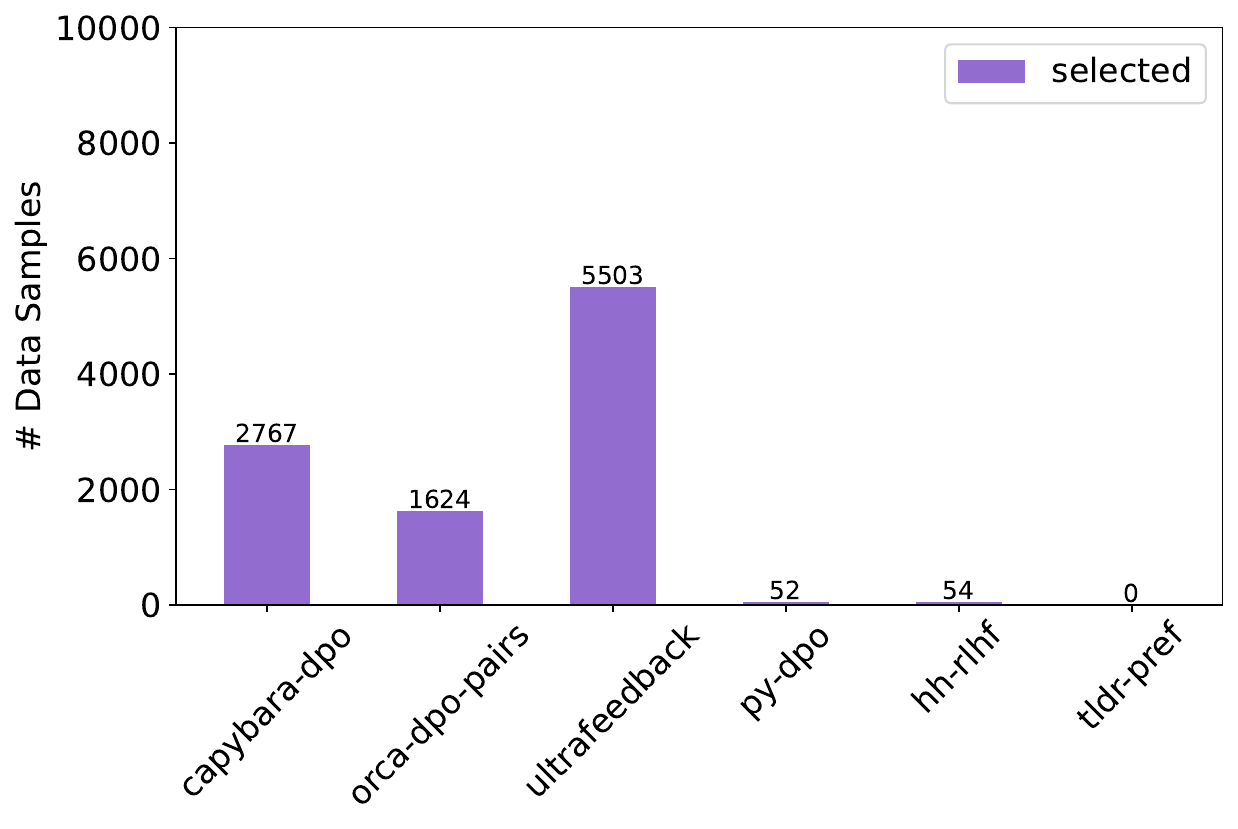}
    }
\caption{Data distribution for \framework{A2}, which 1) removes DPO pairs with score difference less than two, and 2) sample 10k data that has the highest chosen score in each score difference bin.}
\label{fig:data-selection-dist}
\end{figure*}

Note that this reward model achieves 67.72\% accuracy over the entire 264K dataset (both with and without our score transformation). This indicates that  $\sim$30\% of the predicted score might not be accurate. Therefore, we use the original score annotation when available, and use the predicted and rescaled score only when necessary.

\subsection{More Details on Dataset Filtering Algorithms}
\label{subsec:Dataset Filtering Algorithms}

\begin{figure*}[t!]
\centering
\subfigure[Alpagasus-10k]{
    \raisebox{0.18\height}{
        \includegraphics[scale=0.32]{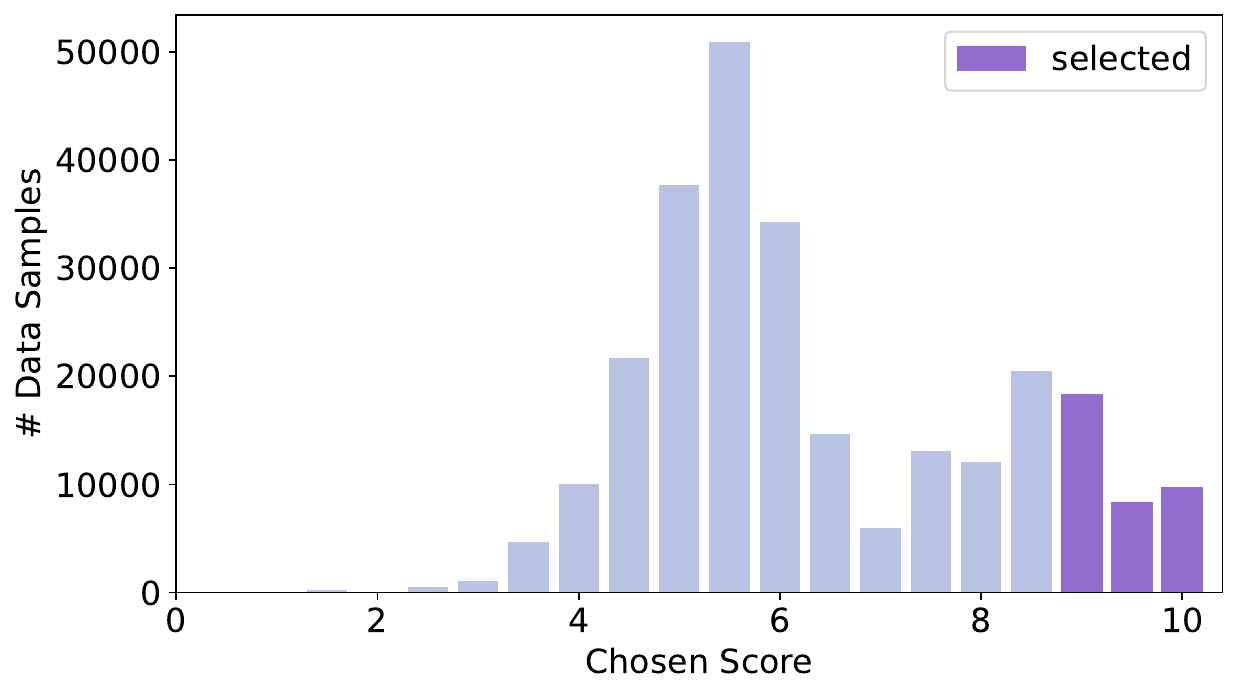}
    }
    \includegraphics[scale=0.32]{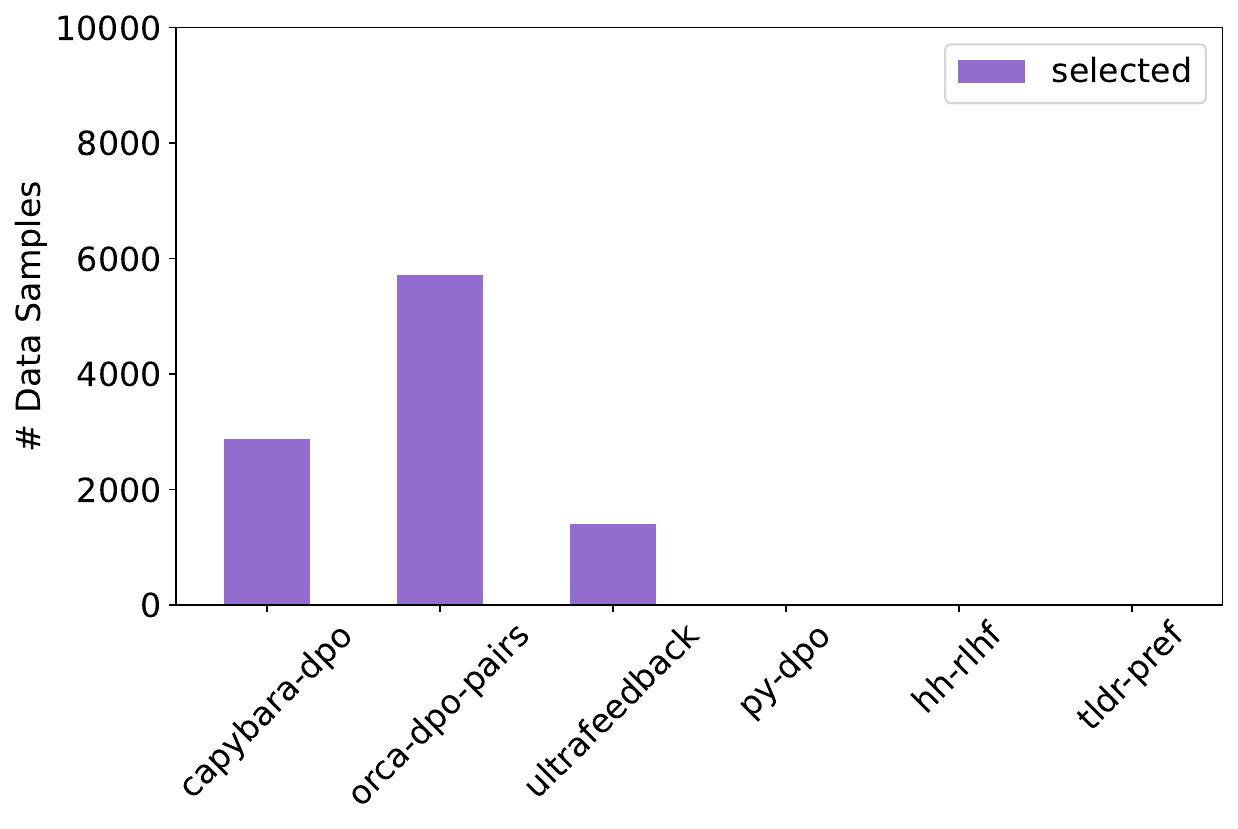}
}
\subfigure[Longest-10k]{
    \raisebox{0.18\height}{
        \includegraphics[scale=0.32]{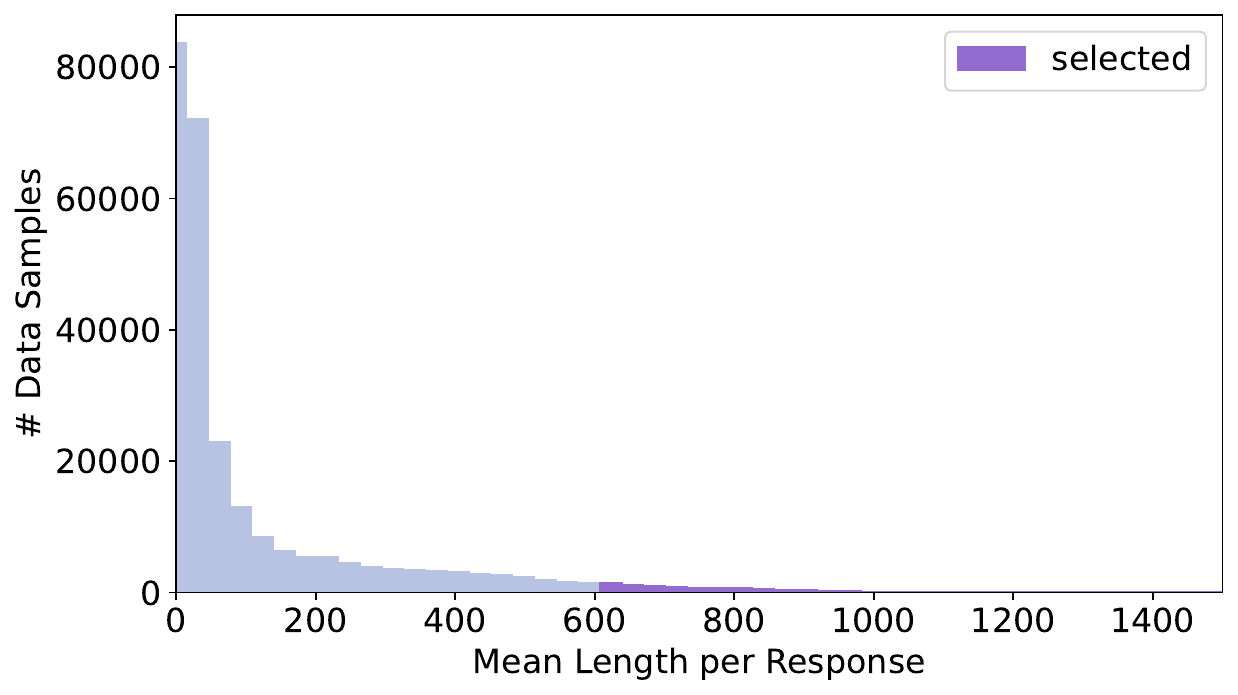}
    }
    \includegraphics[scale=0.32]{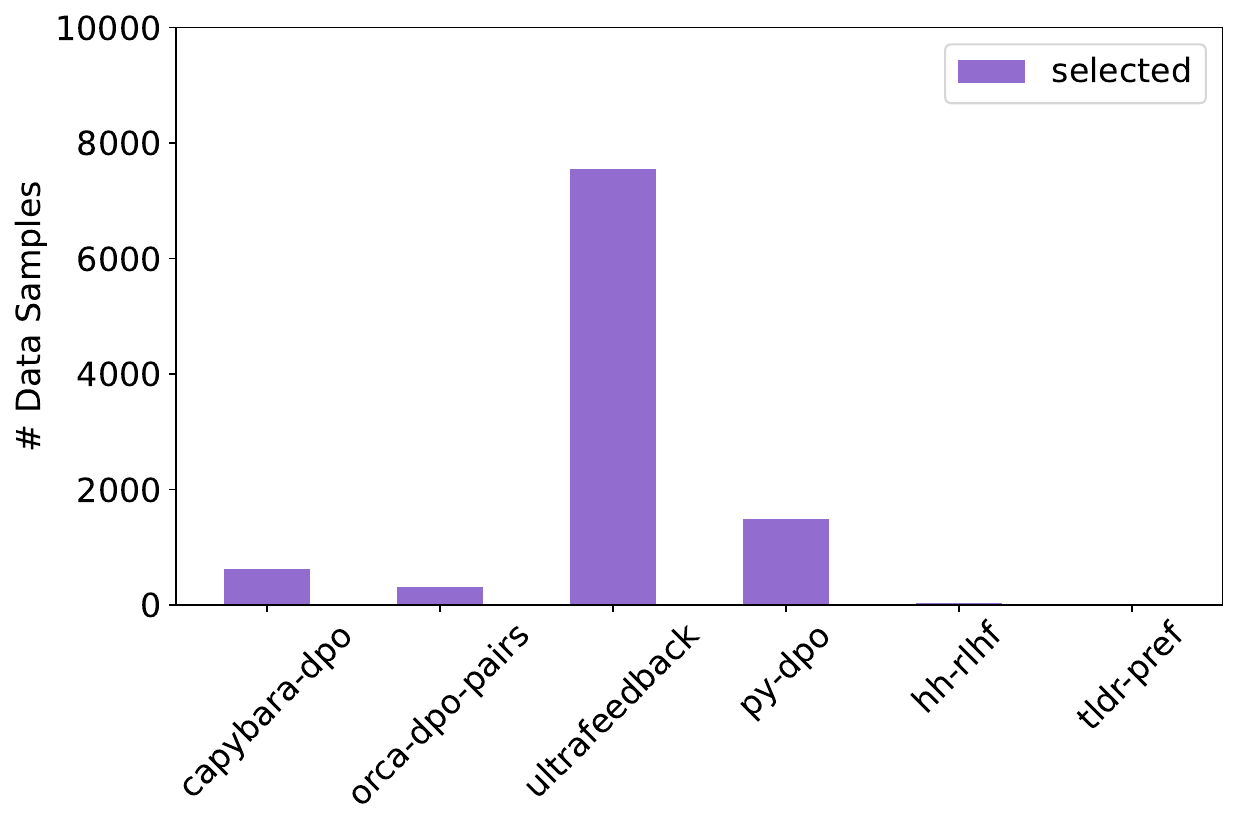}
}
\subfigure[DEITA-10k]{
    \raisebox{0.18\height}{
        \includegraphics[scale=0.32]{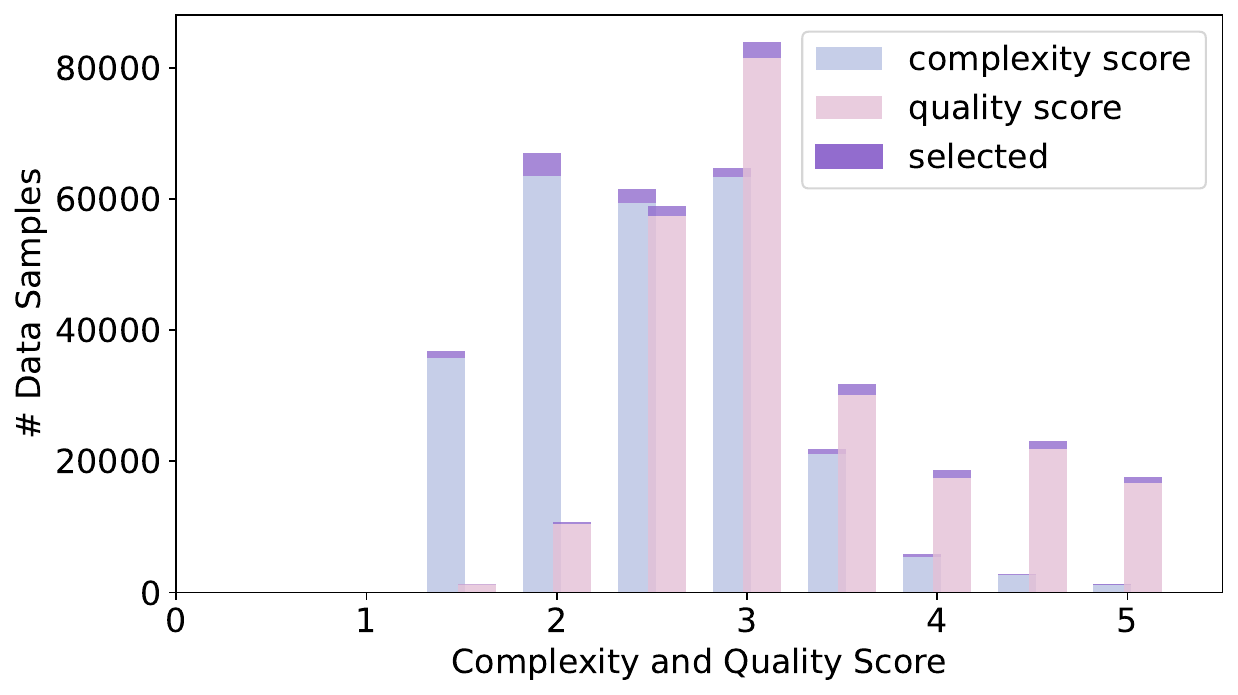}
    }
    \includegraphics[scale=0.32]{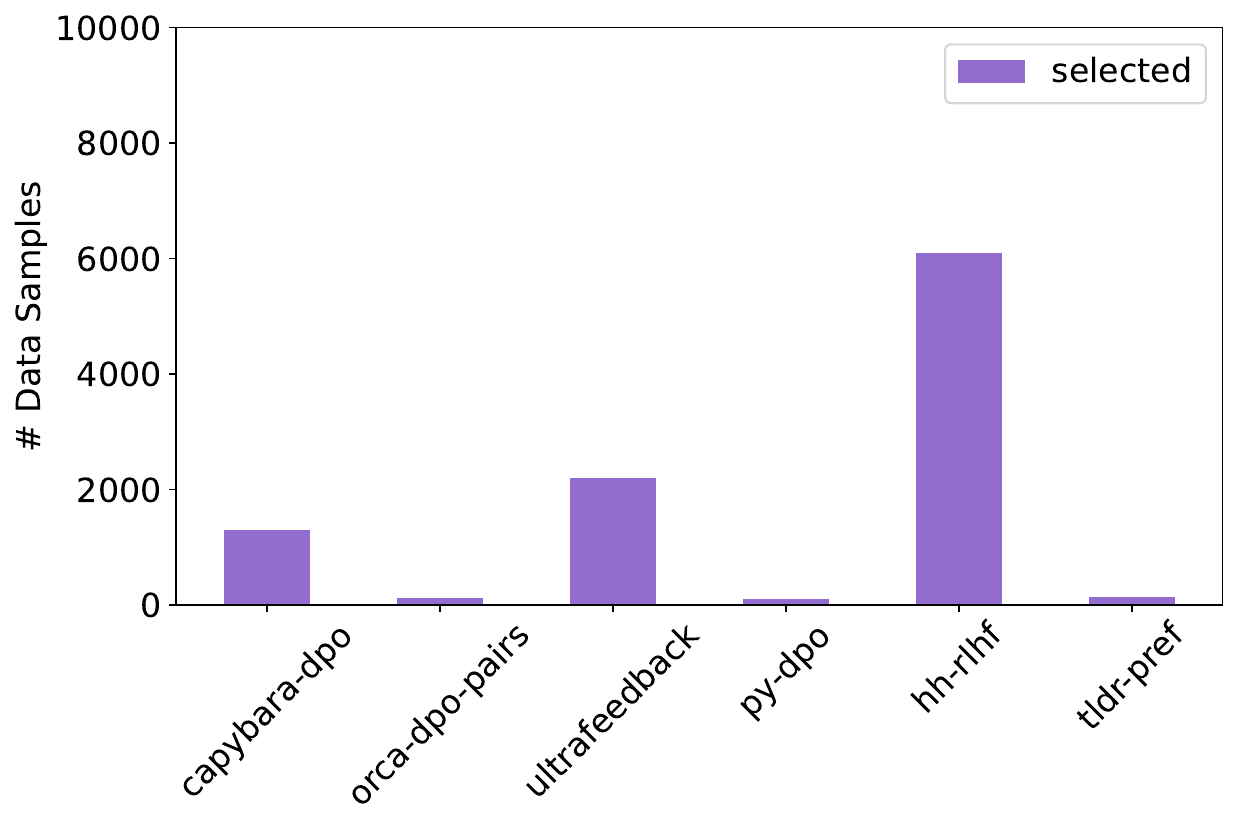}
}
\subfigure[Argilla-10k]{
    \raisebox{0.18\height}{
        \includegraphics[scale=0.32]{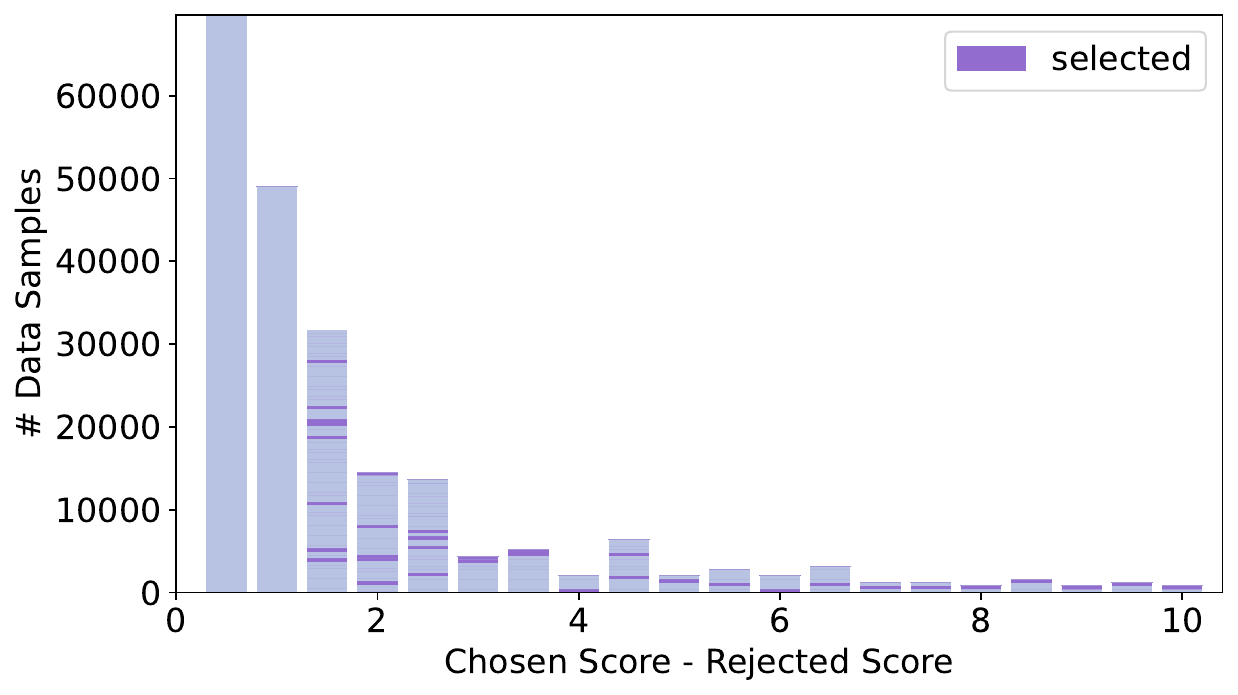}
    }
    \includegraphics[scale=0.32]{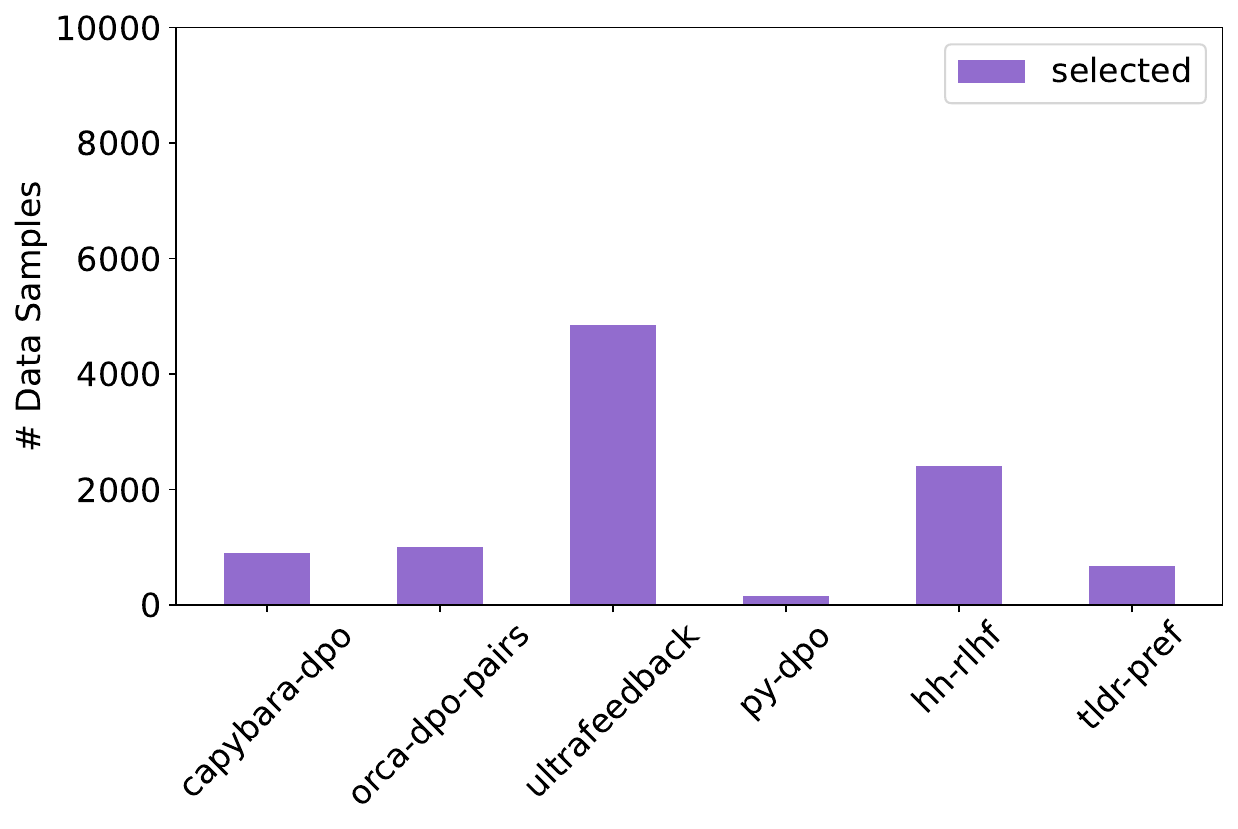}
}
\caption{Data distribution after applying the respective filtering algorithms.}
\label{fig:data-selection-dist-full}
\end{figure*}

We consider data filtering algorithms both from the instruction-tuning domain and from the preference learning domain.
This include algorithms such as \framework{Deita} \cite{liu2024makes-deita}, \framework{Longest} \cite{zhao2024longismore}, \framework{Alpagasus} \cite{chen2024alpagasus}, and \framework{Argilla} \cite{argilla-dpo-mix}. We also consider \framework{A2}, which can be seen as a combination of \framework{Argilla} and \framework{Alpagasus}: first removing pairs that has a score difference of less than two, and then sampling 10k data that has the highest chosen score in each bin. We apply each of the algorithms above to select 10k data from the 264K shown in \Cref{tab:dpo_data_stats}.
We present the selected data distributions for each algorithm (except for \framework{A2}) in \Cref{fig:data-selection-dist-full}, and for \framework{A2} in \Cref{fig:data-selection-dist}.

\section{More Details on Training \framework{Lion} Series}

\subsection{Efficient DPO Implementation}

\begin{table}[h]
    \centering
    \scalebox{0.8}{
    \begin{tabular}{lc}
        \toprule
        \textbf{Configuration} & \textbf{Training Time} \\
        \midrule
        LLaMA3-8b DPO & 15.72 hours \\
        \qquad + Fast Implementation (ours) & 11.40 hours \\
        \midrule
        Improvement & 27.48\% \\
        \bottomrule
    \end{tabular}
    }
    \caption{DPO Training times for different configurations.}
    \label{tab:fast_dpo_training_times}
\end{table}

In this work, we introduced an efficient DPO implementation for Transformers.
The motivation is to eliminate the computation overhead caused by padding tokens, as in DPO, chosen and rejected samples normally have varied lengths.
Our approach involves removing all padding tokens within a batch and concatenating the remaining sequences into a single, continuous sequence.
To handle sequence boundaries effectively in the self-attention layers, we utilize FlashAttention \cite{dao2023flashattention2}. This ensures that the removed padding tokens do not interfere with the processing of the valid tokens. An illustration of this process can be found in Figure~\ref{fig:fast_dpo_implementation}.

We evaluated the training times for LLAMA3-DPO with and without the fast DPO model implementation. 
The experiments were conducted using the specified offline preference dataset, running on a setup of four A100 80GB GPUs.
As shown in Table~\ref{tab:fast_dpo_training_times}, our fast DPO model implementation achieves a 27.48\% speed improvement.

\subsection{Training Details}

All the training experiments in this paper were conducted on 4×A100 80GB GPUs.
We used Deepspeed \cite{Rasley2020DeepSpeedSO} for all our experiments as we find that storingin model weights in fp32 is essential for DPO's performance as learning rate is small.
For other training details, please see Table~\ref{tab:sft_training_details} and Table~\ref{tab:dpo_training_details}.

\begin{table}[t]
\centering
\scalebox{0.8}{
\begin{tabular}{lccc}
\toprule
\bf Hyperparam  & \bf Gemma-2b & \bf LLaMA-3-8b \\
\midrule 
Warmup ratio & 0.1 & 0.1 \\
Peak Learning Rate & 2e-5 & 2e-5 \\
Max Sequence Length & 8192 & 8192 \\
Batch Size & 16 & 64\\
Weight Decay & 0.0 & 0.0 \\
Number Epochs & 3 & 3 \\
Learning Rate Decay & Cosine & Cosine \\
Max Grad Norm & 1.0 & 1.0 \\
\midrule
Training Time & 25.80 hrs & 56.36 hrs \\
\bottomrule
\end{tabular}
}
\caption{SFT Training Details. Training time is measured on 4 GPUs configuration.}
\label{tab:sft_training_details}
\end{table}

\begin{table}[t]
\centering
\scalebox{0.8}{
\begin{tabular}{lccc}
\toprule
\bf Hyperparam  & \bf Gemma-2b & \bf LLaMA-3-8b \\
\midrule 
Beta & 0.05 & 0.01 \\
Warmup ratio & 0.1 & 0.1 \\
Max Sequence Length & 2048 & 2048 \\
Peak Learning Rate & 5e-7 & 5e-7 \\
Batch Size & 64 & 128\\
Weight Decay & 0.0 & 0.0 \\
Number Epochs & 2 & 1 \\
Learning Rate Decay & Cosine & Cosine \\
Max Grad Norm & 1.0 & 1.0 \\
\midrule
Training Time & 10.15 hrs & 11.40 hrs \\
\bottomrule
\end{tabular}
}
\caption{DPO Training Details. Training time is measured on 4 GPUs configuration.}
\label{tab:dpo_training_details}
\end{table}

\subsection{Training Datasets}
\label{subsec:LION Training Datasets}

Following our findings in \Cref{sec:Training Procedure Analysis}, we train the \framework{Lion} series using a combination of datasets from the instruction-tuning domain and the preference learning domain. For SFT, we use the same dataset collection as in \Cref{tab:sft_data_stats}. Given the scaling trends of offline DPO, we add in more preference datasets such as Nectar \cite{starling2023} in addition to \Cref{tab:dpo_data_stats}. We summarize the datasets for training the \framework{Lion} below:

\paragraph{Details of Supervised Fine-Tuning Data}
\begin{itemize}
    \item \textbf{OpenHermes-2.5}  \cite{OpenHermes2.5}:
    The OpenHermes-2.5 dataset contains 1 million diverse, synthetic samples. It includes data from various sources like \href{https://huggingface.co/datasets/jondurbin/airoboros-3.2?not-for-all-audiences=true}{Airoboros}, CamelAI \cite{li2023camel}, \href{https://huggingface.co/datasets/lmsys/lmsys-chat-1m}{ChatBot Arena}, and several others, each contributing to fields ranging from physics and mathematics to code assistance and medical tasks. Please check the \href{https://huggingface.co/datasets/teknium/OpenHermes-2.5}{repo} for details.

    \item \textbf{MetaMathQA} \cite{yu2023metamath}:
    MetaMathQA is created using question bootstrapping, where mathematical questions are rewritten from GSM \cite{cobbe2021gsm8k} and Math \cite{hendrycksmath2021} dataset. The dataset is further enriched by rephrasing questions and using rejection sampling to select only correctly answered paths, enhancing diversity and reasoning capabilities.
    
    \item \textbf{SlimOrca} \cite{SlimOrca}:
    The SlimOrca dataset is a curated subset of the OpenOrca \cite{mukherjee2023orca} data, containing about 500,000 GPT-4 completions refined using human annotations from the FLAN \cite{longpre2023flan} dataset to remove incorrect answers. 

    \item \textbf{UltraChat} \cite{ding2023enhancing}:
    UltraChat is large-scale, informative, and diverse multi-round dialogue dataset aimed at improving language model conversational skills. 
    It contains 1.5M samples with  a wide range of topics and instructions.

    \item \textbf{OrcaMath} \cite{mitra2024orcamath}:
    OrcaMath comprises 200,000 synthetic mathematical problems created using a collaborative multi-agent setup with GPT-4. 
    
    \item \textbf{Capybara} \cite{daniele2023amplify-instruct}:
    Capybara uses the Amplify-Instruct method to create synthetic multi-turn conversations from quality single-turn seeds. 
    It focuses on diverse, logical reasoning across domains, with each conversation exploring deep, diverse topics. 

    \item \textbf{Deita-10k} \cite{liu2024what}:
    Deita is an open-source dataset aimed at enhancing instruction tuning for Large Language Models (LLMs) through Automatic Data Selection. 
    It incorporates a dataset of 10,000 high-quality, alignment-specific Supervised Fine-Tuning (SFT) data points. This data is primarily selected from larger datasets including 58K entries from ShareGPT \cite{vicuna2023}, 105K from UltraChat \cite{ding2023enhancing}, and a 143K mixture from WizardLM \cite{Xu2023WizardLMEL} data \cite{luo2023wizardcoder}. 
\end{itemize}

\begin{table*}[t]
\centering
\scalebox{0.85}{
\begin{tabular}{lccccccc}
\toprule
\textbf{Model} & \textbf{ARC} & \textbf{HellaSwag} & \textbf{MMLU} & \textbf{TruthfulQA} & \textbf{Winogrande} & \textbf{GSM8k} & \textbf{Average} \\
\midrule
Gemma-2b & 48.38 & 71.77 & 41.77 & 33.08 & \textbf{66.30} & 16.91 & 46.51 \\
Gemma-2b-it & 43.60 &	62.55 & 36.95 & \textbf{45.85} & 61.80 & 10.99 &	43.62 \\
Gemma-2b-lion-sft (ours) & 50.94 & 70.65 & 45.04 & 43.80 & 64.88 & 53.37 & 54.78 \\
Gemma-2b-lion-dpo (ours) & 52.30	& 72.47 &45.31 &	45.06 & 65.19 & 51.78 & 55.35 \\
Gemma-2b-lion-odpo (ours) & \textbf{53.75} & \textbf{73.04} & \textbf{45.52} & 45.66 & 64.40 & \textbf{53.53} & \textbf{55.98} \\
\midrule
LLaMA-3-8b-Base & 58.02	& 82.15 & 65.09&	43.92 & \textbf{77.58} & 51.55 & 63.05 \\
LLaMA-3-8b-Instruct & 61.86	&78.79&	\textbf{65.70} &	51.64&	75.30&	75.13 & 68.07 \\
LLaMA-3-8b-lion-sft (ours) & 59.64 & 80.80 & 64.21 & 54.26 & 76.64 & 76.72 & 68.71 \\
LLaMA-3-8b-lion-dpo (ours) & 63.91	& 82.95 &	63.67 & 60.01 & 76.56 & \textbf{80.59} & 71.28 \\
LLaMA-3-8b-lion-odpo (ours) & \textbf{63.99}	& \textbf{83.18} & 63.59 & \textbf{61.12} & 76.72 & 79.91 & \textbf{71.41} \\
\bottomrule
\end{tabular}
}
\caption{Detailed task evaluation results on OpenLLM.}
\label{tab:openllm_details}
\end{table*}

\paragraph{Details of Offline Preference Data}

\begin{itemize}
    \item \textbf{TLDR} \cite{stienon2020learning}: 
    This data is used to train a reward model for summarization. Summaries for the reward model came from the TL;DR dataset.
    We use the comparisons data, where annotators chose the better of two summaries.
    
    \item \textbf{PKU-SafeRLHF} \cite{safe-rlhf}:
    This dataset contains 83.4K preference entries annotated for harmlessness and helpfulness. 
    Each entry includes two responses to a question, with safety meta-labels and preferences. The responses came from Alpaca-7B, Alpaca2-7B, and Alpaca3-8B models, following SFT performed on Llama2-7B and Llama3-8B with the Alpaca 52K dataset.
    
    \item \textbf{HelpSteer} \cite{wang2023helpsteer}:
    It is open-source Helpfulness Dataset designed by NVIDIA to improve models' helpfulness, factual accuracy, and coherence, with adjustable response complexity and verbosity. It contains 37,120 samples, each including a prompt, a response, and five human-annotated attributes of the response, rated from 0 to 4: Helpfulness (overall helpfulness), Correctness (pertinence and accuracy of facts), Coherence (consistency and clarity), Complexity (intellectual depth), and Verbosity (amount of detail).
    
    \item \textbf{UltraFeedback} \cite{cui2023ultrafeedback}:
    UltraFeedback is a diverse preference dataset with 64k prompts and 256k responses from various sources, annotated by GPT-4 for instruction-following, truthfulness, honesty, and helpfulness. 
    It includes 380k high-quality feedback entries, allowing the creation of 1 million comparison pairs. 
    Prompts are sourced from datasets like UltraChat, ShareGPT, Evol-Instruct, TruthfulQA, FalseQA, and FLAN, ensuring broad representation and diversity.
    
    \item \textbf{Nectar} \cite{starling2023}:
    Nectar is a high-quality 7-wise comparison dataset. 
    It features diverse chat prompts from sources like \href{https://huggingface.co/datasets/lmsys/lmsys-chat-1m}{lmsys-chat-1M}, \href{https://sharegpt.com/}{ShareGPT}, \href{https://huggingface.co/datasets/Anthropic/hh-rlhf}{Antropic/hh-rlhf}, \href{https://huggingface.co/datasets/openbmb/UltraFeedback}{UltraFeedback}, \href{https://huggingface.co/datasets/WizardLM/WizardLM_evol_instruct_V2_196k}{Evol-Instruct}, and \href{https://huggingface.co/datasets/SirNeural/flan_v2}{Flan}. 
    Responses from models such as GPT-4, GPT-3.5-turbo, LLama-2-7B-chat, and Mistral-7B-Instruct are ranked by GPT-4, resulting in 3.8M pairwise comparisons.
    
    \item \textbf{Py-DPO}\footnote{\url{https://huggingface.co/datasets/jondurbin/py-dpo-v0.1}}:
    The DPO dataset enhances Python coding abilities using the validated \href{https://huggingface.co/datasets/Vezora/Tested-22k-Python-Alpaca}{Python-Alpaca} dataset for "chosen" responses. "Rejected" values, generated with a mix of \href{https://huggingface.co/jondurbin/airoboros-l2-13b-3.1.1}{airoboros-l2-13b-3.1.1} and \href{https://huggingface.co/jondurbin/bagel-7b-v0.1}{bagel-7b-v0.1}, are assumed to be of lower quality.
    
    \item \textbf{Distilabel-Capybara}\footnote{\url{https://huggingface.co/datasets/argilla/distilabel-capybara-dpo-7k-binarized}}:
    The Distilabel-Capybara dataset, created by \href{https://github.com/argilla-io/distilabel}{distilabel} addresses the lack of multi-turn open datasets for DPO/RLHF by providing multi-turn dialogue preferences on top of \href{https://huggingface.co/datasets/LDJnr/Capybara}{Capybara}.

    \item \textbf{Distilabel-Orca}\footnote{\url{https://huggingface.co/datasets/argilla/distilabel-intel-orca-dpo-pairs}}:
    Similar to Distilabel-Capybara, this dataset is created by \href{https://github.com/argilla-io/distilabel}{distilabel} 
    to generate preference labels on top of \href{https://huggingface.co/datasets/Intel/orca_dpo_pairs}{Orca}.
    
\end{itemize}

\noindent
\textbf{Details of Online Preference Data}
\vspace{1mm}

We used data from UltraFeedback \cite{cui2023ultrafeedback} as prompts. 
We sample multiple responses from $\pi_\theta$ and use Pair-RM \cite{jiang2023llmblender} as a judge to obtain preference pairs. 
This results in an online collected dataset of 60k in size.

\subsection{Performance Details}

% Table~\ref{tab:openllm_details} shows the detailed evaluation results for OpenLLM tasks.
% The performance of our aligned Gemma-2b models and LLaMA-3-8b models show significant improvements compared to the official instruct models. For the Gemma-2b model, the lion-sft, lion-dpo, and lion-odpo alignment training improved the average scores from 46.51 for the base model to 55.98 for the lion-odpo model. Similarly, for the LLaMA-3-8b model, the performance increased from 63.05 for the base model to 71.28 for the lion-dpo model and 71.22 for the lion-odpo model. These enhancements demonstrate the effectiveness of our alignment pipeline in surpassing the officially tuned instruct models which rely on closed-source data and algorithms.

Table~\ref{tab:openllm_details} presents a comprehensive breakdown of the performance of our Gemma-2b and LLaMA-3-8b models across various OpenLLM tasks, highlighting the improvements brought by the lion-sft, lion-dpo, and lion-odpo alignment training methods. The lion-sft model showed substantial improvements across all tasks, with significant gains in GSM8k (53.37) and TruthfulQA (43.80). Building on these improvements, the lion-dpo model particularly enhanced ARC (52.30) and HellaSwag (72.47), while maintaining strong performance in other tasks. The lion-odpo model achieved the highest scores overall, excelling in ARC (53.75) and HellaSwag (73.04), and maintaining superior performance in GSM8k (53.53).

For the LLaMA-3-8b model, the lion-sft variant displayed robust performance across all tasks, with notable scores in GSM8k (76.72) and TruthfulQA (54.26). The lion-dpo model further improved performance, achieving higher scores in ARC (63.91), HellaSwag (82.95), and significantly in GSM8k (80.59) and TruthfulQA (60.01). The lion-odpo model marginally outperformed the lion-dpo model, attaining the highest scores in ARC (63.99), HellaSwag (83.18), and TruthfulQA (61.12), while maintaining exceptional performance across all tasks.

Both the Gemma-2b and LLaMA-3-8b models benefit significantly from the alignment training, with each subsequent model (sft, dpo, odpo) showing progressive improvements. The lion-odpo models generally achieve the highest scores, demonstrating the effectiveness of this alignment method in enhancing model performance across diverse tasks.
% \begin{table*}[!t]
%     \centering
%     \scalebox{0.82}{
%       \begin{tabular}{l p{0.35\textwidth} p{0.5\textwidth}}
%         \toprule
%         Stage & Datasets & Descriptions \\
%         \midrule
%         SFT  & 2 & 3 \\
%         \cmidrule{1-2}
%         Offline DPO  & TLDR \cite{stienon2020learning}; PKU-SafeRLHF \cite{safe-rlhf}; HelpSteer \cite{wang2023helpsteer} & Preference pairs where responses are written by human/LLMs, and judged by human raters. \\
%         & UltraFeedback \cite{cui2023ultrafeedback}; Nectar \cite{starling2023}; Py-DPO\footnote{\url{https://huggingface.co/datasets/jondurbin/py-dpo-v0.1}}; Distilabel-Capybara\footnote{\url{https://huggingface.co/datasets/argilla/distilabel-capybara-dpo-7k-binarized}}; Distilabel-Orca\footnote{\url{https://huggingface.co/datasets/argilla/distilabel-intel-orca-dpo-pairs}} & Preference pairs where responses are written by LLMs, and and judged by GPT-4 or GPT-4-turbo \\
%         \cmidrule{1-2}
%         Online DPO & UltraFeedback \cite{cui2023ultrafeedback} & We follow \citet{meng2024simpo} and use prompts from UltraFeedback to generate model responses.\\
%         \bottomrule
%         \end{tabular}
%     }
%     \caption{Training datasets used by LION-series. All datasets are publicly available. Please refer to our GitHub repo for the portion of the dataset used in each stage.}
%     \label{tbl:training_datasets}
% \end{table*}

% \section{Efficient DPO Training}
\subsection{More Details on Qualitative Analysis}
\label{subsec:Details on Qualitative Analysis}
To provide insights into the opaque training process during preference learning, we additionally record how the sequence probability for $y_w$ and $y_l$ change after DPO training.
We use a fixed unseen test set of $(x,y_w,y_l)$ obtained from public datasets (see \Cref{subsec:DPO Analysis Dataset Selection}), and compute $\pi_\theta(y_w) - \pi_\theta(y_l)$ for each preference pair before and after training.
We then qualitatively compare various models from \Cref{tbl:main_exp} and from \Cref{sec:Training Procedure Analysis}. We present the visualizations in \Cref{fig:dpo-seqprob-change-full}.

In \Cref{fig:dpo-seqprob-change-full}, we find that many of the best-performing models have a parabolic shape as shown in \Cref{fig:dpo-seqprob-best}.
This indicates that well-trained models learn to not only increase confidence in pairs they could distinguish correctly before training (i.e., $\pi_\theta(y_w) > \pi_\theta(y_l)$), but also improve on pairs where they previously could not (i.e., $\pi_\theta(y_w) < \pi_\theta(y_l)$).
Undertrained models (\Cref{fig:dpo-seqprob-undertrained}) achieve a similar shape but with a much smaller magnitude.
While overtrained models (\Cref{fig:dpo-seqprob-overtrained}) show a change in probability at an even greater scale (e.g., for $\pi_\theta(y_w) < \pi_\theta(y_l)$), it is often achieved by sacrifising performance on the other side of the plot (e.g., $\pi_\theta(y_w) > \pi_\theta(y_l)$).

\end{document}